\documentclass[10pt,twocolumn,letterpaper]{article}
\usepackage[T1]{fontenc}
\usepackage[pagenumbers]{cvpr}
\usepackage{booktabs}
\usepackage{multirow}
\usepackage{placeins}
\usepackage{tikz}
\usetikzlibrary{arrows.meta,calc,positioning}
\definecolor{cvprblue}{rgb}{0.21,0.49,0.74}
\definecolor{blue}{rgb}{0,0,0} %
\usepackage[pagebackref,breaklinks,colorlinks,allcolors=cvprblue]{hyperref}

\def\confName{CVPR}
\def\confYear{2027}
\newcommand{\method}{D\'ej\`a-VAE\xspace}
\newcommand{\R}{\mathbb{R}}

\newcommand{\Ep}{E'}
\newcommand{\Dp}{D'}

\newcommand{\papertitle}{D\'ej\`a-VAE\xspace: Training-Free Consistent Editing\\ with Pseudo-Invertible Encoders and Latent Back-Projection}
\title{\vspace{-0.3cm}\papertitle}
\author{Ron Libman$^{1,}$\thanks{Equal contribution.} \qquad Yamit Ehrlich$^{1,}$\footnotemark[1] \qquad Ohad Fried$^{2}$ \qquad Assaf Shocher$^{1}$\\
$^{1}$Technion -- Israel Institute of Technology \qquad $^{2}$Reichman University}

\makeatletter
\g@addto@macro\@maketitle{%
  \begingroup
  \vspace{-0.4cm}
  \centering
  \begin{tikzpicture}[x=\textwidth,y=\textwidth,every node/.style={font=\footnotesize}]
\node[anchor=base,inner sep=0pt] at (0.27800,-0.01500) {\small\bfseries Edit / undo chain};
\node[anchor=base,inner sep=0pt] at (0.66300,-0.01500) {\small\bfseries Encode--decode};
\node[anchor=base,inner sep=0pt] at (0.89100,-0.01500) {\small\bfseries No-op edits};
\node[anchor=base,inner sep=0pt] at (0.05750,-0.03400) {Original};
\node[anchor=base,inner sep=0pt] at (0.13100,-0.03400) {+ bow};
\node[anchor=base,inner sep=0pt] at (0.20450,-0.03400) {+ glasses};
\node[anchor=base,inner sep=0pt] at (0.27800,-0.03400) {+ hat};
\node[anchor=base,inner sep=0pt] at (0.35150,-0.03400) {+ scarf};
\node[anchor=base,inner sep=0pt] at (0.42500,-0.03400) {+ bag};
\node[anchor=base,inner sep=0pt] at (0.49850,-0.03400) {Undo};
\node[anchor=base,inner sep=0pt] at (0.58950,-0.03400) {Original};
\node[anchor=base,inner sep=0pt] at (0.66300,-0.03400) {20$\times$};
\node[anchor=base,inner sep=0pt] at (0.73650,-0.03400) {60$\times$};
\node[anchor=base,inner sep=0pt] at (0.81750,-0.03400) {Original};
\node[anchor=base,inner sep=0pt] at (0.89100,-0.03400) {20$\times$};
\node[anchor=base,inner sep=0pt] at (0.96450,-0.03400) {60$\times$};
\node[anchor=north west,inner sep=0pt] at (0.02200,-0.03900) {\includegraphics[width=0.07100\textwidth]{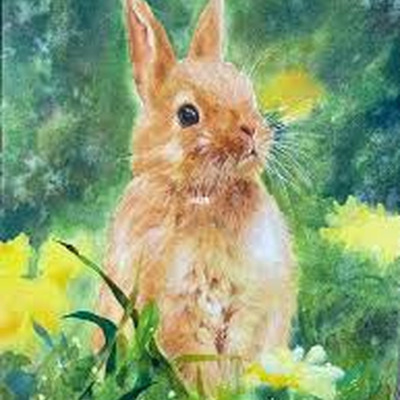}};
\node[anchor=north west,inner sep=0pt] at (0.09550,-0.03900) {\includegraphics[width=0.07100\textwidth]{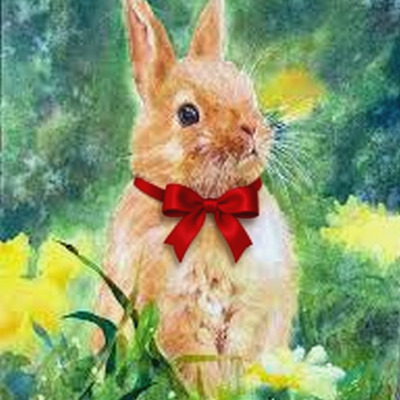}};
\node[anchor=north west,inner sep=0pt] at (0.16900,-0.03900) {\includegraphics[width=0.07100\textwidth]{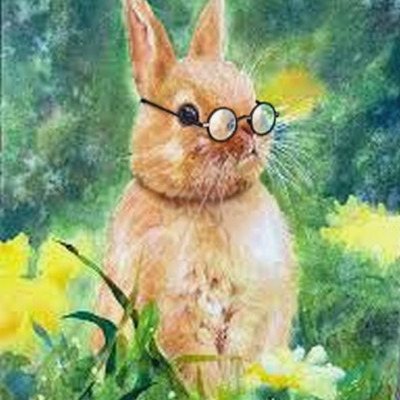}};
\node[anchor=north west,inner sep=0pt] at (0.24250,-0.03900) {\includegraphics[width=0.07100\textwidth]{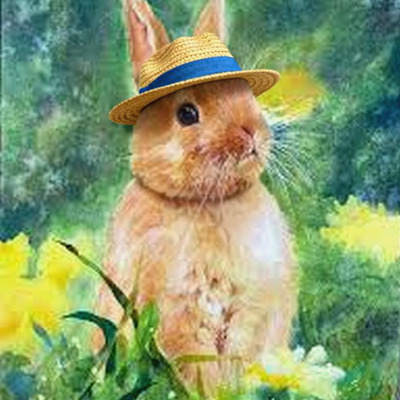}};
\node[anchor=north west,inner sep=0pt] at (0.31600,-0.03900) {\includegraphics[width=0.07100\textwidth]{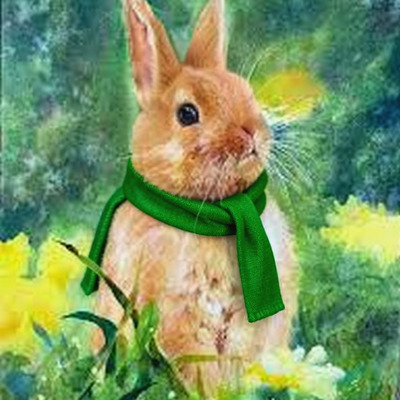}};
\node[anchor=north west,inner sep=0pt] at (0.38950,-0.03900) {\includegraphics[width=0.07100\textwidth]{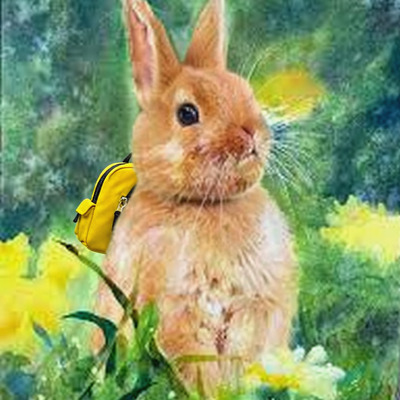}};
\node[anchor=north west,inner sep=0pt] at (0.46300,-0.03900) {\includegraphics[width=0.07100\textwidth]{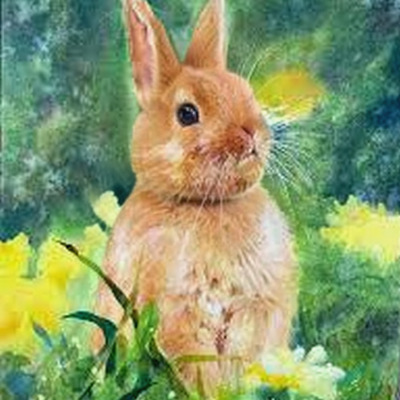}};
\node[anchor=north west,inner sep=0pt] at (0.02200,-0.11250) {\includegraphics[width=0.07100\textwidth]{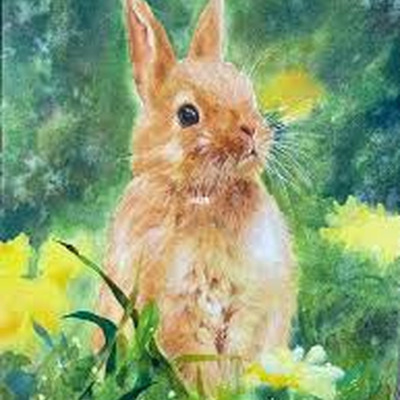}};
\node[anchor=north west,inner sep=0pt] at (0.09550,-0.11250) {\includegraphics[width=0.07100\textwidth]{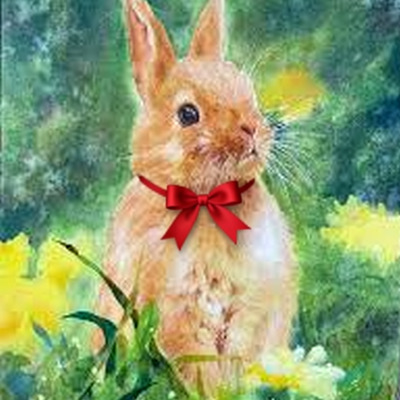}};
\node[anchor=north west,inner sep=0pt] at (0.16900,-0.11250) {\includegraphics[width=0.07100\textwidth]{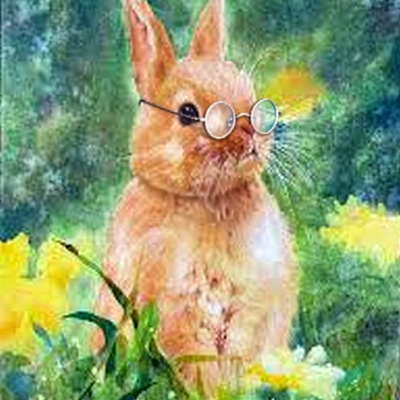}};
\node[anchor=north west,inner sep=0pt] at (0.24250,-0.11250) {\includegraphics[width=0.07100\textwidth]{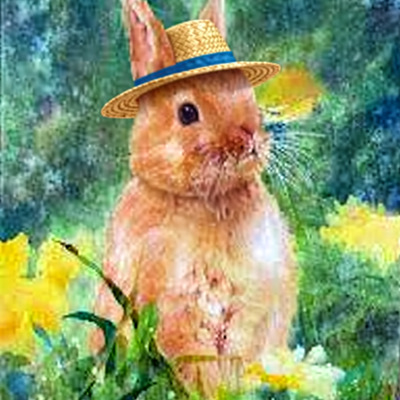}};
\node[anchor=north west,inner sep=0pt] at (0.31600,-0.11250) {\includegraphics[width=0.07100\textwidth]{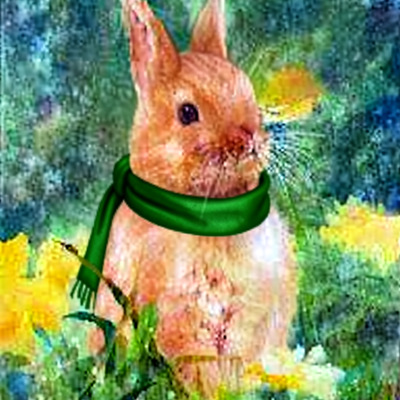}};
\node[anchor=north west,inner sep=0pt] at (0.38950,-0.11250) {\includegraphics[width=0.07100\textwidth]{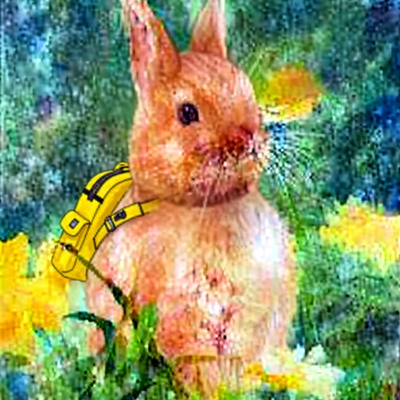}};
\node[anchor=north west,inner sep=0pt] at (0.46300,-0.11250) {\includegraphics[width=0.07100\textwidth]{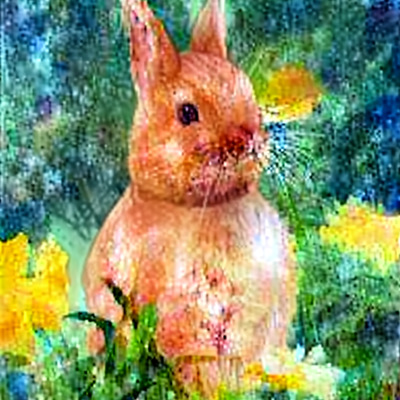}};
\node[anchor=north west,inner sep=0pt] at (0.55400,-0.03900) {\includegraphics[width=0.07100\textwidth]{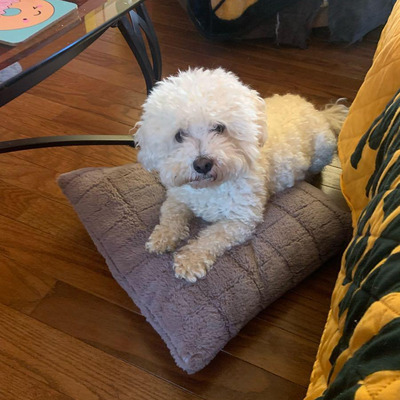}};
\node[anchor=north west,inner sep=0pt] at (0.62750,-0.03900) {\includegraphics[width=0.07100\textwidth]{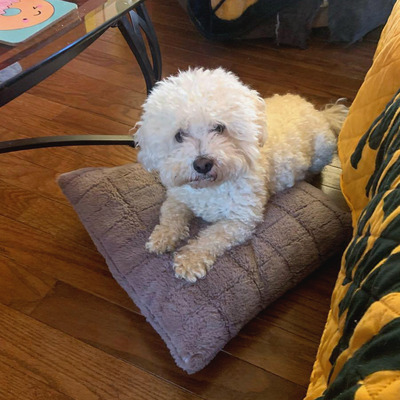}};
\node[anchor=north west,inner sep=0pt] at (0.70100,-0.03900) {\includegraphics[width=0.07100\textwidth]{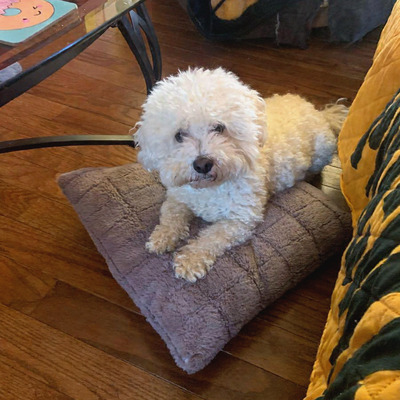}};
\node[anchor=north west,inner sep=0pt] at (0.55400,-0.11250) {\includegraphics[width=0.07100\textwidth]{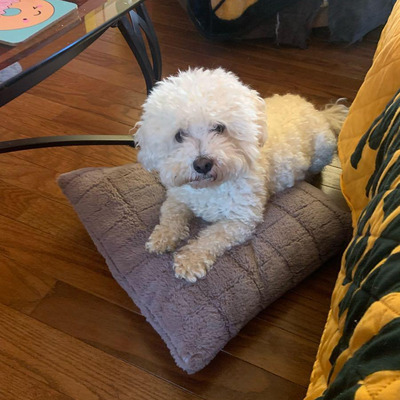}};
\node[anchor=north west,inner sep=0pt] at (0.62750,-0.11250) {\includegraphics[width=0.07100\textwidth]{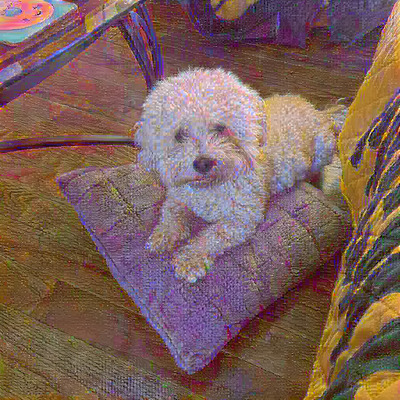}};
\node[anchor=north west,inner sep=0pt] at (0.70100,-0.11250) {\includegraphics[width=0.07100\textwidth]{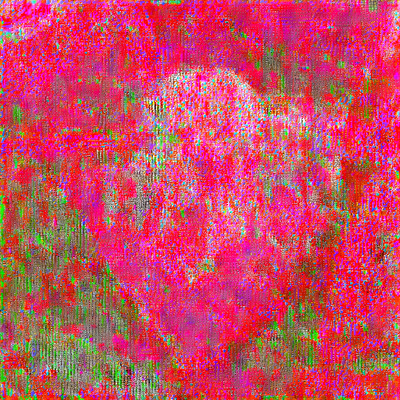}};
\node[anchor=north west,inner sep=0pt] at (0.78200,-0.03900) {\includegraphics[width=0.07100\textwidth]{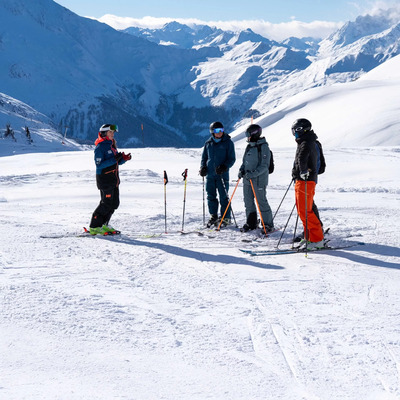}};
\node[anchor=north west,inner sep=0pt] at (0.85550,-0.03900) {\includegraphics[width=0.07100\textwidth]{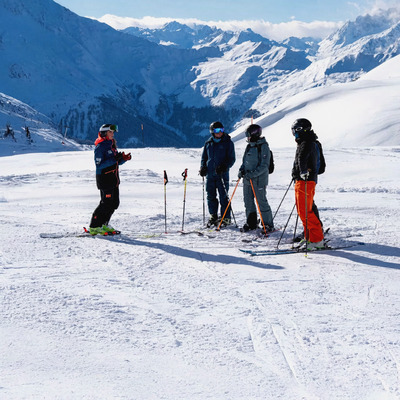}};
\node[anchor=north west,inner sep=0pt] at (0.92900,-0.03900) {\includegraphics[width=0.07100\textwidth]{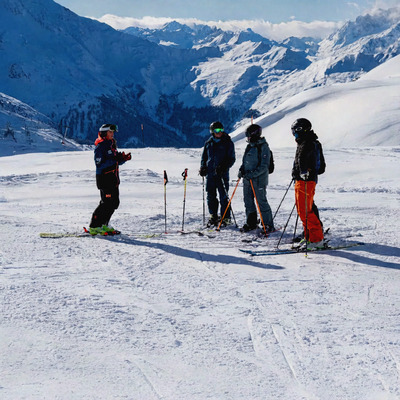}};
\node[anchor=north west,inner sep=0pt] at (0.78200,-0.11250) {\includegraphics[width=0.07100\textwidth]{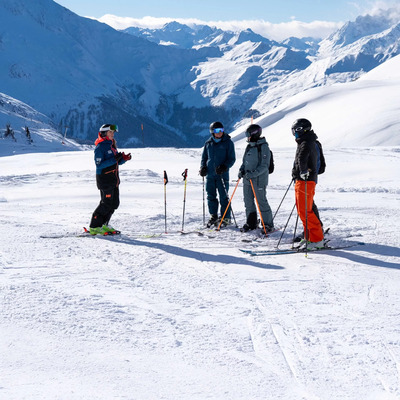}};
\node[anchor=north west,inner sep=0pt] at (0.85550,-0.11250) {\includegraphics[width=0.07100\textwidth]{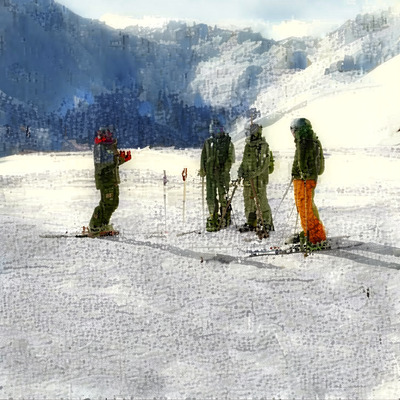}};
\node[anchor=north west,inner sep=0pt] at (0.92900,-0.11250) {\includegraphics[width=0.07100\textwidth]{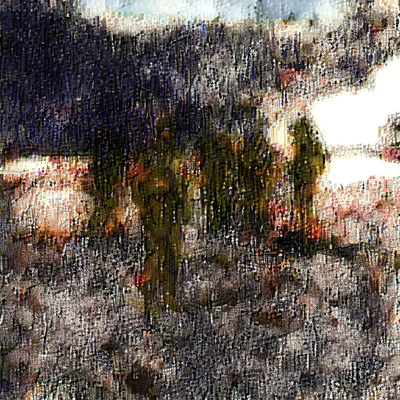}};
\node[rotate=90,anchor=center,inner sep=0pt] at (0.00900,-0.07450) {\bfseries\color{cvprblue}Ours};
\node[rotate=90,anchor=center,inner sep=0pt] at (0.00900,-0.14800) {\bfseries FLUX.2};
\draw[black!25,line width=0.4pt] (0.54400,-0.01100) -- (0.54400,-0.18350);
\draw[black!25,line width=0.4pt] (0.77700,-0.01100) -- (0.77700,-0.18350);
\end{tikzpicture}
\par
  \vspace{3pt}
  \includegraphics[width=\textwidth]{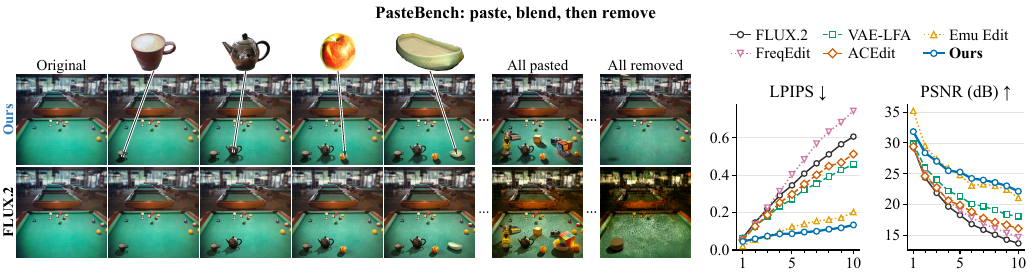}
  \par\vspace{-0.1cm}
  \captionof{figure}{\small \textbf{Stable repeated processing without training.}
  Top, left to right: one edit/undo chain (each edit and the final undo shown), encode--decode cycles, and no-op edits. Bottom left: PasteBench with ten sequential pastes and harmonization,
  then reverse-order removals (turns 5--9 and 11--19 omitted). Bottom right: final whole-image
  LPIPS~\cite{lpips} and PSNR versus the input after pasting and removing $K=1,\ldots,10$ objects ($2K$ turns),
  averaged over 20 scenes. Ours increasingly limits drift on longer chains;
  $K=10$ corresponds to Table~\ref{tab:pastebench}. Ours uses the \method codec for reconstruction
  and latent back-projection (BP) for editing; FLUX.2 uses the stock pipeline.}
  \label{fig:teaser}
  \endgroup
  \vspace{12pt}
}
\makeatother

\begin{document}
\maketitle

\begin{abstract}
Consistency is essential to generative workflows: repeated processing should preserve information beyond the changes explicitly requested. In latent diffusion models, this requires stability both when moving between latent and pixel spaces and when applying successive edits. Autoencoding errors and unintended changes from the editing model can accumulate across these operations, progressively altering image structure, texture, and color. We introduce \method, a framework that combines exact autoencoding consistency with calibrated editing stabilization, without training or fine-tuning. Its autoencoding component converts a pretrained variational autoencoder into a surjective pseudo-invertible neural network, with a fixed gain balancing deviations from the pretrained encoder and decoder. The modified decoder $D'$ is an explicit nonlinear right inverse of the encoder $E'$, satisfying $E'(D'(z))=z$ by construction. Consequently, $D'$ is a reflexive generalized inverse:
latent round trips preserve their input, up to numerical precision, and repeated image reconstruction is idempotent. To suppress editing drift, adaptive wavelet back-projection finds the closest latent to the editor's output in $\ell_2$, subject to matching the previous latent on wavelet coefficients whose changes fall below calibrated thresholds. This correction is applied once per turn, with no additional network evaluations after calibration. Together, these components support workflows that remain in latent space or interleave generation with pixel-space operations and constraints. Tests across models and editing benchmarks show less drift than stock pipelines and dedicated consistency methods.
\end{abstract}

\section{Introduction}
\label{sec:intro}
\vspace{-0.1cm}

{Generative image editing is commonly no longer a single shot. Users and automated pipelines refine an image over many turns, mixing instruction-based edits and pixel-space retouching (Figure~\ref{fig:teaser}). Most state-of-the-art editors are latent diffusion models~\cite{ldm}, which work in the latent space of a variational autoencoder (VAE). Whenever a turn needs pixels, to store, composite, or constrain the image~\cite{resample}, the image is decoded to pixels and later encoded back. We call this round trip a \emph{hand-off}. Such sessions are only useful if whatever was not explicitly changed stays exactly as it was.}

{In practice, images drift, and two errors compound over turns. The VAE is not invertible: re-encoding a decoded image does not recover its latent, so each hand-off loses detail and shifts colors, and tens of hand-offs corrupt an image even when nothing is edited~\cite{reed}. The editor adds drift of its own, altering content it was never asked to touch; even an instruction to keep the image as is gradually changes it~\cite{vaelfa}. Existing remedies address one part at a time. Some retrain a component for each model: REED-VAE fine-tunes the VAE decoder~\cite{reed}, ACEdit fine-tunes the editor on its own multi-turn outputs~\cite{acedit}, and manifold alignment trains for latent consistency~\cite{alignment}. Others correct a single frequency band: VAE-LFA aligns low frequencies~\cite{vaelfa}, and FreqEdit re-injects high frequencies~\cite{freqedit}. None makes round trips exact.}

We introduce \method, which makes autoencoding and editing consistent without training or fine-tuning. It works with existing editors as they are: all weights stay frozen, and only a one-time calibration is needed. Its codec turns the pretrained VAE into a surjective pseudo-invertible neural network (SPNN)~\cite{spnn}, whose encoder exactly undoes its decoder by construction: $\Ep(\Dp(z))=z$ in exact arithmetic, making $\Dp$ a reflexive generalized inverse and $\Dp\circ\Ep$ \textbf{idempotent}: once decoded, an image survives any number of further hand-offs, $\Dp\Ep\Dp\Ep(x) = \Dp\Ep(x)$.

Exact round trips cannot stop the editor's own drift, but intended edits tend to be strong and localized, whereas drift is small and widespread. Emu Edit exploits this distinction with fixed pixel thresholds~\cite{emuedit}. Our latent back-projection (BP) applies this distinction across frequency bands using the same generalized-inverse principle as our VAE: it projects the editor's output to the closest latent that preserves components judged unchanged. It runs once per edit without extra network calls. Neither component suffices alone: after fifty no-op edits, either leaves the image almost as degraded as the stock pipeline, while together they keep it nearly intact. Without training, \method also preserves unedited content better than fine-tuned REED-VAE and ACEdit, with a growing advantage on longer PasteBench chains (Figure~\ref{fig:teaser}, bottom right).

Our contributions are:
\begin{itemize}
    \item A training-free SPNN construction for pretrained autoencoders with exact latent round trips.
    \item Adaptive wavelet back-projection once per edit.
    \item PasteBench, testing recovery after sequential pixel-space pasting, harmonization, and removal.
\end{itemize}

\section{Preliminaries}
\label{sec:preliminaries}
\vspace{-0.1cm}

\subsection{Generalized inverses}
\label{sec:prelim-inverses}
\vspace{-0.1cm}

Dimensionality reduction generally prevents a mapping from having a two-sided inverse: several inputs can share the same output. A generalized inverse can instead select a representative input while preserving its output under the forward mapping. This distinction allows compression and exact consistency to coexist.

For a matrix $A\in\mathbb{R}^{m\times n}$, a generalized inverse $B\in\mathbb{R}^{n\times m}$ satisfies $ABA=A$. If it also satisfies $BAB=B$, it is called a reflexive generalized inverse. The Moore--Penrose pseudoinverse $A^\dagger$ is the unique choice of $B$ satisfying all four Penrose identities~\cite{penrose1955}:
\vspace{-0.1cm}
\begin{equation}
\label{eq:penrose}
\begin{aligned}
ABA &= A, & BAB &= B,\\
(AB)^\top &= AB, & (BA)^\top &= BA.
\end{aligned}
\end{equation}
The first two identities make $AB$ and $BA$ idempotent; the last two make these projections orthogonal. Together, they characterize the minimum-norm least-squares inverse. For full-row-rank $A$, any right inverse $B$ with $AB=I_m$ obeys the first two identities, but need not equal $A^\dagger$.

The first two identities extend directly to nonlinear maps through composition. For $g:\mathbb{R}^{n}\to\mathbb{R}^{m}$ and $h:\mathbb{R}^{m}\to\mathbb{R}^{n}$, they become $g\circ h\circ g=g$ and $h\circ g\circ h=h$. If $g$ is surjective, these hold precisely when $h$ is a right inverse: 
\vspace{-0.1cm}
\begin{equation}
\label{eq:right-inverse}
g(h(y))=y,\qquad y\in\mathbb{R}^{m}.
\end{equation}
The transpose conditions have no direct counterpart for general nonlinear maps. Selecting a particular nonlinear pseudoinverse therefore requires additional geometric structure. Here, we require a reflexive generalized right inverse, without imposing a minimum-norm criterion.

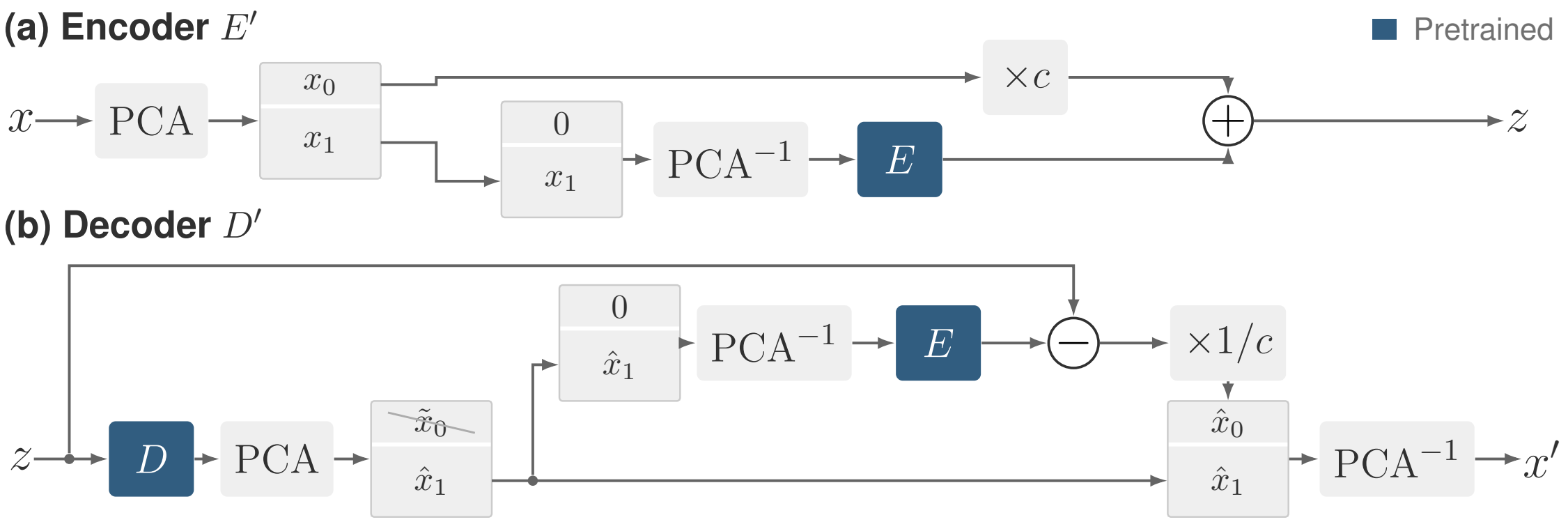
\begin{figure*}
 \centering
 \resizebox{0.9\textwidth}{!}{\definecolor{dvInk}{HTML}{303030}
\definecolor{dvMuted}{HTML}{707070}
\definecolor{dvCloud}{HTML}{EFEFEF}
\definecolor{dvPretrained}{HTML}{315D80}
\begin{tikzpicture}[x=1cm,y=1cm,
 font=\sffamily\fontsize{10}{12}\selectfont,
 >={Latex[length=1.8mm,width=1.25mm]},
 block/.style={rounded corners=2pt,draw=none,fill=dvCloud,text=dvInk,
  minimum width=.88cm,minimum height=.78cm,align=center,inner sep=4pt,
  font=\sffamily\fontsize{13}{15}\selectfont},
 vae/.style={block,fill=dvPretrained,text=white},
 coupling/.style={block},
 tensor/.style={draw=none,text=dvInk,inner sep=0pt},
 op/.style={circle,draw=dvInk,line width=.75pt,fill=white,
  minimum size=.5cm,inner sep=0pt,font=\sffamily\fontsize{15}{16}\selectfont},
 main/.style={->,draw=dvInk!75,line width=.8pt},
 fork/.style={circle,fill=dvInk!75,draw=none,minimum size=3pt,inner sep=0pt},
 label/.style={font=\sffamily\fontsize{9}{11}\selectfont,text=dvMuted}]
\path[use as bounding box] (0,1.12) rectangle (16.7,6.45);

\newcommand{\splitTensor}[5]{%
 \node[minimum width=1.25cm,minimum height=1.2cm,inner sep=0pt] (#1) at (#2,#3) {};
 \path[fill=dvCloud] (#2-.625,#3+.15) rectangle (#2+.625,#3+.6);
 \path[fill=dvCloud] (#2-.625,#3-.6) rectangle (#2+.625,#3+.15);
 \draw[white,line width=1.2pt] (#2-.625,#3+.15)--(#2+.625,#3+.15);
 \draw[dvInk!25,line width=.5pt,rounded corners=1pt]
  (#2-.625,#3-.6) rectangle (#2+.625,#3+.6);
 \node[tensor] (#1zero) at (#2,#3+.375) {$#4$};
 \node[tensor] (#1one) at (#2,#3-.225) {$#5$};
 \coordinate (#1zeroeast) at (#2+.625,#3+.375);
 \coordinate (#1oneeast) at (#2+.625,#3-.225);
 \coordinate (#1zerowest) at (#2-.625,#3+.375);
 \coordinate (#1onewest) at (#2-.625,#3-.225);
 \coordinate (#1zeronorth) at (#2,#3+.6);
}

\node[anchor=west,text=dvInk,font=\sffamily\bfseries\fontsize{11}{13}\selectfont]
 at (.1,6.2) {(a) Encoder $E'$};
\node[anchor=east,label] at (16.4,6.2)
 {\textcolor{dvPretrained}{\rule{7pt}{6pt}}\; Pretrained};

\node[tensor,font=\fontsize{14}{16}\selectfont] (x) at (.4,5.25) {$x$};
\node[block] (eq) at (1.75,5.25) {$\mathrm{PCA}$};
\splitTensor{xs}{3.5}{5.25}{x_0}{x_1}
\node[coupling] (gain) at (10.8,5.7) {$\times c$};
\splitTensor{ezero}{6}{4.85}{0}{x_1}
\node[block] (einversepca) at (7.75,4.85) {$\mathrm{PCA}^{-1}$};
\node[vae] (e) at (9.5,4.85) {$E$};
\node[op] (sum) at (12.9,5.25) {$+$};
\node[tensor,font=\fontsize{14}{16}\selectfont] (z) at (15.9,5.25) {$z$};
\draw[main] (x.east)--(eq.west);
\draw[main] (eq.east)--(xs.west);
\draw[main] (xszeroeast)--(4.7,5.625)--(4.7,5.7)--(gain.west);
\draw[main] (gain.east)--(12.9,5.7)--(sum.north);
\draw[main] (xsoneeast)--(4.7,5.025)--(4.7,4.625)--(ezeroonewest);
\draw[main] (ezero.east)--(einversepca.west);
\draw[main] (einversepca.east)--(e.west);
\draw[main] (e.east)--(12.9,4.85)--(sum.south);
\draw[main] (sum.east)--(z.west);

\node[anchor=west,text=dvInk,font=\sffamily\bfseries\fontsize{11}{13}\selectfont]
 at (.1,4.15) {(b) Decoder $D'$};
\node[tensor,font=\fontsize{14}{16}\selectfont] (dz) at (.4,1.75) {$z$};
\node[vae] (d) at (1.75,1.75) {$D$};
\node[block] (dq) at (3.05,1.75) {$\mathrm{PCA}$};
\splitTensor{ds}{4.65}{1.75}{\tilde{x}_0}{\hat{x}_1}
\draw[dvInk!40,line width=.6pt] (4.2,2.23)--(5.1,2.02);
\splitTensor{dzero}{6.6}{2.95}{0}{\hat{x}_1}
\node[block] (dinversepca) at (8.2,2.95) {$\mathrm{PCA}^{-1}$};
\node[vae] (de) at (9.9,2.95) {$E$};
\node[op] (minus) at (11.3,2.95) {$-$};
\node[coupling] (inversegain) at (12.9,2.95) {$\times 1/c$};
\splitTensor{out}{12.9}{1.75}{\hat{x}_0}{\hat{x}_1}
\node[block] (inverseq) at (14.65,1.75) {$\mathrm{PCA}^{-1}$};
\node[tensor,font=\fontsize{14}{16}\selectfont] (dx) at (16.15,1.75) {$x'$};
\node[fork] (zfork) at (.9,1.75) {};
\node[fork] (contentfork) at (5.7,1.525) {};
\draw[main] (dz.east)--(d.west);
\draw[main] (d.east)--(dq.west);
\draw[main] (dq.east)--(ds.west);
\draw[main] (dsoneeast)--(outonewest);
\draw[main] (contentfork.north)--(5.7,2.725)--(dzeroonewest);
\draw[main] (dzero.east)--(dinversepca.west);
\draw[main] (dinversepca.east)--(de.west);
\draw[main] (de.east)--(minus.west);
\draw[main] (zfork.north)--(.9,3.75)--(11.3,3.75)--(minus.north);
\draw[main] (minus.east)--(inversegain.west);
\draw[main] (inversegain.south)--(outzeronorth);
\draw[main] (out.east)--(inverseq.west);
\draw[main] (inverseq.east)--(dx.west);

\end{tikzpicture}}
 \par\vspace{-0.1cm}
 \caption{\small \textbf{The \method codec.} PCA separates reserved and retained coordinates. Zero-filling and inverse PCA recover the retained image. Decoding keeps the retained coordinates $\hat x_1$ of $D(z)$ and replaces its reserved coordinates $\tilde x_0$ with $\hat x_0$, computed with the frozen encoder $E$ (Eq.~\ref{eq:decoder}). Blue blocks are pretrained; $\times c$ and $\times 1/c$ multiply by a fixed gain $c$. The DoF is chosen to be the frozen $D$. Any other mapping satisfies right inverseness $E'(D'(z))=z$. $D$ provides a prior of similarity to original input and closeness to the data distribution.
 }
\label{fig:method}
\end{figure*}

\subsection{Surjective pseudo-invertible neural networks}
\label{sec:prelim-spnn}
\vspace{-0.1cm}

Surjective pseudo-invertible neural networks (SPNNs)~\cite{spnn} are architectures with output size smaller than input size, that admit a tractable non-linear right-inverse by construction and satisfy the corresponding geometric properties. A block partitions its input after an orthogonal transform $U$:
\vspace{-0.1cm}
\begin{equation}
\label{eq:spnn-forward}
\begin{aligned}
(x_0,x_1)&=Ux,\\
g(x)&=x_0\odot s(x_1)+t(x_1),
\end{aligned}
\end{equation}
where $x_0\in\mathbb{R}^{m}$ will determine the output size, $x_1\in\mathbb{R}^{n-m}$, $s,t:\mathbb{R}^{n-m}\to\mathbb{R}^{m}$, and $s$ is strictly positive. For any output $y$ and any choice of $x_1$, the coupling can be solved uniquely for $x_0$, making $g$ surjective. The degree of freedom comes from a map $r:\mathbb{R}^{m}\to\mathbb{R}^{n-m}$ that selects $x_1$, yielding the reverse:
\vspace{-0.1cm}
\begin{equation}
\label{eq:spnn-reverse}
\begin{aligned}
\hat{x}_1&=r(y),\\
\hat{x}_0&=\bigl(y-t(\hat{x}_1)\bigr)\oslash s(\hat{x}_1),\\
h(y)&=U^\top(\hat{x}_0,\hat{x}_1),
\end{aligned}
\end{equation}
Here $\odot$ and $\oslash$ denote elementwise multiplication and division. Substitution gives $g(h(y))=y$ for any $r$. The original SPNN learns its functions, but this identity follows from its architecture. We use this right-inverse property; SPNNs also define a particular nonlinear pseudoinverse.

For an autoencoder, taking $g=E'$ and $h=D'$ gives exact latent round trips and an idempotent image reconstruction map $P=D'\circ E'$:
\vspace{-0.1cm}
\begin{equation}
\label{eq:autoencoder-consistency}
E'\circ D'=\mathrm{Id},\qquad P\circ P=P.
\end{equation}
Thus, reconstruction may change an input image once, but subsequent reconstructions leave the result unchanged. These identities concern consistency; they do not require lossless image reconstruction. \method instantiates this structure using the pretrained encoder and decoder, without learning new functions or fine-tuning their parameters.

\subsection{Back-projection}
\label{sec:prelim-back-projection}
\vspace{-0.1cm}

For a linear measurement operator $A$ and a consistent target $y\in\operatorname{range}(A)$, back-projection finds the closest point to $z$ with $Au=y$:
\vspace{-0.1cm}
\begin{equation}
\label{eq:back-projection}
\begin{aligned}
z^+ &= \arg\min_{u:\,Au=y}\|u-z\|_2^2 \\
    &= z+A^\dagger(y-Az).
\end{aligned}
\end{equation}
Here $A^\dagger$ is the Moore--Penrose pseudoinverse. For $A=SW$, where $W$ is an orthonormal wavelet transform and $S$ selects coefficients, $A^\dagger=W^\top S^\top$. Back-projection replaces the selected coefficients and leaves the others unchanged. With $y=Az_{\mathrm{ref}}$, it preserves these reference coefficients with minimum $\ell_2$ change.

\section{Method}
\label{sec:method}
\vspace{-0.1cm}

\subsection{Consistent autoencoding}
\label{sec:codec}
\vspace{-0.1cm}
\vspace{-0.1cm}\paragraph{Coordinates.}
Let $\mu$ be a calibration mean and $U=[U_0;U_1]$ an orthogonal matrix, with $U_0\in\R^{k\times n}$. As in Eq.~\ref{eq:spnn-forward} and Figure~\ref{fig:method}, an image $x$ has reserved coordinates $x_0$ and retained coordinates $x_1$:
\vspace{-0.1cm}
\begin{equation}
 x_0=U_0(x-\mu),\label{eq:y0} \quad\quad
 x_1=U_1(x-\mu).
\end{equation}
{Setting the reserved coordinates to zero gives the retained image $\mu+U_1^\top x_1$.} PCA supplies low-variance reserved directions (Appendix~\ref{app:basis}).

\vspace{-0.1cm}\paragraph{Encoder.}
For a fixed gain $c>0$, {the encoder is the SPNN block of Eq.~\ref{eq:spnn-forward} with $s\equiv c$ and $t(x_1)=E(\mu+U_1^\top x_1)$, the frozen encoder applied to the retained image:}
\vspace{-0.1cm}
\begin{equation}
 \boxed{\Ep(x)=c\,x_0+t(x_1).}\label{eq:encoder}
\end{equation}
The frozen encoder processes retained content; reserved coordinates{, scaled by $c$,} enter additively (Figure~\ref{fig:method}).

\vspace{-0.1cm}\paragraph{Decoder.}
The decoder keeps the retained content of $D(z)$ and replaces its reserved coordinates{, as in Eq.~\ref{eq:spnn-reverse} with $r(z)=U_1(D(z)-\mu)$}:
\vspace{-0.1cm}
\begin{equation}
 \boxed{\begin{aligned}
 \hat x_1&=U_1(D(z)-\mu),\qquad \hat x_0=\frac{z-t(\hat x_1)}{c},\\
 \Dp(z)&=\mu+U_0^\top\hat x_0+U_1^\top\hat x_1.
 \end{aligned}}\label{eq:decoder}
\end{equation}

\vspace{-0.1cm}\paragraph{Consistency.}
{Because $U$ is orthogonal, the coordinates of $\Dp(z)$ are exactly $(\hat x_0,\hat x_1)$, so Eq.~\ref{eq:encoder} gives $\Ep(\Dp(z))=c\,\hat x_0+t(\hat x_1)=z$. Hence}
\vspace{-0.1cm}
\begin{equation}
 \Ep(\Dp(z))=z,\qquad (\Dp\Ep)^2=\Dp\Ep.
 \label{eq:guarantee}
\end{equation}
The second equality preserves the \emph{first reconstruction} under further unchanged cycles, even if it differs from the input.

\vspace{-0.1cm}\paragraph{Deviation from the pretrained VAE.}
For the construction above, with $\tilde x_0=U_0(D(z)-\mu)$, the deviations are
\vspace{-0.1cm}
\begin{equation}
\begin{aligned}
 \|\Ep(x)-E(x)\|_2
 &=\big\|c\,x_0+t(x_1)-E(x)\big\|_2,\\
 \|\Dp(z)-D(z)\|_2
 &=\big\|\tfrac{1}{c}\big[z-t(\hat x_1)\big]-\tilde x_0\big\|_2.
\end{aligned}
\end{equation}
The gain $c$ trades off closeness to $E$ and $D$ (values: Table~\ref{tab:codecs}).
At our settings, relative encoder deviations are 1.3--5.9\%, and decoder outputs match the pretrained decoder at 42--46\,dB PSNR on real and generated latents, see Appendix~\ref{app:deviation}.

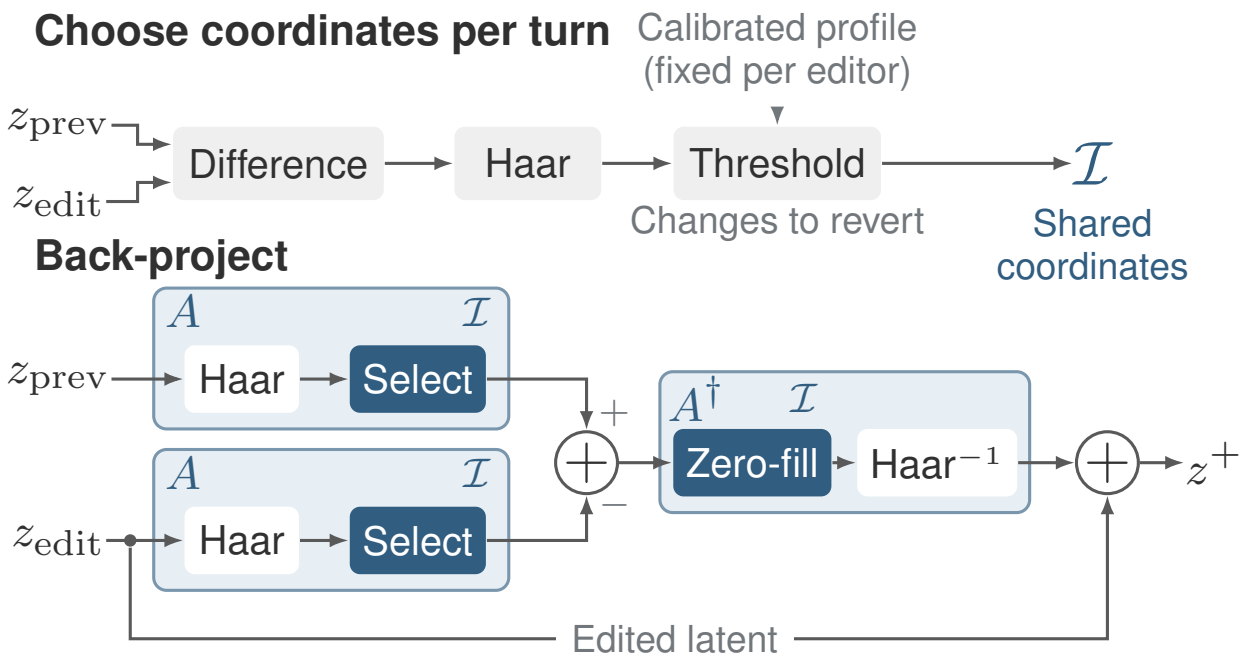
\begin{figure}
\centering
\definecolor{bpInk}{HTML}{303030}
\definecolor{bpMuted}{HTML}{72777C}
\definecolor{bpGray}{HTML}{EFEFEF}
\definecolor{bpBlue}{HTML}{315D80}
\definecolor{bpPale}{HTML}{E8EFF4}
\resizebox{\columnwidth}{!}{%
\begin{tikzpicture}[x=1cm,y=1cm,
  font=\sffamily\fontsize{7.7}{9}\selectfont,text=bpInk,
  >={Latex[length=1.3mm,width=.95mm]},
  flow/.style={->,draw=bpInk!80,line width=.6pt},
  control/.style={->,draw=bpMuted,line width=.55pt,densely dashed},
  block/.style={rounded corners=2pt,fill=bpGray,minimum height=.5cm,
    minimum width=1.0cm,align=center,inner sep=3pt},
  inner/.style={block,fill=white,minimum width=.78cm,minimum height=.46cm,
    inner sep=2.5pt},
  selection/.style={inner,fill=bpBlue,text=white},
  dot/.style={circle,fill=bpInk!80,inner sep=0pt,minimum size=2.4pt},
  op/.style={circle,draw=bpInk!80,line width=.6pt,fill=white,
    minimum size=.39cm,inner sep=0pt,font=\fontsize{12}{13}\selectfont},
  lab/.style={text=bpMuted,font=\sffamily\fontsize{7.2}{8.3}\selectfont},
  heading/.style={anchor=west,font=\sffamily\bfseries\fontsize{8.2}{9.5}\selectfont},
  latent/.style={inner sep=0pt,font=\fontsize{9.5}{11}\selectfont}]
\path[use as bounding box] (0,.02) rectangle (8.5,4.5);

\node[heading] at (.06,4.28) {Choose coordinates per turn};
\node[latent] (prevtop) at (.35,3.66) {$z_{\mathrm{prev}}$};
\node[latent] (edittop) at (.35,3.14) {$z_{\mathrm{edit}}$};
\node[block] (diff) at (1.85,3.40) {Difference};
\node[block] (wavelet) at (3.55,3.40) {Haar};
\node[block] (threshold) at (5.25,3.40) {Threshold};
\node[lab,align=center] (tau) at (5.25,4.16) {Calibrated profile\\(fixed per editor)};
\node[latent,text=bpBlue,font=\fontsize{12}{13}\selectfont]
  (indices) at (7.4,3.40) {$\mathcal I$};
\node[lab,text=bpBlue,align=center,anchor=north,inner ysep=0pt]
  at (7.4,3.10) {Shared\\coordinates};
\node[lab] at (5.25,2.99) {Changes to revert};
\draw[flow] (prevtop.east)--(.89,3.66)--(.89,3.53)--($(diff.west)+(0,.13)$);
\draw[flow] (edittop.east)--(.89,3.14)--(.89,3.27)--($(diff.west)+(0,-.13)$);
\draw[flow] (diff.east)--(wavelet.west);
\draw[flow] (wavelet.east)--(threshold.west);
\draw[flow] (threshold.east)--(indices.west);
\draw[control] (tau.south)--(threshold.north);

\node[heading] at (.06,2.75) {Back-project};
\foreach \yy/\suffix in {1.95/ref,.85/edit}{
  \path[rounded corners=2.5pt,fill=bpPale,draw=bpBlue!65,line width=.6pt]
    (1.0,\yy-.36) rectangle (3.42,\yy+.60);
  \node[anchor=center,text=bpBlue,inner sep=0pt,
    font=\fontsize{9.5}{11}\selectfont] at (1.2,\yy+.44) {$A$};
  \node[inner] (haar\suffix) at (1.6,\yy-.02) {Haar};
  \node[selection] (select\suffix) at (2.8,\yy-.02) {Select};
  \node[anchor=center,text=bpBlue,inner sep=0pt,
    font=\fontsize{8}{9}\selectfont] at (3.20,\yy+.44) {$\mathcal I$};
  \draw[flow] (haar\suffix.east)--(select\suffix.west);
}
\node[latent] (prev) at (.35,1.93) {$z_{\mathrm{prev}}$};
\node[latent] (edit) at (.35,.83) {$z_{\mathrm{edit}}$};
\node[op] (subtract) at (3.95,1.36) {$+$};

\path[rounded corners=2.5pt,fill=bpPale,draw=bpBlue!65,line width=.6pt]
  (4.45,1.02) rectangle (6.97,1.98);
\node[anchor=center,text=bpBlue,inner sep=0pt,
  font=\fontsize{9.5}{11}\selectfont] at (4.70,1.80) {$A^\dagger$};
\node[selection,minimum width=.95cm] (scatter) at (5.08,1.36) {Zero-fill};
\node[anchor=center,text=bpBlue,inner sep=0pt,
  font=\fontsize{8}{9}\selectfont] at (5.43,1.80) {$\mathcal I$};
\node[inner,minimum width=.95cm] (ihaar) at (6.34,1.36) {Haar$^{-1}$};
\draw[flow] (scatter.east)--(ihaar.west);

\node[op] (add) at (7.5,1.36) {$+$};
\node[latent] (output) at (8.23,1.36) {$z^+$};
\draw[flow] (prev.east)--(haarref.west);
\draw[flow] (edit.east)--(haaredit.west);
\draw[flow] (selectref.east)--(3.95,1.93)--(subtract.north);
\draw[flow] (selectedit.east)--(3.95,.83)--(subtract.south);
\node[lab,anchor=west,inner sep=1pt] at (3.99,1.70) {$+$};
\node[lab,anchor=west,inner sep=1pt] at (3.99,1.07) {$-$};
\draw[flow] (subtract.east)--(scatter.west);
\draw[flow] (ihaar.east)--(add.west);
\draw[flow] (add.east)--(output.west);
\node[dot] (skip) at (.84,.83) {};
\draw[flow] (skip.south)--(.84,.16)--(7.5,.16)--(add.south);
\node[lab,fill=white,inner xsep=3pt,inner ysep=1pt] at (4.55,.16) {Edited latent};
\end{tikzpicture}
}
\par\vspace{-0.1cm}
\caption{\small \textbf{Adaptive latent back-projection (BP).} Per-turn $\mathrm{HH}_1$ noise estimates and a band profile calibrated once per editor set thresholds selecting $\mathcal I$. Shared selection in $A$ and $A^\dagger$ reverts these coordinates to their input values; all others remain edited.}
\label{fig:bp}
\end{figure}

\subsection{Latent back-projection}
\label{sec:bp}
\vspace{-0.1cm}

An exact codec cannot stop an editor from changing content it was asked to preserve. Such changes accumulate even when turns stay in latent space, shifting colors and erasing detail~\cite{vaelfa,acedit,freqedit}. We exploit a simple distinction: local edits tend to produce strong, concentrated changes, while drift is weaker and widespread. Latent back-projection (BP) uses this cue to suppress drift selectively across all frequency bands, allowing substantial edits to pass through.

\vspace{-0.1cm}\paragraph{Correction.}
Let $z_{\mathrm{prev}}$ be the latent of the turn's input and $z_{\mathrm{edit}}$ the editor's output, both in raw VAE coordinates. In a latent route, $z_{\mathrm{prev}}$ is the previous turn's output; in a pixel route, it is the encoding of the turn's input image.
Let $d=z_{\mathrm{edit}}-z_{\mathrm{prev}}$ and let $W$ be a two-level orthonormal Haar transform of each channel $\ell$, with bands $b\in\{\mathrm{LL}_2,\mathrm{LH}_2,\mathrm{HL}_2,\mathrm{HH}_2,\mathrm{LH}_1,\mathrm{HL}_1,\mathrm{HH}_1\}$. The thresholds define a measurement operator $A=SW$, where $S$ extracts the coefficients of $Wd$ with magnitude at most $\tau_{b,\ell}$. With this selection fixed for the current correction, $A$ is surjective onto the selected-coordinate space, with $A^\dagger=W^\top S^\top$ and $AA^\dagger=I$. Back-projection (Section~\ref{sec:prelim-back-projection}, Figure~\ref{fig:bp}) gives the closest latent to $z_{\mathrm{edit}}$ in $\ell_2$ satisfying $Az^+=Az_{\mathrm{prev}}$:
\vspace{-0.1cm}
\begin{equation}
\label{eq:bp}
\begin{aligned}
z^+ &= z_{\mathrm{edit}}+A^\dagger(Az_{\mathrm{prev}}-Az_{\mathrm{edit}})\\
    &= z_{\mathrm{prev}}+W^\top(I-S^\top S)\,Wd.
\end{aligned}
\end{equation}
BP thus hard-thresholds the wavelet coefficients of $d$~\cite{donoho1994wavelet}: selected coefficients return to their input values, and the rest stay as edited. It acts once per turn on the final latent, needs no user mask, training, or change to the editor or sampler, and adds no denoiser call.

\vspace{-0.1cm}\paragraph{Thresholds.}
Thresholds combine a per-turn noise estimate with a band profile $\rho_{b,\ell}$ in a bandwise form of the universal threshold of Donoho and Johnstone~\cite{donoho1994wavelet}:
\vspace{-0.1cm}
\begin{equation}
\label{eq:bp-threshold}
\tau_{b,\ell}=\rho_{b,\ell}\,\hat\sigma_\ell\sqrt{2\ln N_b},\quad
\hat\sigma_\ell=\frac{\operatorname{median}\lvert\mathrm{HH}_{1,\ell}(d)\rvert}{0.6745},
\end{equation}
where $N_b$ is the number of coefficients per channel in band $b$. The robust finest-scale rule~\cite{donoho1994wavelet,chang2000adaptive} adapts thresholds to each image and turn; its use as a drift estimate assumes sparse edits in each channel's $\mathrm{HH}_1$ band. Drift is colored, however: for example, on FLUX.2 [klein], its calibrated level in $\mathrm{LL}_2$ averages about three times that in $\mathrm{HH}_1$. Level-dependent thresholds address correlated noise~\cite{johnstone1997correlated}, but estimating each level from $d$ would also measure the edit. We therefore measure only $\mathrm{HH}_1$ on the current turn and take the spectral shape, assumed more stable than the level, from calibration: $\rho_{b,\ell}=\sigma^{\mathrm{cal}}_{b,\ell}/\sigma^{\mathrm{cal}}_{\mathrm{HH}_1,\ell}$. The factor $\sqrt{2\ln N_b}$ is a heuristic here, since editor drift need not be i.i.d. Gaussian.

\vspace{-0.1cm}\paragraph{Calibration.}
We compute $\sigma^{\mathrm{cal}}_{b,\ell}$ only once per editor, from no-op chains with BP disabled, pooling absolute coefficients over images, turns, and positions into one median per band and channel, divided by 0.6745. For FLUX.2 [klein] and SD3-UE (UltraEdit~\cite{ultraedit} on Stable Diffusion~3~\cite{sd3}), the chains run ten ``Keep the image unchanged.'' turns in the latent route on eight external images, and we use turns 2--10 (all editors: Table~\ref{tab:bpcalib}). In both editors' VAE-LFA no-op and edit/undo protocols (Section~\ref{sec:vaelfa}), with wrapped pixel or latent routes, Eq.~\ref{eq:bp-threshold} lowers final LPIPS by 0.015--0.063 relative to either $\rho_{b,\ell}=1$ or replacing $\rho_{b,\ell}\hat\sigma_\ell$ with $\sigma^{\mathrm{cal}}_{b,\ell}$; LPIPS and L1 improve significantly in all 16 (95\% paired bootstrap; Tables~\ref{tab:vaelfa-bp-rules} and~\ref{tab:vaelfa-bp-ablation}).

\begin{figure}
\centering
\includegraphics[width=0.8\columnwidth]{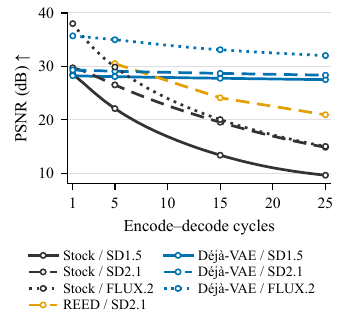}
\par\vspace{-0.1cm}
\caption{\small \textbf{The \method codec limits reconstruction drift.}
PSNR (dB) over 179 images, 8-bit hand-offs.
Stock and \method: all 25 cycles. REED: cycles 5, 15, 25.
Cycle-25 LPIPS: Table~\ref{tab:cycles}.}
\label{fig:cycles}
\end{figure}

\begin{table}
\centering
\footnotesize
\setlength{\tabcolsep}{2pt}
\caption{\small \textbf{REED editing benchmark.}
179 examples; alternating edits and reversals (SD Inpainting: repeated inpainting), 8-bit hand-offs; rounds 5, 15, 25 versus the first edited image (round 1).
Below each Ours row: the mean paired difference Ours$-$Stock $\pm$ the half-width of its 95\% bootstrap CI over the 179 matched chains (same images, instructions, and seeds).
Bold/underlined: best/second per editor and column.
Metrics and controls: Table~\ref{tab:reed-editing-full}.}
\label{tab:reed-editing}
\par\vspace{-0.1cm}
\resizebox{0.9\columnwidth}{!}{%
\begin{tabular}{@{}clrrrrrr@{}}
\toprule
& & \multicolumn{3}{c}{PSNR$\uparrow$} & \multicolumn{3}{c}{LPIPS$\downarrow$} \\
\cmidrule(lr){3-5}\cmidrule(l){6-8}
& Method & 5 & 15 & 25 & 5 & 15 & 25 \\
\midrule
\multirow{4}{*}{\rotatebox{90}{\footnotesize\emph{\shortstack{Instruct\\Pix2Pix}}}} & Stock & 17.60 & 10.15 & 8.56 & 0.341 & 0.699 & 0.765 \\
 & REED-VAE & 19.81 & 13.36 & \underline{11.36} & 0.180 & \underline{0.450} & \underline{0.580} \\
 & Ours w/o BP & \underline{20.65} & \underline{13.65} & 11.33 & \underline{0.172} & 0.458 & 0.608 \\
 & \textbf{Ours} & \textbf{22.78} & \textbf{18.16} & \textbf{16.31} & \textbf{0.148} & \textbf{0.252} & \textbf{0.324} \\
 & {\scriptsize\shortstack[l]{{Ours$-$Stock}\\{95\% CI}}} & {\scriptsize\shortstack[r]{{$+$5.18}\\{$\pm$0.87}}} & {\scriptsize\shortstack[r]{{$+$8.00}\\{$\pm$0.80}}} & {\scriptsize\shortstack[r]{{$+$7.75}\\{$\pm$0.75}}} & {\scriptsize\shortstack[r]{{$-$0.193}\\{$\pm$0.022}}} & {\scriptsize\shortstack[r]{{$-$0.447}\\{$\pm$0.026}}} & {\scriptsize\shortstack[r]{{$-$0.441}\\{$\pm$0.028}}} \\
\midrule
\multirow{4}{*}{\rotatebox{90}{\footnotesize\emph{MagicBrush}}} & Stock & 19.08 & 11.37 & 8.80 & 0.308 & 0.705 & 0.790 \\
 & REED-VAE & 21.53 & 16.66 & \underline{14.09} & 0.190 & 0.510 & 0.690 \\
 & Ours w/o BP & \underline{22.32} & \underline{16.80} & 13.60 & \underline{0.147} & \underline{0.437} & \underline{0.615} \\
 & \textbf{Ours} & \textbf{23.14} & \textbf{20.71} & \textbf{19.53} & \textbf{0.113} & \textbf{0.175} & \textbf{0.220} \\
 & {\scriptsize\shortstack[l]{{Ours$-$Stock}\\{95\% CI}}} & {\scriptsize\shortstack[r]{{$+$4.06}\\{$\pm$0.65}}} & {\scriptsize\shortstack[r]{{$+$9.35}\\{$\pm$0.71}}} & {\scriptsize\shortstack[r]{{$+$10.73}\\{$\pm$0.66}}} & {\scriptsize\shortstack[r]{{$-$0.196}\\{$\pm$0.015}}} & {\scriptsize\shortstack[r]{{$-$0.530}\\{$\pm$0.023}}} & {\scriptsize\shortstack[r]{{$-$0.570}\\{$\pm$0.024}}} \\
\midrule
\multirow{4}{*}{\rotatebox{90}{\footnotesize\emph{\shortstack{DiffEdit\\(SD 2.1)}}}} & Stock & 15.53 & 12.35 & 10.94 & 0.362 & 0.639 & 0.735 \\
 & REED-VAE & \textbf{16.24} & \textbf{12.91} & 11.44 & 0.300 & 0.550 & 0.680 \\
 & Ours w/o BP & 15.80 & 12.50 & \underline{11.63} & \underline{0.294} & \underline{0.473} & \underline{0.551} \\
 & \textbf{Ours} & \underline{15.87} & \underline{12.57} & \textbf{11.69} & \textbf{0.289} & \textbf{0.444} & \textbf{0.502} \\
 & {\scriptsize\shortstack[l]{{Ours$-$Stock}\\{95\% CI}}} & {\scriptsize\shortstack[r]{{$+$0.34}\\{$\pm$0.15}}} & {\scriptsize\shortstack[r]{{$+$0.23}\\{$\pm$0.15}}} & {\scriptsize\shortstack[r]{{$+$0.75}\\{$\pm$0.13}}} & {\scriptsize\shortstack[r]{{$-$0.073}\\{$\pm$0.007}}} & {\scriptsize\shortstack[r]{{$-$0.195}\\{$\pm$0.010}}} & {\scriptsize\shortstack[r]{{$-$0.233}\\{$\pm$0.011}}} \\
\midrule
\multirow{4}{*}{\rotatebox{90}{\footnotesize\emph{\shortstack{Paint-by-\\Example}}}} & Stock & 18.72 & 13.91 & 11.83 & 0.254 & 0.596 & 0.708 \\
 & REED-VAE & 19.62 & 16.19 & 14.43 & 0.200 & 0.440 & 0.590 \\
 & Ours w/o BP & \underline{20.57} & \underline{18.01} & \underline{16.80} & \underline{0.149} & \underline{0.273} & \underline{0.378} \\
 & \textbf{Ours} & \textbf{20.90} & \textbf{18.95} & \textbf{18.13} & \textbf{0.129} & \textbf{0.187} & \textbf{0.228} \\
 & {\scriptsize\shortstack[l]{{Ours$-$Stock}\\{95\% CI}}} & {\scriptsize\shortstack[r]{{$+$2.18}\\{$\pm$0.33}}} & {\scriptsize\shortstack[r]{{$+$5.04}\\{$\pm$0.43}}} & {\scriptsize\shortstack[r]{{$+$6.30}\\{$\pm$0.42}}} & {\scriptsize\shortstack[r]{{$-$0.125}\\{$\pm$0.007}}} & {\scriptsize\shortstack[r]{{$-$0.409}\\{$\pm$0.016}}} & {\scriptsize\shortstack[r]{{$-$0.480}\\{$\pm$0.017}}} \\
\midrule
\multirow{4}{*}{\rotatebox{90}{\footnotesize\emph{\shortstack{SD\\Inpainting}}}} & Stock & 20.78 & 12.15 & 9.53 & 0.284 & 0.698 & 0.793 \\
 & REED-VAE & \underline{23.14} & 16.66 & 13.61 & 0.170 & 0.470 & 0.650 \\
 & Ours w/o BP & 23.06 & \underline{17.17} & \underline{14.67} & \underline{0.143} & \underline{0.403} & \underline{0.534} \\
 & \textbf{Ours} & \textbf{24.21} & \textbf{20.00} & \textbf{18.35} & \textbf{0.095} & \textbf{0.168} & \textbf{0.215} \\
 & {\scriptsize\shortstack[l]{{Ours$-$Stock}\\{95\% CI}}} & {\scriptsize\shortstack[r]{{$+$3.43}\\{$\pm$0.56}}} & {\scriptsize\shortstack[r]{{$+$7.85}\\{$\pm$0.43}}} & {\scriptsize\shortstack[r]{{$+$8.82}\\{$\pm$0.40}}} & {\scriptsize\shortstack[r]{{$-$0.189}\\{$\pm$0.010}}} & {\scriptsize\shortstack[r]{{$-$0.530}\\{$\pm$0.013}}} & {\scriptsize\shortstack[r]{{$-$0.577}\\{$\pm$0.013}}} \\
\bottomrule

\end{tabular}}
\vspace{-0.2cm} \end{table}

\begin{figure*}
\centering
\includegraphics[width=0.9\textwidth,trim={0 17cm 0 0},clip]{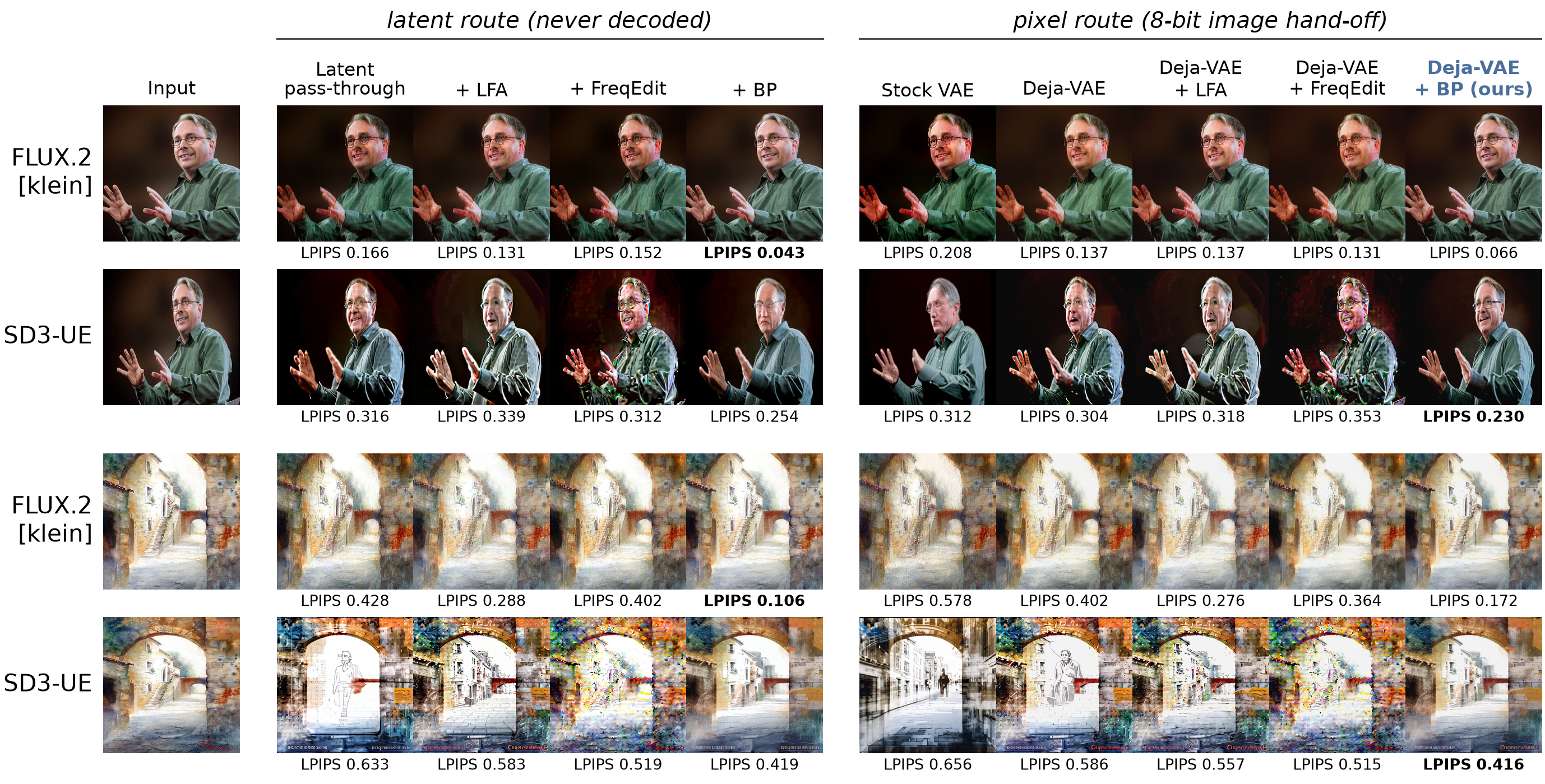}
\par\vspace{-0.35cm}
\caption{\small \textbf{Ten no-op turns.} One photograph after ten no-op turns, seed 0; pixel routes use 8-bit hand-offs; latent routes pass latents directly.
Numbers below panels: LPIPS versus the input; bold: lowest per editor.
Ours preserves this image best across editors and routes.}
\label{fig:noop-routes}
\end{figure*}

\vspace{-0.1cm}\paragraph{Role of the codec.}
BP restores the selected coefficients of $z_{\mathrm{prev}}$, including any hand-off error they contain. With the stock VAE, that error enters the reference every turn and accumulates; with the wrapped codec, Eq.~\ref{eq:guarantee} limits it to quantization and clipping at the 8-bit hand-off.

\vspace{-0.1cm}\paragraph{Prior work and limits.}
After each turn, Emu Edit~\cite{emuedit} reverts pixels whose low-passed RGB change falls below a fixed threshold; BP applies this idea to latent wavelet coefficients, with thresholds set by the editor's measured drift. Eq.~\ref{eq:bp} has the form of DDNM's range-space rectification~\cite{ddnm}, which enforces known linear measurements at each sampling step; FreqEdit's injection is a related wavelet-domain update with continuous weights, applied to velocities during sampling; moved to BP's position, it does not reproduce BP's results (Appendix~\ref{app:bp-placement}). Beyond the global-edit limits of Section~\ref{sec:limitations}, BP assumes spatial alignment: re-rendering or shifts can cause block or ghosting artifacts.

\section{Experiments}
\label{sec:experiments}
\vspace{-0.1cm}
We use complementary benchmarks to compare each baseline on its supported models and test whether our gains extend across editors and workflows. Existing consistency methods differ in model support: some, such as REED-VAE~\cite{reed} and ACEdit~\cite{acedit}, require model-specific fine-tuning, and FreqEdit~\cite{freqedit} intervenes in the editor's denoising. Our construction adapts across pretrained editors through their VAEs and output latents, with calibration but no training. We first separate errors from autoencoding and image hand-offs from drift introduced by the editor. We then test whether stability carries over to longer sessions as edits accumulate or are reversed, measuring both preservation of unedited content and execution of requested edits, as a method that blocks edits would otherwise look stable. PasteBench extends this evaluation to workflows that require pixel access, where sequential pasting, harmonization, and removal exercise both the autoencoder and editor.

\subsection{Evaluation setup}
\label{sec:protocol}
\vspace{-0.1cm}
\vspace{-0.1cm}\paragraph{Routes and comparisons.} \emph{Pixel routes} pass 8-bit images between turns; \emph{latent routes} pass latents directly. Baselines include stock pipelines, VAE-LFA~\cite{vaelfa}, FreqEdit~\cite{freqedit}, Emu Edit thresholding~\cite{emuedit} (Appendix~\ref{app:methods}), and two trained baselines: REED-VAE~\cite{reed} (decoder) and ACEdit~\cite{acedit} (FLUX.2 [klein] editor). \emph{Ours} combines the \method codec (Section~\ref{sec:codec}) and BP with frozen weights; \emph{w/o BP} uses only the \method codec, and \emph{w/o wrapper} uses BP with the stock VAE (stock-codec BP). Competing corrections use the stock VAE unless marked \emph{wrapped} (\method codec); direct latent BP is labeled \emph{BP}. Beyond each correction's native hand-off ($^\ast$ in tables; e.g., latents for VAE-LFA's white-box mode, decoded images for FreqEdit), we run VAE-LFA, FreqEdit, and BP in the other routes and with the \method codec, separating autoencoding drift from editor drift. Table~\ref{tab:metrics} and Appendix~\ref{app:reproducibility} give the metrics and settings.

\subsection{Repeated encode--decode cycles}
\label{sec:cycles}
\vspace{-0.1cm}
We first remove the editor and reconstruct REED's 179 images 25 times with 8-bit hand-offs (Figure~\ref{fig:cycles}, Table~\ref{tab:cycles}). From rounds 5 to 25, SD~1.5 loses 12.48\,dB PSNR, versus 0.57\,dB with the \method codec; SD~2.1 and FLUX.2 show similar gains. Fine-tuned REED leads on SD~2.1 at round 5, but at round 25 the \method codec reaches 28.34\,dB and 0.07 LPIPS, versus 20.93\,dB and 0.25. It reduces accumulated error. Eq.~\ref{eq:guarantee} excludes 8-bit clipping and rounding.

\subsection{Iterative editing on the REED benchmark}
\label{sec:reed-editing}
\vspace{-0.1cm}
We compare the training-free \method with REED-VAE, a decoder fine-tuned for repeated editing, on REED's 25-round protocol: its 179 ImagenHub~\cite{imagenhub} examples with InstructPix2Pix~\cite{ip2p}, MagicBrush~\cite{magicbrush}, DiffEdit~\cite{diffedit}, Paint-by-Example~\cite{pbe}, and SD Inpainting~\cite{ldm}. Rounds 5, 15, and 25 are compared with each method's first edited image, which measures retention, not instruction following.

\vspace{-0.2cm}\paragraph{Results.} At round 25, \method surpasses fine-tuned REED-VAE in all four metrics for every editor (Tables~\ref{tab:reed-editing} and~\ref{tab:reed-editing-full}). PSNR and LPIPS gains over stock are significant across all evaluated rounds (95\% paired bootstrap). BP improves on the codec alone for every editor at round 25; for DiffEdit, the benefit is mainly perceptual, with lower LPIPS but little PSNR change.

\begin{table}
\centering
\footnotesize
\setlength{\tabcolsep}{3pt}
\caption{\small \textbf{VAE-LFA protocols.} 120 images, drift at round 10 vs. the input.
$\Delta E$: mean CIELAB color difference; Cycle: edit/undo pairs with both steps verified by EdiVal's evaluation~\cite{edival} (\%, Appendix~\ref{app:edit-verification}).
Bold/underlined: best/second within route and editor; $^\ast$: native hand-off route;
Full results: Tables~\ref{tab:vaelfa-noop} and~\ref{tab:vaelfa-cycle}.}
\label{tab:vaelfa-summary}
\par\vspace{-0.35cm}
\begin{tabular}{@{}cclrrrr@{}}
\toprule
& & & \multicolumn{2}{c}{No-op} & \multicolumn{2}{c}{Edit/undo} \\
\cmidrule(lr){4-5}\cmidrule(l){6-7}
& & Method & LPIPS$\downarrow$ & $\Delta E\downarrow$ & LPIPS$\downarrow$ & Cycle$\uparrow$ \\
\midrule
\multirow{10}{*}{\rotatebox{90}{\footnotesize\emph{FLUX.2 [klein]}}} & \multirow{6}{*}{\rotatebox{90}{\footnotesize\emph{Pixel}}} & Stock & .41 & 18.7 & .48 & \textbf{89.0} \\
 &  & VAE-LFA & .33 & \underline{11.8} & .43 & 87.2 \\
 &  & FreqEdit$^\ast$ & .38 & 17.3 & .46 & 73.6 \\
 &  & Ours w/o BP & .35 & 17.3 & .44 & \underline{88.6} \\
 &  & Ours w/o wrapper & \underline{.32} & 14.1 & \underline{.41} & 88.0 \\
 &  & \textbf{Ours} & \textbf{.16} & \textbf{9.0} & \textbf{.31} & 87.0 \\
\cmidrule(l){2-7}
 & \multirow{4}{*}{\rotatebox{90}{\footnotesize\emph{Latent}}} & Uncorrected & .34 & 16.3 & .43 & \underline{88.1} \\
 &  & VAE-LFA$^\ast$ & \underline{.27} & \underline{10.6} & \underline{.39} & 87.2 \\
 &  & FreqEdit & .30 & 14.4 & .41 & 73.1 \\
 &  & BP & \textbf{.15} & \textbf{8.1} & \textbf{.31} & \textbf{88.2} \\
\midrule
\multirow{10}{*}{\rotatebox{90}{\footnotesize\emph{SD3-UE}}} & \multirow{6}{*}{\rotatebox{90}{\footnotesize\emph{Pixel}}} & Stock & .48 & 25.1 & .60 & \underline{55.2} \\
 &  & VAE-LFA & .46 & 19.6 & .58 & \textbf{58.8} \\
 &  & FreqEdit$^\ast$ & .43 & 20.9 & \underline{.54} & 45.1 \\
 &  & Ours w/o BP & .47 & 25.0 & .60 & 54.7 \\
 &  & Ours w/o wrapper & \underline{.42} & \underline{15.9} & .55 & 51.4 \\
 &  & \textbf{Ours} & \textbf{.28} & \textbf{14.6} & \textbf{.52} & 54.0 \\
\cmidrule(l){2-7}
 & \multirow{4}{*}{\rotatebox{90}{\footnotesize\emph{Latent}}} & Uncorrected & .47 & 25.5 & .60 & 53.8 \\
 &  & VAE-LFA$^\ast$ & .47 & \underline{20.5} & .58 & \textbf{54.8} \\
 &  & FreqEdit & \underline{.42} & 21.7 & \underline{.54} & 43.7 \\
 &  & BP & \textbf{.28} & \textbf{14.4} & \textbf{.52} & \underline{54.7} \\
\bottomrule
\end{tabular}

\vspace{-0.5cm} \end{table}

\begin{figure*}
\centering
\input{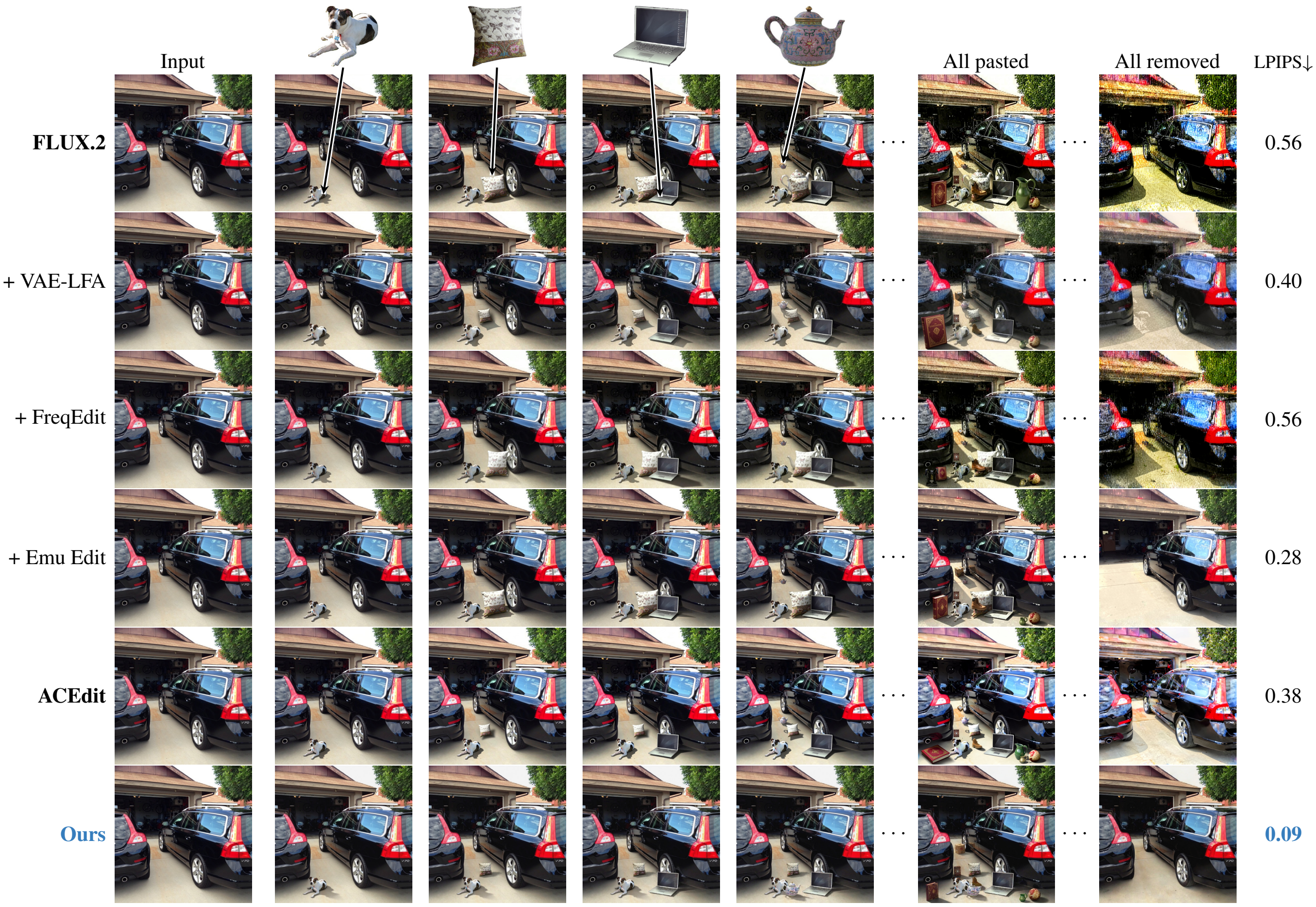}
\par\vspace{-1.1cm}
\caption{\small \textbf{A PasteBench chain.} FLUX.2 [klein] 9B, 8-bit hand-offs. {Turns 1--4 paste and harmonize objects; arrows mark the pasted objects. Turns 5--10 paste six more. Removal follows in reverse order (all removed: turn 20). Right: final whole-image LPIPS to the input. Ours removes every object and keeps the scene. Emu Edit erases the parked car on the left; the other methods shift colors and corrupt the garage.} Averages: Table~\ref{tab:pastebench}.}
\label{fig:pastebench}
\end{figure*}

\begin{figure*}
\centering
\input{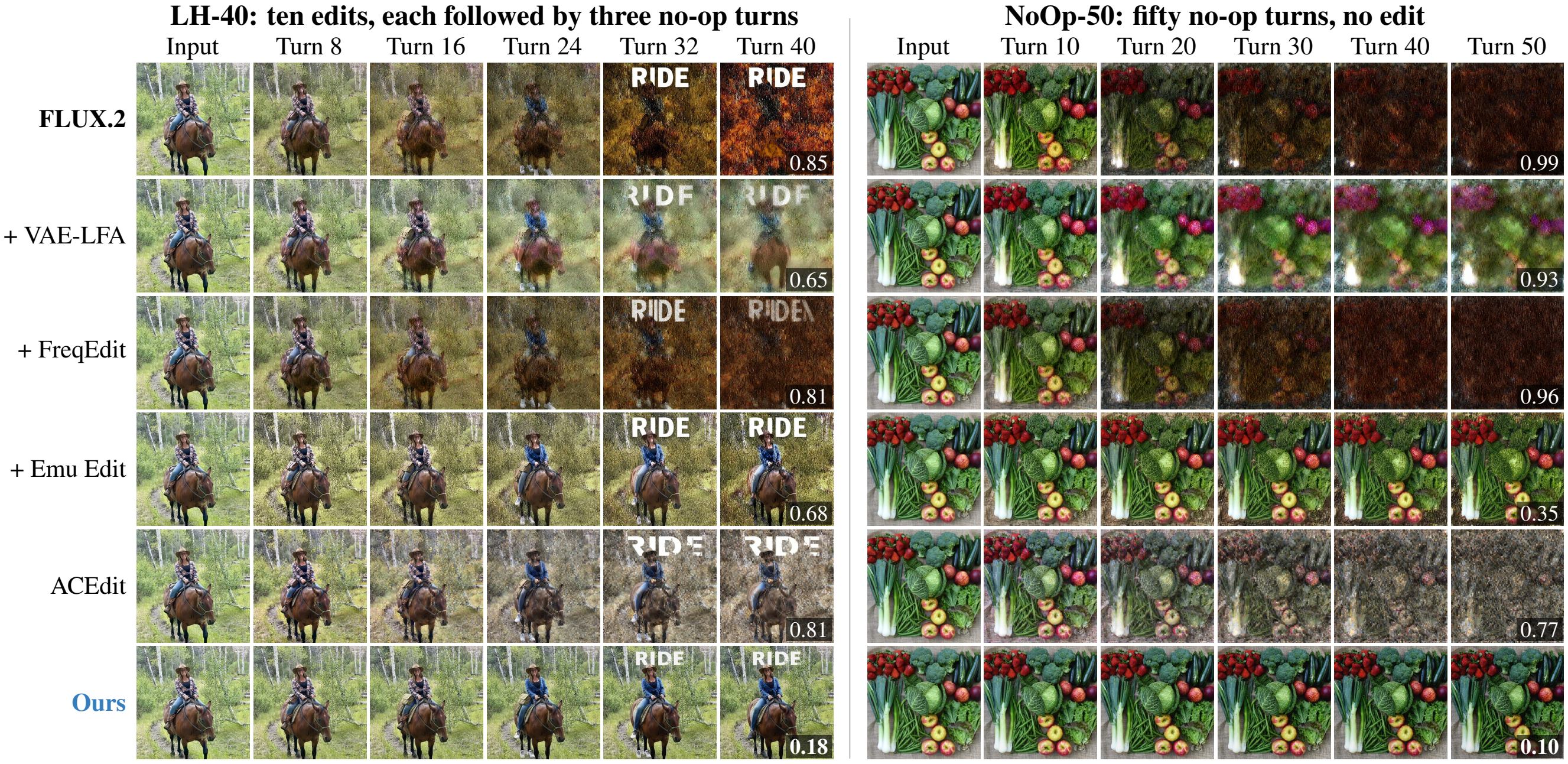}
\par\vspace{-0.75cm}
\caption{\small \textbf{Long-horizon drift.} FLUX.2 [klein]. Left: ten edits, each followed by three no-op turns. Right: fifty no-op turns. Last-panel values are LPIPS to the input. Ours preserves more content while completing seven of ten edits here, as does ACEdit; the other methods complete six (Appendix~\ref{app:edival10-full}). Averages: Table~\ref{tab:edival10}.}
\label{fig:long-horizon}
\end{figure*}

\vspace{-0.2cm}\paragraph{Editor drift remains.} For the three editors that accept latents, the \method codec alone comes within 0.007 round-25 LPIPS of the latent route despite 8-bit hand-offs, while soft-limiting the stock decoder changes PSNR by at most 0.35\,dB (Tables~\ref{tab:reed-editing-full} and~\ref{tab:no-softlimit}). It nearly matches latent-route retention, yet substantial drift remains without hand-offs.

\subsection{No-op and edit/undo evaluation}
\label{sec:vaelfa}
\vspace{-0.1cm}
We evaluate all methods with VAE-LFA's ``no-op editing'' and ``cycle editing'' protocols~\cite{vaelfa} on 120 images with FLUX.2 [klein] and SD3-UE (with corrected decoding, Appendix~\ref{app:vaelfa-setup}). No-op editing isolates drift; cycle editing alternates edits with their undos (edit/undo below). Cycle is the share of pairs whose edit and undo EdiVal's evaluation~\cite{edival} both verifies (Appendix~\ref{app:edit-verification}; examples: Figures~\ref{fig:noop-routes} and~\ref{fig:cycle-chain}).

\begin{table*}
\begin{minipage}[t]{0.58\textwidth}
\vspace{0pt}\centering
\caption{\small \textbf{Cumulative editing and long horizons.}
197 chains, matched seeds; scoring at $512^2$.
Cells: Edit success (\%) / background LPIPS versus the input; NoOp-50: whole-image LPIPS after fifty no-op turns.
EdiVal-10$^\dagger$: our ten-edit chains following ACEdit's EdiVal-10~\cite{acedit}; LH-40 adds three no-op turns after each edit. $^\ast$: native hand-off route.
The appendix provides full results in Table~\ref{tab:edival10-all} and results at $1024^2$ resolution in Table~\ref{tab:edival10-full}.}
\label{tab:edival10}
\par\vspace{-0.35cm}
\fontsize{7.7}{9}\selectfont
\setlength{\tabcolsep}{1.2pt}
\begin{tabular*}{\linewidth}{@{\extracolsep{\fill}}clrrr@{\hspace{5pt}}rrr@{}}
\toprule
& & \multicolumn{3}{c}{\shortstack{\textbf{FLUX.2 [klein]}\\[-1pt]{\scriptsize (Edit$\uparrow$ / Bg LPIPS$\downarrow$)}}}
  & \multicolumn{3}{c}{\shortstack{\textbf{SD3-UE}\\[-1pt]{\scriptsize (Edit$\uparrow$ / Bg LPIPS$\downarrow$)}}} \\
\cmidrule(lr){3-5}\cmidrule(l){6-8}
& Method & EdiVal-10$^\dagger$ & LH-40 & NoOp-50 & EdiVal-10$^\dagger$ & LH-40 & NoOp-50 \\
\midrule
\multirow{7}{*}{\rotatebox{90}{\emph{Pixel hand-off}}}
 & Stock & \textbf{66.35}\,/\,.402 & 64.26\,/\,.697 & .730 & \underline{46.14}\,/\,.453 & 40.76\,/\,.655 & .755 \\
 & VAE-LFA & \textbf{66.35}\,/\,.341 & 64.16\,/\,.550 & .590 & 46.04\,/\,.432 & \underline{43.35}\,/\,.598 & .703 \\
 & FreqEdit$^\ast$ & 61.42\,/\,.367 & 59.70\,/\,.759 & .848 & 42.13\,/\,\underline{.400} & 33.71\,/\,.738 & .784 \\
 & Emu Edit$^\ast$ & 66.09\,/\,\underline{.265} & \textbf{64.97}\,/\,\underline{.357} & \underline{.231} & \textbf{47.87}\,/\,.415 & \textbf{44.21}\,/\,\underline{.444} & \underline{.332} \\
 & Ours w/o BP & \underline{66.25}\,/\,.386 & \underline{64.72}\,/\,.671 & .705 & 44.06\,/\,.441 & 33.05\,/\,.676 & .739 \\
 & Ours w/o wrapper & 65.13\,/\,.277 & 62.08\,/\,.670 & .763 & 40.25\,/\,.411 & 23.60\,/\,.806 & .843 \\
 & \textbf{Ours} & 64.37\,/\,\textbf{.216} & 63.66\,/\,\textbf{.249} & \textbf{.082} & 43.55\,/\,\textbf{.339} & 40.81\,/\,\textbf{.408} & \textbf{.314} \\
\midrule
\multirow{5}{*}{\rotatebox{90}{\emph{Latent}}}
 & Uncorrected & \underline{65.48}\,/\,.383 & \underline{64.77}\,/\,.671 & .700 & \textbf{46.55}\,/\,.436 & 36.45\,/\,\underline{.637} & \underline{.715} \\
 & VAE-LFA$^\ast$ & 65.18\,/\,.325 & \textbf{65.23}\,/\,.526 & \underline{.529} & \underline{46.45}\,/\,.436 & \underline{38.32}\,/\,.645 & \underline{.715} \\
 & FreqEdit & 62.18\,/\,.340 & 60.56\,/\,.658 & .730 & 40.61\,/\,\underline{.386} & 26.95\,/\,.698 & .750 \\
 & ACEdit$^\ast$ & \textbf{66.04}\,/\,\underline{.271} & 63.71\,/\,\underline{.516} & .560 & \multicolumn{3}{c}{---} \\
 & BP & 65.13\,/\,\textbf{.206} & 64.47\,/\,\textbf{.219} & \textbf{.030} & 46.04\,/\,\textbf{.336} & \textbf{41.47}\,/\,\textbf{.364} & \textbf{.254} \\
\bottomrule
\end{tabular*}
\end{minipage}\hfill
\begin{minipage}[t]{0.4\textwidth}
\vspace{0pt}\centering\footnotesize
\setlength{\tabcolsep}{1.2pt}
\newcommand{\pbkept}[3]{%
{%
    \shortstack[r]{#1 {\scriptsize[#2, #3]}}}}
\caption{\small \textbf{PasteBench: paste ten objects, then remove them.}
Recovery: turn 20 vs.\ input; 20 scenes, FLUX.2 [klein] 9B, 8-bit hand-offs.
Bg PSNR excludes object footprints dilated by 32 pixels.
DINO~\cite{dino}: ViT-B/16 similarity.
Kept: VLM-recognizable targets at turn 10 (mean/10; brackets: 95\% scene-bootstrap CI).
Bold/underlined: best/second means.
Details: Appendices~\ref{app:pastebench-full} and~\ref{app:pastebench-retention}; full protocol and per-image annotated results: supplementary material.}
\label{tab:pastebench}
\par\vspace{-0.1cm}
\begin{tabular*}{\linewidth}{@{\extracolsep{\fill}}lrrrrr@{}}
\toprule
Method & LPIPS$\downarrow$ & PSNR$\uparrow$ &
\shortstack{Bg\\PSNR$\uparrow$} & DINO$\uparrow$ &
\multicolumn{1}{c}{\shortstack{Kept$\uparrow$ (/10)}} \\
\midrule
FLUX.2 & .606 & 13.66 & 13.96 & .662 &
\pbkept{8.65}{7.90}{9.25} \\[1pt]
VAE-LFA & .458 & 18.06 & 18.48 & .701 &
\pbkept{8.85}{8.35}{9.30} \\[1pt]
FreqEdit & .743 & 14.67 & 15.01 & .520 &
\pbkept{\textbf{9.25}}{8.85}{9.60} \\[1pt]
Emu Edit & \underline{.203} & \underline{21.06} &
\underline{22.80} & \underline{.945} &
\pbkept{8.60}{7.90}{9.20} \\[1pt]
ACEdit & .513 & 16.07 & 16.28 & .726 &
\pbkept{\underline{8.95}}{8.50}{9.35} \\[1pt]
\midrule
\textbf{Ours} & \textbf{.134} & \textbf{22.13} &
\textbf{23.90} & \textbf{.961} &
\pbkept{8.80}{8.30}{9.25} \\[1pt]
\bottomrule
\end{tabular*}
\end{minipage}
\end{table*}

\vspace{-0.2cm}\paragraph{Results.} \method gives the lowest pixel-route LPIPS in both protocols on both editors, approaching latent BP despite 8-bit hand-offs (Table~\ref{tab:vaelfa-summary}). Edit/undo success is slightly below stock. Figure~\ref{fig:noop-routes} shows reduced no-op drift; in Figure~\ref{fig:cycle-chain}, ours adds and removes each accessory, while FreqEdit leaves the collar behind.

\subsection{Cumulative editing and long horizons}
\label{sec:edival10}
\vspace{-0.1cm}
\vspace{-0.1cm}\paragraph{Protocol.} Following ACEdit's EdiVal-10 protocol~\cite{acedit}, EdiVal-10$^\dagger$ extends EdiVal~\cite{edival} to 197 chains of ten cumulative local edits on GEdit-Bench photographs~\cite{step1x}, using EdiVal's released generation and evaluation code. LH-40 inserts three no-op turns after each edit; NoOp-50 requests no changes for fifty turns. We evaluate FLUX.2 [klein] 9B (four denoising steps) and SD3-UE, with matched seeds. We compare corrections in both routes and with the stock VAE and the \method codec; construction details are in Appendix~\ref{app:edival10-setup}.

\vspace{-0.2cm}\paragraph{Metrics.} \emph{Edit} is the share of edits EdiVal's checker~\cite{edival} judges successful (using detector or VLM, by edit type); Background LPIPS~\cite{lpips} excludes all object boxes, as in ACEdit~\cite{acedit}; NoOp-50 uses whole-image metrics. Appendix~\ref{app:metrics} defines the remaining scores; Tables~\ref{tab:edival10-all} and~\ref{tab:edival10-full} give full metrics and resolution checks.

\vspace{-0.2cm}\paragraph{Results.} \method achieves the lowest pixel-route LPIPS on both editors across all three protocols (Table~\ref{tab:edival10}). On FLUX NoOp-50, either component alone leaves drift near stock levels; together they largely suppress it. Figure~\ref{fig:long-horizon} shows ours preserving the rider and woodland through intervening no-ops. Preservation can cost edit success: fine-tuned ACEdit does better on FLUX EdiVal-10$^\dagger$, and Emu Edit and VAE-LFA on SD3-UE LH-40.

\subsection{PasteBench}
\label{sec:pastebench}
\vspace{-0.2cm}

PasteBench contains 20 Open Images~\cite{openimages} scenes, each with ten objects from a library of 100 cutouts from its instance masks~\cite{benenson2019}. Each chain sequentially pastes and harmonizes the objects, then removes them in reverse order, targeting recovery of the original scene after 20 turns. Methods share objects and placements but evolve on their own outputs. Table~\ref{tab:pastebench} reports turn-10 edit retention and turn-20 recovery. For edit retention, a VLM counts recognizable targets against the original scene and cutouts, allowing appearance and position changes. Figure~\ref{fig:pastebench} shows intermediate edits and failures; Figure~\ref{fig:teaser} (bottom right) compares each scene's first $K$ objects. The supplement provides the scoring protocol, annotations, and an HTML viewer of all scenes and methods at turns 0, 10, and 20.

\vspace{-0.2cm}\paragraph{Results.} Without training, \method{} leads all four recovery metrics, with 8.80/10 targets visible at turn 10 (FLUX.2: 8.65). Its recovery advantage grows with chain length (Figure~\ref{fig:teaser}, bottom right). In Figure~\ref{fig:pastebench}, it removes the inserted objects while preserving both parked cars; Emu Edit also erases the original car on the left, and other methods distort the garage or its colors.

\subsection{Cost}
\label{sec:cost}
\vspace{-0.1cm}
A wrapped cycle uses two encoder calls and one decoder call, versus one of each for the stock codec. BP adds per-turn wavelet transforms and correction without network calls after calibration (Appendix~\ref{app:runtime}).

\section{Limitations}
\label{sec:limitations}
\vspace{-0.1cm}
Global edits, such as style transfer or broad color and lighting changes, are harder to stabilize. Their intended changes can span many wavelet coefficients with magnitudes comparable to drift, so BP may suppress part of the requested transformation. Per-turn noise estimation can amplify this problem: widespread edits raise the estimated drift level and hence the thresholds. The resulting bias toward preservation can weaken global edits. {Appendix~\ref{app:global-edits} (Figure~\ref{fig:global-edits}) shows this on FLUX.2 [klein]: BP weakened all six global edits tested, almost completely reverting the style edit, while the \method codec alone followed them like the stock pipeline.}

\clearpage
\section*{LLM Usage Statement}
\vspace{-0.1cm}
We used OpenAI Codex and Anthropic Claude to assist with technical sanity checks, experiment organization, coding, processing and analyzing experimental results, figure preparation, and text refinement. The technical ideas and methodological decisions are ours. We take full responsibility for all content, results, and conclusions presented in this paper.
\section*{Acknowledgments}
\vspace{-0.1cm}
A.S. is a Chaya Fellow and acknowledges support from the Chaya Career Advancement Chair and the Alon Fellowship of the Israeli Council for Higher Education. This research was supported by the Israel Science Foundation, Grant No. 1039/26. This research was made possible through a GPU compute resource grant funded by the Association of University Heads, the Council for Higher Education and the AI Research Compute Center.

\noindent We thank Sara Dorfman for her help with ACEdit.

{\small
\bibliographystyle{ieeenat_fullname}
\bibliography{references}
}

\clearpage
\renewcommand{\thetitle}{\papertitle}
\maketitlesupplementary
\appendix
\raggedbottom
\section{Method details}
\label{app:general}

\subsection{Pixel-range handling}

Images are in $[-1,1]$, with $x=u/127.5-1$ for 8-bit values $u$. The coordinates of Eq.~\ref{eq:y0} act on non-overlapping $8\times8\times3$ patches, one per latent position ($n=192$), and $k$ equals the number of latent channels: 4 for SD~1.5 and SD~2.1, 16 for SD3-UE, and 32 for FLUX.2. Latents are the encoder's unscaled posterior means; the diffusion model's latent scaling is applied outside the codec.

Clipping or rounding a decoded image breaks the cancellation behind Eq.~\ref{eq:guarantee}. Two fixed elementwise maps, neither learned, handle the pixel range. The \emph{soft limit}

\begin{equation}
 \mbox{}\mbox{$\psi$}(v)=\begin{cases}v, & |v|\le \kappa,\\ \operatorname{sgn}(v)\big[\kappa+(\lambda-\kappa)\tanh\frac{|v|-\kappa}{\lambda-\kappa}\big], & |v|>\kappa,\end{cases}
\end{equation}
with $\kappa=0.82$ and $\lambda=0.92$, maps $\R$ onto $(-\lambda,\lambda)$. It acts on the pretrained decoder's output, replacing $D$ by $\psi\circ D$ in Eq.~\ref{eq:decoder}, and leaves headroom below $\pm1$ for the reserved-coordinate correction. The \emph{pixel map}

\begin{equation}
 \phi(v)=\begin{cases}v, & |v|\le\beta,\\ \operatorname{sgn}(v)\big[\beta+(1-\beta)\operatorname{artanh}\frac{|v|-\beta}{1-\beta}\big], & |v|>\beta,\end{cases}
\end{equation}
with $\beta=0.98$, is a bijection from $(-1,1)$ onto $\R$; $\phi^{-1}$ replaces $\operatorname{artanh}$ by $\tanh$. With the pixel map, the coordinates of Eq.~\ref{eq:y0} are those of $\phi(x)$, and every image the codec forms is mapped back by $\phi^{-1}$, so $t(x_1)=E\big(\phi^{-1}(\mu+U_1^\top x_1)\big)$, $\hat x_1=U_1(\tilde x-\mu)$ with $\tilde x=\psi(D(z))$ (on which $\phi$ is the identity, since $\lambda<\beta$), and $\hat x_0=(z-t(\hat x_1))/c$:

\begin{align}
 \Ep(x)&=c\,x_0+t(x_1),\\
 \Dp(z)&=\phi^{-1}\big(\mu+U_0^\top\hat x_0+U_1^\top\hat x_1\big)\mbox{}\mbox{.}
\end{align}
 The encoder and decoder share $U$, $\mu$, $c$, and $\phi$; $\psi$ acts only in the decoder. Because $\psi$ only redefines the pretrained decoder and $\phi$ only changes coordinates, Eq.~\ref{eq:guarantee} holds unchanged. With the pixel map, $\Dp$ never leaves $(-1,1)$, so the clamp before 8-bit serialization is a no-op and only rounding remains; the frozen encoder also receives only in-range images. Saturated pixels ($|x|=1$) would map to infinity, so the $\operatorname{artanh}$ argument is capped at $1-4\epsilon/(1-\beta)$, with $\epsilon$ the float32 machine epsilon, and they map to $|\phi(x)|\approx1.093$. Every \method codec uses the soft limit; the codecs of REED editing also use the pixel map (Table~\ref{tab:codecs}).

The ``no range handling'' rows of Table~\ref{tab:no-softlimit} remove both maps. Without the pixel map, $\Dp$ can leave $[-1,1]$; such values are clipped at the 8-bit hand-off, outside the guarantee.

\subsection{Codec configurations}

Table~\ref{tab:codecs} lists every codec.

\begin{table*}
\centering
\footnotesize
\setlength{\tabcolsep}{3pt}
\begin{tabular}{@{}p{0.24\textwidth}ccp{0.25\textwidth}p{0.15\textwidth}p{0.22\textwidth}@{}}
\toprule
Codec (VAE) & $k$ & $c$ & PCA basis & Range handling & Used in \\
\midrule
SD~1.5 & 4 & 29.8 & 256 LAION~\cite{laion5b} images & soft limit, pixel map & Section~\ref{sec:reed-editing} \\
SD~2.1 (\texttt{sd-vae-ft-mse}) & 4 & 29.8 & same as SD~1.5 & soft limit, pixel map & Section~\ref{sec:reed-editing} \\
SD~1.5, SD~2.1 & 4 & 30 & 256 LAION images & soft limit & Section~\ref{sec:cycles} \\
FLUX.2 [klein] & 32 & 26 & 64 LAION images & soft limit & Sections~\ref{sec:cycles}, \ref{sec:vaelfa},~\ref{sec:edival10}, \ref{sec:pastebench};\newline Figure~\ref{fig:teaser}; Appendix~\ref{app:global-edits} \\
SD3-UE & 16 & 2.828 & the 16 lowest-variance directions of the FLUX.2 basis & soft limit & Sections~\ref{sec:vaelfa}, \ref{sec:edival10} \\
\bottomrule
\end{tabular}
\par
\caption{\footnotesize \textbf{Codec configurations.} Every \method codec reserves $k$ coordinates, one per latent channel, in non-overlapping $8\times8$ RGB patches ($n=192$), along the lowest-variance PCA directions of LAION patches; $c$ is the gain of Eq.~\ref{eq:encoder}. In the reconstruction runs, the \method codec and the stock VAE run in fp32; in the DiT editing runs, the \method codec runs in fp32 and the stock VAE in bf16 (FLUX.2) or fp16 (SD3-UE).}
\label{tab:codecs}
\end{table*}

\subsection{Back-projection on other grid sizes}

On latent grids not divisible by four, $d$ is reflect-padded at the bottom and right before $W$ and cropped after $W^\top$; Eq.~\ref{eq:bp} minimizes distance on the padded grid.

\subsection{BP calibration}
\label{app:bpcalib}

Table~\ref{tab:bpcalib} lists the calibration of each editor. In REED editing, the calibration chains run through the wrapped pixel route with 8-bit hand-offs and record the change between the editor's output latent and the \method codec's encoding of the turn's input image.

\begin{table*}
\centering
\footnotesize
\setlength{\tabcolsep}{3pt}
\begin{tabular}{@{}p{0.22\textwidth}p{0.18\textwidth}p{0.24\textwidth}p{0.2\textwidth}@{}}
\toprule
Editor & Threshold (Eq.~\ref{eq:bp-threshold}) & Calibration chains & Used in \\
\midrule
InstructPix2Pix & band profile & first 100 LAION crops, fixed seed per crop & Section~\ref{sec:reed-editing} \\
MagicBrush & band profile & as above & Section~\ref{sec:reed-editing} \\
{\raggedright DiffEdit, Paint-by-Example, SD Inpainting\par} & $\rho_{b,\ell}=1$: no no-op prompt & --- & Section~\ref{sec:reed-editing} \\
FLUX.2 [klein] 9B & band profile & eight LAION crops, base seed 0 & Sections~\ref{sec:vaelfa}, \ref{sec:edival10}, \ref{sec:pastebench} \\
SD3-UE & band profile & eight LAION crops, base seed 0 & Sections~\ref{sec:vaelfa}, \ref{sec:edival10} \\
\bottomrule
\end{tabular}
\par
\caption{\footnotesize \textbf{BP calibration per editor.} Each calibration chain runs ten ``Keep the image unchanged.'' turns with BP disabled, and turns 2--10 are pooled (Section~\ref{sec:bp}). Calibration uses $512\times512$ LAION crops outside the evaluation set, with one profile per editor across routes.}
\label{tab:bpcalib}
\end{table*}

\subsection{A bijective completion of the wrapper}

With $t$ and $r$ of Section~\ref{sec:codec}, the explicit completion

\begin{equation}
 H(x)=\big(\Ep(x),\ \mbox{}\mbox{$x_1$}-r(\Ep(x))\big)
\end{equation}
is bijective: from $(z,q)$ recover $x_1=q+r(z)$, $x_0=(z-t(x_1))/c$, and $x=\mu+U_0^\top x_0+U_1^\top x_1$. In these coordinates,

\begin{equation}
 \Dp\Ep=H^{-1}\,\mathrm{diag}(I_k,0)\,H.
\end{equation}
Thus $\Dp\Ep$ is a coordinate projection; it is not the closest-point projection in the Euclidean image metric.

\subsection{Runtime}
\label{app:runtime}

On an A100 GPU at $512^2$, an encode--decode cycle takes 204\,ms with the \method codec versus 152\,ms with the stock codec, for each of the SD~1.5, SD~2.1, and FLUX.2 VAEs. A FLUX.2 [klein] turn at $1024^2$ takes 4.27\,s with \method versus 4.05\,s with the stock pipeline (+5\%, including our fp32 codec), of which BP takes 0.01\,s.

\section{Experimental setup}
\label{app:reproducibility}

Table~\ref{tab:settings} gives the settings of each benchmark except PasteBench, whose protocol is in Appendix~\ref{app:pastebench-full}; Table~\ref{tab:metrics} defines every metric.

\subsection{Benchmarks and protocols}
\label{app:settings}

The paragraphs below add what does not fit in Table~\ref{tab:settings}.

\begin{table*}
\centering
\footnotesize
\setlength{\tabcolsep}{3pt}
\begin{tabular}{@{}p{0.1\textwidth}p{0.17\textwidth}p{0.235\textwidth}p{0.215\textwidth}p{0.215\textwidth}@{}}
\toprule
& Reconstruction\newline (Section~\ref{sec:cycles}) & REED editing (Section~\ref{sec:reed-editing}) & No-op, edit/undo (Section~\ref{sec:vaelfa}) & Cumulative editing (Section~\ref{sec:edival10}) \\
\midrule
Data & REED's 179 source images & REED's 179 ImagenHub examples & VAE-LFA's 120 images (60 photos, 60 illustrations) & 197 GEdit-Bench photo chains \\
Turns & 25 encode--decode cycles & 25 editing rounds & 10: no-ops, or five edit/undo pairs & 10 edits; LH-40: 40 turns; NoOp-50: 50 \\
Models & SD~1.5, SD~2.1, FLUX.2 VAEs & InstructPix2Pix, MagicBrush: 100 steps, guidance 7.5, image guidance 1.5, Euler ancestral; DiffEdit on SD~2.1; Paint-by-Example, SD Inpainting: 50 steps & FLUX.2 [klein] 9B: 4 steps, guidance 1.0; SD3-UE: 50 steps, guidance 7.5, image guidance 1.5 & FLUX.2 [klein] 9B: 4 steps, guidance 1.0; SD3-UE: 50 steps, guidance 7.5, image guidance 1.5 \\
Seeds & none (posterior mean) & per chain; per round for Paint-by-Example & $s+r-1$ ($s=0,10$; round $r$) & $t-1$ for turn $t$ \\
Resolution & $512^2$ & $512^2$ & FLUX.2: output long edge $\le1280$, area capped at $1024^2$, scored at input size; SD3-UE: $512^2$ & FLUX.2: $1024^2$ (score: $512^2$); SD3-UE: $512^2$ \\
Codec, BP & Table~\ref{tab:codecs} & Tables~\ref{tab:codecs}, \ref{tab:bpcalib} & Tables~\ref{tab:codecs}, \ref{tab:bpcalib} & Tables~\ref{tab:codecs}, \ref{tab:bpcalib} \\
Edit check & --- & none; retention against round 1 & EdiVal's verifiers (Appendix~\ref{app:edit-verification}) & EdiVal's verifiers \\
\bottomrule
\end{tabular}
\par
\caption{\footnotesize \textbf{Settings per benchmark.} Codecs: Table~\ref{tab:codecs}; BP calibration: Table~\ref{tab:bpcalib}.}
\label{tab:settings}
\end{table*}

\paragraph{REED editing.} Direct latent conditioning is available for InstructPix2Pix, MagicBrush, and DiffEdit; the evaluated Paint-by-Example and SD Inpainting pipelines require pixel images. The stock VAE encodes with the posterior mode for InstructPix2Pix and MagicBrush and samples the posterior for the other three editors; the \method codec uses the posterior mean. REED-VAE scores come from its paper.

\paragraph{No-op and edit/undo.} \label{app:vaelfa-setup} The 120 images span photographs and illustrations of creatures, architecture, and scenery. We use VAE-LFA's runners and metric code. Both seeds (0 and 10) are evaluated. EdiVal's evaluation~\cite{edival} verifies each edit and undo separately and jointly, over 1,200 pairs per method and editor (Appendix~\ref{app:edit-verification}). FLUX.2 outputs keep the input's aspect ratio, with the long edge at most 1280 pixels in multiples of 16, and their area is capped at $1024^2$ pixels so that the pipeline never resamples the image handed between turns; outputs are bicubically resized to the input for scoring. Matched SD3-UE runs, here and in cumulative editing, use corrected decoding: VAE-LFA's released SD3-UE code maps the decoder's $[-1,1]$ output to $[0,1]$ by clamping, which turns every negative value black, so we use $(x+1)/2$, as VAE-LFA's FLUX.2 code does. The SD3-UE code's latent pass-through also conditions on latents 1.53 times too large and replaces the edit mask's latent with the image latent; we fix both. Results with VAE-LFA's unmodified SD3-UE code appear only as gray diagnostic rows (Tables~\ref{tab:vaelfa-noop}--\ref{tab:vaelfa-cycle-full}).

\paragraph{Cumulative editing.} \label{app:edival10-setup} We retain GEdit-Bench photographs~\cite{step1x} whose short side is at least 768 pixels, then center-crop and resize to $1024\times1024$. As in ACEdit's EdiVal-10~\cite{acedit}, each chain has ten local edits and reserves one or two anchor objects that no instruction edits or mentions; our 197 chains (EdiVal-10$^\dagger$) are generated with EdiVal's released GPT-4o pipeline~\cite{edival} and its eight local edit types. LH-40 inserts three ``Keep the image as is.'' turns after each of ten edits, a structure that ACEdit's figures show with two or four such turns; NoOp-50 contains fifty such turns without edits, extending VAE-LFA's ten-turn no-op protocol~\cite{vaelfa}. FLUX.2 outputs are resized to $512^2$ with LANCZOS for scoring. Background metrics average over the chains whose background is not empty at the end.

\paragraph{PasteBench.} \label{app:pastebench-full} Scenes and objects come from Open Images~\cite{openimages}: 20 scene photographs at $1024\times1024$, used without cropping or resizing, and 100 object cutouts from its instance masks~\cite{benenson2019}. Each scene receives ten distinct cutouts at fixed, non-overlapping placements, chosen by visual inspection without editor outputs; cutouts recur across scenes. Turns 1--10 alpha-composite the next cutout onto the previous output and ask the editor to blend it in: ``Blend the newly pasted [object] [position] naturally into the scene. Match the lighting, colors, perspective, and contact shadows. Keep the object recognizable and near its current position. Preserve all other objects and the rest of the image. Do not add or remove anything else.'' Turns 11--20 remove the objects in reverse order: ``Remove the [object] [position], including its cast shadow. Fill the area with the natural background. Preserve every other object and the rest of the image. Do not introduce new objects.'' Every hand-off is an 8-bit PNG, and the editor never receives the clean scene or the masks used for scoring. Turn $t$ of scene $s$ uses seed $1000s+t$. BP's reference $z_{\mathrm{prev}}$ is the previous turn's output, encoded from its 8-bit PNG: on removal turns, this is the turn's input; on insertion turns, it is the image before the paste, so the pasted object counts as part of the turn's edit. All methods receive identical objects and placements, applied to their own preceding outputs. The object-count sweep in Figure~\ref{fig:teaser} (bottom right) evaluates each scene's first $K$ objects with the recovery metrics of Table~\ref{tab:pastebench}. The Kept audit is in Appendix~\ref{app:pastebench-retention}.

\begin{table*}
\centering
\footnotesize
\setlength{\tabcolsep}{3pt}
\begin{tabular}{@{}p{0.15\textwidth}p{0.33\textwidth}p{0.27\textwidth}p{0.06\textwidth}p{0.14\textwidth}@{}}
\toprule
Metric & Measures & Images compared & Better & Source \\
\midrule
\multicolumn{5}{@{}l}{\emph{Reconstruction and REED (Sections~\ref{sec:cycles} and~\ref{sec:reed-editing})}} \\
PSNR (dB), MSE, SSIM & Pixel fidelity & cycle $k$ vs.\ the original; REED: round $k$ vs.\ the same method's round 1 & $\uparrow$, $\downarrow$, $\uparrow$ & standard \\
LPIPS & Perceptual distance (AlexNet) & as above & $\downarrow$ & \cite{lpips} \\
\midrule
\multicolumn{5}{@{}l}{\emph{No-op and edit/undo (Section~\ref{sec:vaelfa})}} \\
LPIPS, L1, SSIM & Drift; SSIM is VAE-LFA's global grayscale SSIM & round 10 vs.\ the input, whole image & $\downarrow$, $\downarrow$, $\uparrow$ & VAE-LFA~\cite{vaelfa} \\
PSNR (dB), $\Delta E$ & Pixel drift and mean CIE76 color difference ($\Delta E$) & round 10 vs.\ the input, whole image & $\uparrow$, $\downarrow$ & ours \\
Edit, Undo, Cycle (\%) & EdiVal verifies the edit, undo, or both & each edit/undo pair & $\uparrow$ & adapted from EdiVal~\cite{edival} \\
\midrule
\multicolumn{5}{@{}l}{\emph{Cumulative editing (Section~\ref{sec:edival10}); at $512\times512$}} \\
Edit (\%) & Share of edits that EdiVal's verifiers accept: a Qwen2-VL-7B check, with GroundingDINO detection for object edits & the turn's output vs.\ the previous turn's & $\uparrow$ & EdiVal~\cite{edival} \\
IF@5 (\%) & Share of chains whose first five edits all succeed & first five edits of a chain & $\uparrow$ & EdiVal \\
CC (\%) & DINOv3~\cite{dinov3} similarity of untouched objects and background & last turn vs.\ the input & $\uparrow$ & EdiVal \\
O@5 (\%) & $\sqrt{\mathrm{IF@5}\cdot\mathrm{CC@5}}$, with CC after five edits & --- & $\uparrow$ & EdiVal \\
Bg LPIPS & Drift outside the boxes of all the chain's objects, detected in both images (EdiVal's mask) & last turn vs.\ input; $1024^2$: Table~\ref{tab:edival10-full} & $\downarrow$ & adapted from ACEdit~\cite{acedit} \\
Bg $\Delta E$ & Color drift under the same mask & last turn vs.\ the input & $\downarrow$ & ours \\
$\Delta$HPS & Change in HPSv3~\cite{hpsv3} image quality & first vs.\ last scored turn & $\uparrow$ & as in ACEdit~\cite{acedit} \\
NoOp-50 LPIPS, $\Delta E$, PSNR & Whole-image drift & turn 50 vs.\ the input & $\downarrow$, $\downarrow$, $\uparrow$ & ours \\
\midrule
\multicolumn{5}{@{}l}{\emph{PasteBench (Section~\ref{sec:pastebench})}} \\
LPIPS & Perceptual distance (AlexNet) & last turn vs.\ the input, whole image & $\downarrow$ & \cite{lpips} \\
PSNR (dB) & Pixel fidelity & last turn vs.\ the input, whole image & $\uparrow$ & standard \\
Bg PSNR (dB) & PSNR outside all ten pasted-object footprints, dilated by 32 pixels & last turn vs.\ the input & $\uparrow$ & ours \\
DINO & Cosine similarity, DINO ViT-B/16 & last turn vs.\ the input & $\uparrow$ & \cite{dino} \\
Kept & Pasted objects (of 10) still recognizable, by a VLM audit (Appendix~\ref{app:pastebench-retention}) & turn 10 & $\uparrow$ & ours \\
\bottomrule
\end{tabular}
\par
\caption{\footnotesize \textbf{Metrics.} Columns give what each metric measures, which images it compares, whether higher or lower is better, and whose definition we use. VAE-LFA's L1 averages absolute RGB differences in $[0,1]$; our PSNR is computed per image on $[0,1]$, then averaged. Background LPIPS averages the spatial LPIPS map over background pixels, then over the chains with a non-empty background.}
\label{tab:metrics}
\end{table*}

\subsection{Compared methods}
\label{app:methods}
\paragraph{Native hand-off routes.} REED-VAE fine-tunes the decoder for image hand-offs~\cite{reed}; Emu Edit thresholds pixels after each turn~\cite{emuedit}; FreqEdit's released multi-turn pipeline passes decoded images between turns~\cite{freqedit}; VAE-LFA's white-box mode passes aligned latents without VAE round trips, and its black-box mode aligns through an external VAE~\cite{vaelfa}; ACEdit passes latents between turns~\cite{acedit}. Tables~\ref{tab:vaelfa-summary} and~\ref{tab:edival10} mark these native settings with $^\ast$.

\paragraph{Emu Edit thresholding.} Emu Edit's model is not released, so we apply its sequential edit thresholding~\cite{emuedit} to the compared editors with MT-OPSD's implementation and settings~\cite{mtopsd}: each output pixel reverts to the turn's input where its mean absolute RGB change from that input (images in $[0,1]$), blurred by a Gaussian with $\sigma=3$ pixels, is below $\alpha=0.03$.

\paragraph{VAE-LFA.} Pixel-route VAE-LFA applies its alignment to the editor's output latent before the 8-bit hand-off, outside its published modes and therefore unmarked; its latent row is VAE-LFA's white-box mode.

\paragraph{FreqEdit.} Following ACEdit~\cite{acedit}, we adapt FreqEdit's released code so that each turn is conditioned only on the previous output; the quality guidance toward the original image that its Kontext version adds is therefore not used, and the other settings are taken from the Kontext version (two-level db4, $\alpha_0=1.6$, $\gamma=2$, path compensation). Kontext injects during the first 30\% of its 28 steps; we inject during 15 of SD3-UE's 50 steps and, on FLUX.2, during the first two of four steps, which fall in the same noise-level range.

\paragraph{ACEdit.} ACEdit uses its LoRA checkpoint, provided by the authors, on FLUX.2 [klein] 9B, with latent hand-offs in cumulative editing.

\subsection{Metrics}
\label{app:metrics}

Table~\ref{tab:metrics} defines each metric and gives the images it compares and its source. Drift metrics alone can reward a correction that blocks requested changes, so the edit/undo and cumulative evaluations also report edit success; the REED benchmark measures retention against each method's first edited image.

\paragraph{Verifying edits and undos.}\label{app:edit-verification} For VAE-LFA's cycle editing (Section~\ref{sec:vaelfa}), we apply EdiVal's instruction-following verifiers~\cite{edival} unchanged (Qwen2-VL-7B and GroundingDINO, images resized to $512\times512$) to every edit/undo pair. For object additions, the edit uses EdiVal's add verifier (a VLM check that the object newly appears, then GroundingDINO detection) and the undo its remove verifier (no detection at the object's location). All other pairs use EdiVal's two-image VLM verifier with the pair's instruction, for the edit and for the undo. Cycle requires both.

\paragraph{Cumulative-editing metrics.} Table~\ref{tab:metrics} defines the metrics of EdiVal-10$^\dagger$, LH-40, and NoOp-50 (Section~\ref{sec:edival10}). LH-40 scores edit $k$ at turn $4k$, after its three no-op turns. Background metrics are also evaluated at the $1024\times1024$ generation resolution (Table~\ref{tab:edival10-full}).

\FloatBarrier
\section{Full results}
\label{app:results}

\subsection{Reconstruction}
\label{app:cycles-full}
\paragraph{Residual drift.} The serialized reconstruction test includes first-pass error and accumulated drift. The codec's accumulated drift comes from 8-bit rounding: rounding errors in reserved coordinates are amplified by the gain during re-encoding. Table~\ref{tab:cycles} reports PSNR at cycles 5, 15, and 25 and LPIPS at cycle 25.

\begin{table}
\centering
\small
\setlength{\tabcolsep}{4pt}
\begin{tabular}{@{}lcccc@{}}
\toprule
& \multicolumn{3}{c}{PSNR$\uparrow$} & LPIPS$\downarrow$ \\
\cmidrule(lr){2-4}\cmidrule(l){5-5}
VAE & @5 & @15 & @25 & @25 \\
\midrule
SD 2.1 & 26.52 & 19.53 & 14.84 & 0.70 \\
SD 2.1 / REED-VAE (paper) & \textbf{30.50} & \underline{24.10} & \underline{20.93} & \underline{0.25} \\
SD 2.1 / \method codec & \underline{29.09} & \textbf{28.68} & \textbf{28.34} & \textbf{0.07} \\
\midrule
SD 1.5 & 22.09 & 13.36 & 9.61 & 0.76 \\
SD 1.5 / \method codec & \textbf{28.07} & \textbf{27.76} & \textbf{27.50} & \textbf{0.07} \\
\midrule
FLUX.2 & 29.86 & 20.03 & 15.00 & 0.76 \\
FLUX.2 / \method codec & \textbf{34.98} & \textbf{33.10} & \textbf{32.01} & \textbf{0.03} \\
\bottomrule
\end{tabular}
\par
\caption{\footnotesize \textbf{Repeated encode--decode cycles, no editing} (protocol of REED-VAE's Table~A2~\cite{reed}: 179 images, 8-bit hand-off, metrics against the original image). The \method codec uses no training; REED fine-tunes the SD~2.1 decoder. Bold: best per backbone; underlined: second best (SD~2.1).}
\label{tab:cycles}
 \end{table}

\subsection{REED benchmark}
\label{app:reed-full}

In Table~\ref{tab:reed-editing-full}, MSE, LPIPS, SSIM, and PSNR compare rounds 5, 15, and 25 with the first edited image. 

\begin{table*}
\centering
\small
\resizebox{\textwidth}{!}{%
\begin{tabular}{@{}llcccccccccccc@{}}
\toprule
& & \multicolumn{3}{c}{MSE$\downarrow$} & \multicolumn{3}{c}{LPIPS$\downarrow$} & \multicolumn{3}{c}{SSIM$\uparrow$} & \multicolumn{3}{c}{PSNR$\uparrow$} \\
\cmidrule(lr){3-5}\cmidrule(lr){6-8}\cmidrule(lr){9-11}\cmidrule(l){12-14}
Editor & Method & 5 & 15 & 25 & 5 & 15 & 25 & 5 & 15 & 25 & 5 & 15 & 25 \\
\midrule
\multirow{6}{*}{InstructPix2Pix} & Stock VAE (SD 1.5) & 0.025 & 0.107 & 0.149 & 0.341 & 0.699 & 0.765 & 0.595 & 0.226 & 0.183 & 17.596 & 10.151 & 8.563 \\
 & Soft-limit control & 0.025 & 0.102 & 0.140 & 0.341 & 0.690 & 0.751 & 0.619 & 0.277 & 0.231 & 17.664 & 10.313 & 8.801 \\
 & REED-VAE (paper) & 0.020 & 0.060 & 0.090 & 0.180 & \underline{0.450} & \underline{0.580} & 0.800 & \underline{0.530} & \underline{0.410} & 19.810 & 13.360 & \underline{11.360} \\
 & Ours w/o BP & \underline{0.016} & \underline{0.056} & \underline{0.086} & \underline{0.172} & 0.458 & 0.608 & \underline{0.813} & 0.476 & 0.323 & \underline{20.653} & \underline{13.653} & 11.332 \\
 & \textbf{Ours} & \textbf{0.015} & \textbf{0.033} & \textbf{0.044} & \textbf{0.148} & \textbf{0.252} & \textbf{0.324} & \textbf{0.850} & \textbf{0.744} & \textbf{0.675} & \textbf{22.779} & \textbf{18.155} & \textbf{16.314} \\
 & \textcolor{gray}{\emph{latent pass-through}} & \textcolor{gray}{0.017} & \textcolor{gray}{0.059} & \textcolor{gray}{0.091} & \textcolor{gray}{0.166} & \textcolor{gray}{0.453} & \textcolor{gray}{0.609} & \textcolor{gray}{0.807} & \textcolor{gray}{0.461} & \textcolor{gray}{0.301} & \textcolor{gray}{20.539} & \textcolor{gray}{13.416} & \textcolor{gray}{11.085} \\
\midrule
\multirow{6}{*}{MagicBrush} & Stock VAE (SD 1.5) & 0.019 & 0.080 & 0.136 & 0.308 & 0.705 & 0.790 & 0.659 & 0.207 & 0.127 & 19.079 & 11.368 & 8.800 \\
 & Soft-limit control & 0.019 & 0.076 & 0.127 & 0.307 & 0.691 & 0.770 & 0.685 & 0.270 & 0.191 & 19.032 & 11.542 & 9.088 \\
 & REED-VAE (paper) & \textbf{0.010} & 0.030 & \underline{0.050} & 0.190 & 0.510 & 0.690 & 0.810 & \underline{0.600} & \underline{0.450} & 21.530 & 16.660 & \underline{14.090} \\
 & Ours w/o BP & \underline{0.013} & \underline{0.028} & 0.051 & \underline{0.147} & \underline{0.437} & \underline{0.615} & \underline{0.850} & 0.556 & 0.343 & \underline{22.315} & \underline{16.795} & 13.596 \\
 & \textbf{Ours} & 0.014 & \textbf{0.019} & \textbf{0.022} & \textbf{0.113} & \textbf{0.175} & \textbf{0.220} & \textbf{0.880} & \textbf{0.818} & \textbf{0.780} & \textbf{23.140} & \textbf{20.714} & \textbf{19.534} \\
 & \textcolor{gray}{\emph{latent pass-through}} & \textcolor{gray}{0.014} & \textcolor{gray}{0.031} & \textcolor{gray}{0.057} & \textcolor{gray}{0.145} & \textcolor{gray}{0.434} & \textcolor{gray}{0.617} & \textcolor{gray}{0.842} & \textcolor{gray}{0.521} & \textcolor{gray}{0.303} & \textcolor{gray}{21.970} & \textcolor{gray}{16.379} & \textcolor{gray}{13.140} \\
\midrule
\multirow{6}{*}{DiffEdit (SD 2.1)} & Stock VAE (SD 2.1) & 0.031 & 0.061 & 0.082 & 0.362 & 0.639 & 0.735 & 0.649 & 0.348 & 0.205 & 15.533 & 12.346 & 10.944 \\
 & Soft-limit control & \textbf{0.028} & \textbf{0.056} & 0.075 & 0.356 & 0.633 & 0.738 & 0.665 & 0.373 & 0.223 & 15.846 & \underline{12.686} & 11.297 \\
 & REED-VAE (paper) & \underline{0.030} & 0.070 & 0.080 & 0.300 & 0.550 & 0.680 & 0.690 & 0.480 & 0.400 & \textbf{16.240} & \textbf{12.910} & 11.440 \\
 & Ours w/o BP & \textbf{0.028} & 0.058 & \underline{0.070} & \underline{0.294} & \underline{0.473} & \underline{0.551} & \underline{0.713} & \underline{0.530} & \underline{0.450} & 15.804 & 12.502 & \underline{11.627} \\
 & \textbf{Ours} & \textbf{0.028} & \underline{0.057} & \textbf{0.069} & \textbf{0.289} & \textbf{0.444} & \textbf{0.502} & \textbf{0.715} & \textbf{0.548} & \textbf{0.482} & \underline{15.871} & 12.573 & \textbf{11.691} \\
 & \textcolor{gray}{\emph{latent pass-through}} & \textcolor{gray}{0.031} & \textcolor{gray}{0.066} & \textcolor{gray}{0.084} & \textcolor{gray}{0.292} & \textcolor{gray}{0.467} & \textcolor{gray}{0.544} & \textcolor{gray}{0.704} & \textcolor{gray}{0.518} & \textcolor{gray}{0.435} & \textcolor{gray}{15.481} & \textcolor{gray}{11.964} & \textcolor{gray}{10.878} \\
\midrule
\multirow{5}{*}{Paint-by-Example} & Stock VAE (SD 1.5) & 0.019 & 0.044 & 0.068 & 0.254 & 0.596 & 0.708 & 0.671 & 0.329 & 0.223 & 18.723 & 13.914 & 11.829 \\
 & Soft-limit control & 0.018 & 0.042 & 0.069 & 0.252 & 0.597 & 0.712 & 0.688 & 0.352 & 0.238 & 18.862 & 14.044 & 11.793 \\
 & REED-VAE (paper) & 0.020 & 0.030 & 0.040 & 0.200 & 0.440 & 0.590 & 0.770 & 0.610 & 0.540 & 19.620 & 16.190 & 14.430 \\
 & Ours w/o BP & \underline{0.016} & \underline{0.022} & \underline{0.027} & \underline{0.149} & \underline{0.273} & \underline{0.378} & \underline{0.802} & \underline{0.666} & \underline{0.585} & \underline{20.573} & \underline{18.010} & \underline{16.800} \\
 & \textbf{Ours} & \textbf{0.015} & \textbf{0.019} & \textbf{0.021} & \textbf{0.129} & \textbf{0.187} & \textbf{0.228} & \textbf{0.837} & \textbf{0.762} & \textbf{0.714} & \textbf{20.899} & \textbf{18.951} & \textbf{18.127} \\
\midrule
\multirow{5}{*}{SD Inpainting} & Stock VAE (SD 1.5) & 0.010 & 0.064 & 0.115 & 0.284 & 0.698 & 0.793 & 0.685 & 0.224 & 0.143 & 20.775 & 12.153 & 9.528 \\
 & Soft-limit control & 0.010 & 0.062 & 0.112 & 0.277 & 0.697 & 0.785 & 0.709 & 0.263 & 0.182 & 21.010 & 12.290 & 9.573 \\
 & REED-VAE (paper) & 0.010 & 0.030 & 0.050 & 0.170 & 0.470 & 0.650 & 0.800 & \underline{0.570} & 0.410 & \underline{23.140} & 16.660 & 13.610 \\
 & Ours w/o BP & \underline{0.008} & \underline{0.025} & \underline{0.041} & \underline{0.143} & \underline{0.403} & \underline{0.534} & \underline{0.834} & 0.568 & \underline{0.418} & 23.059 & \underline{17.170} & \underline{14.665} \\
 & \textbf{Ours} & \textbf{0.007} & \textbf{0.013} & \textbf{0.018} & \textbf{0.095} & \textbf{0.168} & \textbf{0.215} & \textbf{0.889} & \textbf{0.820} & \textbf{0.776} & \textbf{24.205} & \textbf{19.999} & \textbf{18.350} \\
\bottomrule
\end{tabular}}
\par
\caption{\footnotesize \textbf{Iterative editing on the REED benchmark} (179 ImagenHub examples, 25 rounds alternating an edit and its reverse or, for SD Inpainting, repeating the same inpainting, as in REED, 8-bit hand-off between rounds, metrics of round $k$ against round 1). Bold: best, underlined: second best of the five rows that hand off images (stock, control, REED-VAE, Ours w/o BP, Ours). REED rows are from its paper (two decimals, padded). \emph{Soft-limit control}: the vanilla decoder output passed through our soft limit (identity up to $\pm0.82$, squeezed into $\pm0.92$), encoder unchanged. \emph{Ours w/o BP}: the editor's frozen VAE with LAION PCA coordinates, $k=4$, $c=29.8$, soft limit and pixel map; no training. \emph{Ours}: the wrapped codec with latent back-projection of the editor's output latent onto the round's input latent; a per-turn noise estimate with the band profile calibrated on 100 external no-op chains, for the instruction editors (InstructPix2Pix, MagicBrush), and one per-turn noise level shared across all bands (no no-op instruction to calibrate on) for the others. \emph{Latent pass-through} (gray) never decodes between rounds and is the drift-free codec reference; it is unavailable for Paint-by-Example and SD Inpainting, which condition on pixels. DiffEdit runs on SD~2.1 with its own VAE, which our codec wraps; all other editors use SD~1.5.}
\label{tab:reed-editing-full}
\end{table*}

\subsection{VAE-LFA benchmark}
\label{app:vaelfa-full}

\begin{figure*}
\centering
\includegraphics[width=\textwidth]{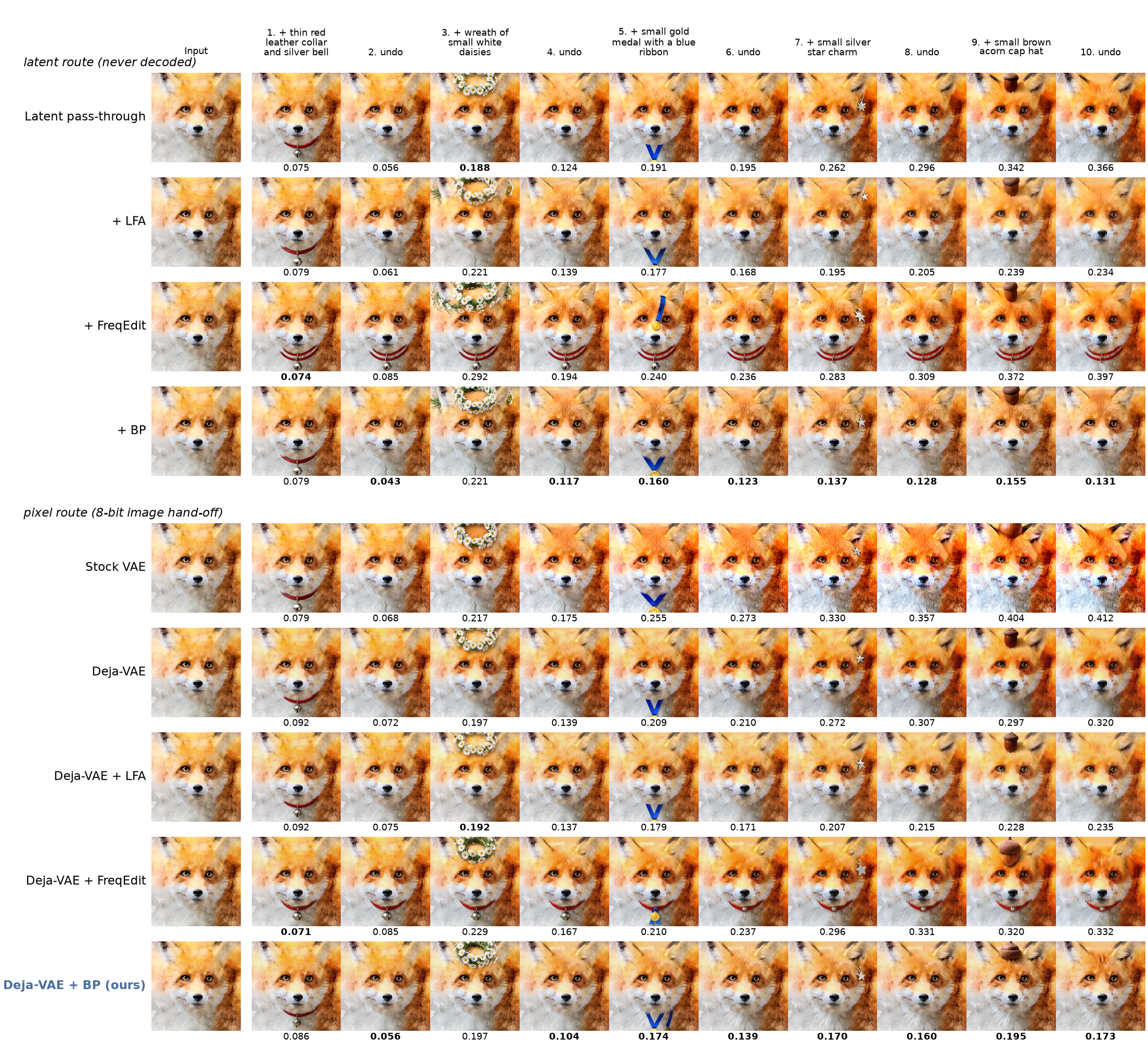}
\par
\caption{\footnotesize \textbf{A cycle-editing chain, every method} (VAE-LFA's cycle protocol, FLUX.2 [klein], seed 0). Odd rounds apply an edit and even rounds undo it: (1) ``Add a thin red leather collar with a small silver bell around the fox's neck.'', (3) ``Place a wreath of small white daisies on top of the fox's head.'', (5) ``Pin a small gold medal with a blue ribbon to the fox's chest fur.'', (7) ``Clip a small silver star charm onto the fur of the fox's right ear.'', (9) ``Balance a small brown acorn cap like a hat on the fox's head.''; each following round asks to remove the object again. Under each panel: LPIPS to the input; bold: lowest within the route in that round. BP uses a per-turn noise estimate with a calibrated band profile. FreqEdit never removes the collar, in either route. With \method every edit appears and is undone, and after every undo the image is the closest to the input in the pixel route.}
\label{fig:cycle-chain}
\end{figure*}

Tables~\ref{tab:vaelfa-noop} and~\ref{tab:vaelfa-cycle} give every no-op and cycle-editing result behind Table~\ref{tab:vaelfa-summary}; Figure~\ref{fig:cycle-chain} shows one cycle chain.

\begin{table*}
\centering
\small
\setlength{\tabcolsep}{4pt}
\resizebox{\textwidth}{!}{%
\begin{tabular}{@{}l ccccc @{\hspace{11pt}} ccccc@{}}
\toprule
& \multicolumn{5}{c}{FLUX.2 [klein]} & \multicolumn{5}{c}{SD3-UE} \\
\cmidrule(lr){2-6}\cmidrule(l){7-11}
Method & LPIPS$\downarrow$ & L1$\downarrow$ & SSIM$\uparrow$ & PSNR$\uparrow$ & $\Delta E\downarrow$ & LPIPS$\downarrow$ & L1$\downarrow$ & SSIM$\uparrow$ & PSNR$\uparrow$ & $\Delta E\downarrow$ \\
\midrule
\textcolor{gray}{Released SD3-UE code} & -- & -- & -- & -- & -- & \textcolor{gray}{.88} & \textcolor{gray}{.46} & \textcolor{gray}{.02} & \textcolor{gray}{5.9} & \textcolor{gray}{55.2} \\
\textcolor{gray}{SD3-UE / VAE-LFA, released code} & -- & -- & -- & -- & -- & \textcolor{gray}{.68} & \textcolor{gray}{.30} & \textcolor{gray}{.47} & \textcolor{gray}{8.8} & \textcolor{gray}{45.6} \\
\midrule
Stock & .41 & .13 & .77 & 15.8 & 18.7 & .48 & .15 & .73 & 14.4 & 25.1 \\
Ours w/o wrapper & .32 & .08 & .83 & 18.5 & 14.1 & \underline{.42} & \underline{.10} & .81 & \underline{17.8} & 15.9 \\
VAE-LFA & .33 & .08 & .83 & 18.7 & 11.8 & .46 & .13 & .73 & 15.4 & 19.6 \\
FreqEdit & .38 & .12 & .83 & 16.7 & 17.3 & .43 & .12 & .80 & 15.9 & 20.9 \\
\midrule
Ours w/o BP & .35 & .12 & .81 & 16.6 & 17.3 & .47 & .14 & .75 & 14.5 & 25.0 \\
\textbf{Ours} & \underline{.16} & \underline{.07} & .87 & \underline{20.3} & \underline{9.0} & \textbf{.28} & \textbf{.09} & \textbf{.86} & \textbf{19.2} & \underline{14.6} \\
VAE-LFA (wrapped) & .28 & .08 & .83 & 18.9 & 11.1 & .47 & .13 & .73 & 15.1 & 20.2 \\
FreqEdit (wrapped) & .30 & .10 & \underline{.88} & 17.9 & 15.5 & \underline{.42} & .12 & .81 & 15.8 & 21.6 \\
\midrule
Latent: uncorrected & .34 & .11 & .81 & 16.9 & 16.3 & .47 & .16 & .73 & 14.0 & 25.5 \\
Latent: BP & \textbf{.15} & \textbf{.06} & .87 & \textbf{20.9} & \textbf{8.1} & \textbf{.28} & \textbf{.09} & \underline{.85} & \textbf{19.2} & \textbf{14.4} \\
Latent: VAE-LFA (white-box) & .27 & .08 & .84 & 19.1 & 10.6 & .47 & .13 & .72 & 14.7 & 20.5 \\
Latent: FreqEdit & .30 & .10 & \textbf{.89} & 18.4 & 14.4 & \underline{.42} & .13 & .81 & 15.4 & 21.7 \\
\bottomrule
\end{tabular}}
\par
\caption{\footnotesize \textbf{No-op editing on the 120 images of VAE-LFA} (their protocol and metric code: ten rounds of ``Keep the image unchanged.''; round 10 against the input, mean over all 120 images and seeds 0 and 10). Rows combine a codec route (stock VAE or the \method codec with an 8-bit image hand-off, or latent pass-through) with one drift correction. BP uses the current round's noise level and a band profile calibrated once per editor on 8 external images. SD3-UE rows use our fix of the decoding and latent pass-through in VAE-LFA's released SD3-UE code; gray: that code, run unmodified. PSNR (dB) and $\Delta E$ (mean CIELAB color difference over all pixels) are ours, computed on the same image pairs as their LPIPS, L1 and SSIM. Bold: best; underlined: second best per editor and column; ties at displayed precision share ranks. Gray diagnostics are excluded. Per-category LPIPS, L1 and SSIM: Table~\ref{tab:vaelfa-noop-full}.}
\label{tab:vaelfa-noop}
\end{table*}

\begin{table*}
\centering
\small
\setlength{\tabcolsep}{4pt}
\resizebox{\textwidth}{!}{%
\begin{tabular}{@{}l ccc@{\hspace{7pt}}ccc @{\hspace{11pt}} ccc@{\hspace{7pt}}ccc@{}}
\toprule
& \multicolumn{6}{c}{FLUX.2 [klein]} & \multicolumn{6}{c}{SD3-UE} \\
\cmidrule(lr){2-7}\cmidrule(l){8-13}
& \multicolumn{3}{c}{Drift (round 10)} & \multicolumn{3}{c}{Edits (\%)} & \multicolumn{3}{c}{Drift (round 10)} & \multicolumn{3}{c}{Edits (\%)} \\
\cmidrule(lr){2-4}\cmidrule(lr){5-7}\cmidrule(lr){8-10}\cmidrule(l){11-13}
Method & LPIPS$\downarrow$ & L1$\downarrow$ & SSIM$\uparrow$ & Edit$\uparrow$ & Undo$\uparrow$ & Cycle$\uparrow$ & LPIPS$\downarrow$ & L1$\downarrow$ & SSIM$\uparrow$ & Edit$\uparrow$ & Undo$\uparrow$ & Cycle$\uparrow$ \\
\midrule
\textcolor{gray}{Released SD3-UE code} & -- & -- & -- & -- & -- & -- & \textcolor{gray}{.86} & \textcolor{gray}{.45} & \textcolor{gray}{.03} & -- & -- & -- \\
\textcolor{gray}{SD3-UE / VAE-LFA, released code} & -- & -- & -- & -- & -- & -- & \textcolor{gray}{.70} & \textcolor{gray}{.32} & \textcolor{gray}{.42} & -- & -- & -- \\
\midrule
Stock & .48 & .16 & .63 & \textbf{99.2} & \textbf{89.1} & \textbf{89.0} & .60 & .21 & .58 & \underline{93.5} & 58.7 & 55.2 \\
Ours w/o wrapper & .41 & .13 & .68 & 98.7 & 88.3 & 88.0 & .55 & \textbf{.15} & \underline{.67} & 84.3 & 57.8 & 51.4 \\
VAE-LFA & .43 & .13 & .65 & \underline{99.0} & 87.7 & 87.2 & .58 & .19 & .54 & \textbf{93.7} & \textbf{61.8} & \textbf{58.8} \\
FreqEdit & .46 & .15 & .67 & 98.7 & 73.8 & 73.6 & .54 & .18 & .67 & 88.4 & 49.8 & 45.1 \\
\midrule
Ours w/o BP & .44 & .15 & .65 & \textbf{99.2} & 88.8 & \underline{88.6} & .60 & .21 & .59 & 91.8 & 58.7 & 54.7 \\
\textbf{Ours} & \underline{.31} & \underline{.12} & \underline{.71} & 98.6 & 87.5 & 87.0 & \textbf{.52} & \underline{.17} & .66 & 89.9 & 57.8 & 54.0 \\
VAE-LFA (wrapped) & .40 & .13 & .66 & 98.8 & 87.6 & 87.1 & .58 & .19 & .54 & 92.2 & \underline{59.6} & \underline{55.7} \\
FreqEdit (wrapped) & .42 & .14 & .71 & 98.4 & 74.1 & 73.9 & .54 & .17 & \textbf{.68} & 85.8 & 50.2 & 44.2 \\
\midrule
Latent: uncorrected & .43 & .15 & .66 & \textbf{99.2} & 88.2 & 88.1 & .60 & .22 & .58 & 92.7 & 57.3 & 53.8 \\
Latent: BP & \textbf{.31} & \textbf{.12} & .70 & 98.5 & \underline{88.9} & 88.2 & \underline{.52} & .17 & .65 & 90.5 & 59.2 & 54.7 \\
Latent: VAE-LFA (white-box) & .39 & .13 & .67 & 98.9 & 87.5 & 87.2 & .58 & .20 & .53 & 92.8 & 58.2 & 54.8 \\
Latent: FreqEdit & .41 & .14 & \textbf{.72} & 98.5 & 73.5 & 73.1 & .54 & .19 & .67 & 85.8 & 49.7 & 43.7 \\
\bottomrule
\end{tabular}}
\par
\caption{\footnotesize \textbf{Cycle editing on the 120 images of VAE-LFA} (their protocol: five local edit / undo pairs per image, 10 rounds; mean over all 120 images and seeds 0 and 10). Rows combine a codec route (stock VAE or the \method codec with an 8-bit image hand-off, or latent pass-through) with one drift correction. Drift: round 10 against the input, with VAE-LFA's metric code. Edits: share (\%) of the 1,200 edit/undo pairs (120 images $\times$ 5 pairs $\times$ 2 seeds) whose edit (Edit), undo (Undo), or both (Cycle) EdiVal's evaluation~\cite{edival} verifies (Appendix~\ref{app:edit-verification}). BP uses the current round's noise level and a band profile calibrated once per editor on 8 external images. SD3-UE rows use our fix of the decoding and latent pass-through in VAE-LFA's released SD3-UE code; gray: that code, run unmodified. Bold: best, underlined: second best per editor (drift ranked on unrounded values). Per-category drift: Table~\ref{tab:vaelfa-cycle-full}; an example chain: Figure~\ref{fig:cycle-chain}.}
\label{tab:vaelfa-cycle}
\end{table*}

Tables~\ref{tab:vaelfa-noop-full} and~\ref{tab:vaelfa-cycle-full} give the per-category results behind Tables~\ref{tab:vaelfa-noop} and~\ref{tab:vaelfa-cycle}, in the format of VAE-LFA's Table~1.

\begin{table*}
\centering
\small
\resizebox{\textwidth}{!}{%
\begin{tabular}{@{}l ccc ccc ccc ccc ccc ccc ccc@{}}
\toprule
& \multicolumn{9}{c}{photograph} & \multicolumn{9}{c}{illustration} & \multicolumn{3}{c}{\multirow{2}{*}{all (120)}} \\
\cmidrule(lr){2-10}\cmidrule(lr){11-19}
& \multicolumn{3}{c}{creature} & \multicolumn{3}{c}{architecture} & \multicolumn{3}{c}{scenery} & \multicolumn{3}{c}{creature} & \multicolumn{3}{c}{architecture} & \multicolumn{3}{c}{scenery} & & & \\
\cmidrule(lr){2-4}\cmidrule(lr){5-7}\cmidrule(lr){8-10}\cmidrule(lr){11-13}\cmidrule(lr){14-16}\cmidrule(lr){17-19}\cmidrule(l){20-22}
Method & LPIPS$\downarrow$ & L1$\downarrow$ & SSIM$\uparrow$ & LPIPS$\downarrow$ & L1$\downarrow$ & SSIM$\uparrow$ & LPIPS$\downarrow$ & L1$\downarrow$ & SSIM$\uparrow$ & LPIPS$\downarrow$ & L1$\downarrow$ & SSIM$\uparrow$ & LPIPS$\downarrow$ & L1$\downarrow$ & SSIM$\uparrow$ & LPIPS$\downarrow$ & L1$\downarrow$ & SSIM$\uparrow$ & LPIPS$\downarrow$ & L1$\downarrow$ & SSIM$\uparrow$ \\
\midrule
\multicolumn{22}{@{}l}{\emph{No-op, round 10}} \\
\midrule
\multicolumn{22}{@{}l}{\textbf{FLUX.2 [klein]}} \\
Stock & .32 & .09 & .85 & .38 & .14 & .73 & .40 & .12 & .78 & .34 & .11 & .82 & .49 & .16 & .73 & .52 & .15 & .73 & .41 & .13 & .77 \\
Ours w/o wrapper & .24 & .05 & \underline{.92} & .31 & .10 & .78 & .34 & \underline{.08} & .83 & .23 & \underline{.07} & .88 & .37 & .10 & .78 & .44 & .09 & .80 & .32 & .08 & .83 \\
VAE-LFA & .24 & .05 & .91 & .31 & .10 & .77 & .33 & \underline{.08} & .82 & .28 & .08 & .87 & .39 & .10 & .79 & .42 & \underline{.08} & .82 & .33 & .08 & .83 \\
FreqEdit & .32 & .08 & .87 & .33 & .12 & .82 & .38 & .11 & .85 & .33 & .11 & .84 & .45 & .14 & \underline{.82} & .50 & .14 & .79 & .38 & .12 & .83 \\
\cmidrule(l){2-22}
Ours w/o BP & .27 & .08 & .89 & .35 & .13 & .76 & .34 & .11 & .81 & .33 & .12 & .84 & .38 & .14 & .77 & .42 & .13 & .79 & .35 & .12 & .81 \\
\textbf{Ours} & \underline{.09} & \underline{.04} & \textbf{.95} & \underline{.17} & \textbf{.07} & .82 & \underline{.16} & \textbf{.07} & .86 & \underline{.15} & \underline{.07} & \underline{.90} & \underline{.19} & \underline{.09} & \underline{.82} & \underline{.22} & \textbf{.07} & \underline{.86} & \underline{.16} & \underline{.07} & .87 \\
VAE-LFA (wrapped) & .20 & .05 & .91 & .27 & \underline{.09} & .78 & .27 & \underline{.08} & .83 & .26 & .08 & .87 & .31 & .10 & .79 & .34 & \underline{.08} & .83 & .28 & .08 & .83 \\
FreqEdit (wrapped) & .24 & .07 & \underline{.92} & .30 & .11 & \underline{.86} & .29 & .10 & \underline{.88} & .31 & .11 & .87 & .32 & .12 & \textbf{.87} & .37 & .12 & \textbf{.87} & .30 & .10 & \underline{.88} \\
\cmidrule(l){2-22}
Latent: uncorrected & .27 & .07 & .90 & .33 & .12 & .77 & .33 & .11 & .81 & .31 & .10 & .86 & .38 & .13 & .77 & .40 & .13 & .79 & .34 & .11 & .81 \\
Latent: BP & \textbf{.07} & \textbf{.03} & \textbf{.95} & \textbf{.15} & \textbf{.07} & .83 & \textbf{.14} & \textbf{.07} & .86 & \textbf{.13} & \textbf{.05} & \textbf{.91} & \textbf{.18} & \textbf{.08} & \underline{.82} & \textbf{.21} & \textbf{.07} & \underline{.86} & \textbf{.15} & \textbf{.06} & .87 \\
Latent: VAE-LFA (white-box) & .19 & .05 & \underline{.92} & .26 & \underline{.09} & .79 & .26 & \underline{.08} & .83 & .24 & \underline{.07} & .88 & .31 & \underline{.09} & .79 & .33 & \underline{.08} & .83 & .27 & .08 & .84 \\
Latent: FreqEdit & .25 & .07 & \underline{.92} & .28 & .10 & \textbf{.87} & .29 & .09 & \textbf{.89} & .26 & .09 & \underline{.90} & .32 & .12 & \textbf{.87} & .37 & .11 & \textbf{.87} & .30 & .10 & \textbf{.89} \\
\midrule
\multicolumn{22}{@{}l}{\textbf{SD3-UE}} \\
\textcolor{gray}{Released SD3-UE code$^\dagger$} & \textcolor{gray}{.86} & \textcolor{gray}{.42} & \textcolor{gray}{.03} & \textcolor{gray}{.89} & \textcolor{gray}{.48} & \textcolor{gray}{.00} & \textcolor{gray}{.88} & \textcolor{gray}{.41} & \textcolor{gray}{.01} & \textcolor{gray}{.82} & \textcolor{gray}{.46} & \textcolor{gray}{.08} & \textcolor{gray}{.89} & \textcolor{gray}{.52} & \textcolor{gray}{.02} & \textcolor{gray}{.92} & \textcolor{gray}{.45} & \textcolor{gray}{.00} & \textcolor{gray}{.88} & \textcolor{gray}{.46} & \textcolor{gray}{.02} \\
\textcolor{gray}{SD3-UE / VAE-LFA, released code$^\dagger$} & \textcolor{gray}{.65} & \textcolor{gray}{.26} & \textcolor{gray}{.50} & \textcolor{gray}{.68} & \textcolor{gray}{.32} & \textcolor{gray}{.42} & \textcolor{gray}{.69} & \textcolor{gray}{.30} & \textcolor{gray}{.48} & \textcolor{gray}{.61} & \textcolor{gray}{.27} & \textcolor{gray}{.55} & \textcolor{gray}{.69} & \textcolor{gray}{.33} & \textcolor{gray}{.45} & \textcolor{gray}{.74} & \textcolor{gray}{.32} & \textcolor{gray}{.43} & \textcolor{gray}{.68} & \textcolor{gray}{.30} & \textcolor{gray}{.47} \\
\cmidrule(l){2-22}
Stock & .44 & .12 & .76 & .42 & .15 & .73 & .52 & .16 & .71 & .42 & .13 & .79 & .50 & .17 & .69 & .57 & .17 & .71 & .48 & .15 & .73 \\
Ours w/o wrapper & .38 & \underline{.08} & \underline{.84} & .40 & .11 & .77 & \underline{.44} & \textbf{.11} & \textbf{.80} & .37 & .08 & \underline{.89} & .45 & \underline{.11} & .78 & .50 & .10 & .81 & \underline{.42} & \underline{.10} & .81 \\
VAE-LFA & .42 & .10 & .78 & .41 & .12 & .72 & .50 & .14 & .69 & .42 & .11 & .80 & .48 & .14 & .70 & .55 & .14 & .69 & .46 & .13 & .73 \\
FreqEdit & .41 & .10 & .82 & .40 & .13 & .77 & .47 & \underline{.13} & .77 & .37 & .11 & .84 & .44 & .13 & .78 & .50 & .13 & .79 & .43 & .12 & .80 \\
\cmidrule(l){2-22}
Ours w/o BP & .44 & .12 & .79 & .41 & .15 & .73 & .53 & .16 & .70 & .40 & .13 & .80 & .49 & .17 & .71 & .55 & .15 & .74 & .47 & .14 & .75 \\
\textbf{Ours} & \textbf{.24} & \textbf{.07} & \textbf{.86} & \textbf{.22} & \textbf{.08} & \textbf{.86} & \textbf{.36} & \textbf{.11} & \textbf{.80} & \underline{.21} & \underline{.07} & \textbf{.92} & \textbf{.32} & \textbf{.10} & \textbf{.82} & \textbf{.33} & \textbf{.08} & \textbf{.88} & \textbf{.28} & \textbf{.09} & \textbf{.86} \\
VAE-LFA (wrapped) & .44 & .10 & .78 & .42 & .12 & .72 & .51 & .14 & .70 & .41 & .12 & .78 & .47 & .15 & .70 & .54 & .14 & .71 & .47 & .13 & .73 \\
FreqEdit (wrapped) & .42 & .10 & \underline{.84} & .40 & .12 & .80 & .47 & \underline{.13} & \underline{.79} & .34 & .11 & .85 & .43 & .14 & \underline{.79} & .49 & .13 & .81 & \underline{.42} & .12 & .81 \\
\cmidrule(l){2-22}
Latent: uncorrected & .43 & .13 & .78 & .41 & .16 & .72 & .52 & .17 & .70 & .42 & .14 & .79 & .49 & .18 & .69 & .55 & .17 & .73 & .47 & .16 & .73 \\
Latent: BP & \underline{.25} & \underline{.08} & \underline{.84} & \underline{.23} & \underline{.09} & \underline{.84} & \textbf{.36} & \textbf{.11} & \underline{.79} & \textbf{.19} & \textbf{.06} & \textbf{.92} & \textbf{.32} & \textbf{.10} & \textbf{.82} & \underline{.34} & \underline{.09} & \underline{.87} & \textbf{.28} & \textbf{.09} & \underline{.85} \\
Latent: VAE-LFA (white-box) & .44 & .11 & .77 & .42 & .13 & .70 & .51 & .15 & .69 & .42 & .13 & .77 & .48 & .15 & .69 & .54 & .15 & .71 & .47 & .13 & .72 \\
Latent: FreqEdit & .41 & .11 & .83 & .39 & .13 & .80 & .47 & .14 & .78 & .34 & .11 & .85 & \underline{.42} & .15 & \underline{.79} & .48 & .14 & .81 & \underline{.42} & .13 & .81 \\
\bottomrule
\end{tabular}}
\par
\caption{\footnotesize \textbf{No-op editing on the 120 images of VAE-LFA} (their protocol and metric code: ten rounds of ``Keep the image unchanged.'', round 10 against the input, mean of seeds 0 and 10). Rows combine a codec route between rounds (stock VAE or the \method codec with an 8-bit image hand-off, or latent pass-through without decoding) with one drift correction (none, BP, LFA, FreqEdit). BP: latent back-projection using the current round's noise level and a band profile calibrated once per editor on 8 external images. SD3-UE rows use our fix of the decoding and latent pass-through in VAE-LFA's released SD3-UE code; $^\dagger$: that code, run unmodified. Bold/underlined: best/second per editor and column. Ties at displayed precision share ranks. Rankings exclude gray diagnostics.}
\label{tab:vaelfa-noop-full}
\end{table*}

\begin{table*}
\centering
\small
\resizebox{\textwidth}{!}{%
\begin{tabular}{@{}l ccc ccc ccc ccc ccc ccc ccc@{}}
\toprule
& \multicolumn{9}{c}{photograph} & \multicolumn{9}{c}{illustration} & \multicolumn{3}{c}{\multirow{2}{*}{all (120)}} \\
\cmidrule(lr){2-10}\cmidrule(lr){11-19}
& \multicolumn{3}{c}{creature} & \multicolumn{3}{c}{architecture} & \multicolumn{3}{c}{scenery} & \multicolumn{3}{c}{creature} & \multicolumn{3}{c}{architecture} & \multicolumn{3}{c}{scenery} & & & \\
\cmidrule(lr){2-4}\cmidrule(lr){5-7}\cmidrule(lr){8-10}\cmidrule(lr){11-13}\cmidrule(lr){14-16}\cmidrule(lr){17-19}\cmidrule(l){20-22}
Method & LPIPS$\downarrow$ & L1$\downarrow$ & SSIM$\uparrow$ & LPIPS$\downarrow$ & L1$\downarrow$ & SSIM$\uparrow$ & LPIPS$\downarrow$ & L1$\downarrow$ & SSIM$\uparrow$ & LPIPS$\downarrow$ & L1$\downarrow$ & SSIM$\uparrow$ & LPIPS$\downarrow$ & L1$\downarrow$ & SSIM$\uparrow$ & LPIPS$\downarrow$ & L1$\downarrow$ & SSIM$\uparrow$ & LPIPS$\downarrow$ & L1$\downarrow$ & SSIM$\uparrow$ \\
\midrule
\multicolumn{22}{@{}l}{\emph{Cycle, round 10}} \\
\midrule
\multicolumn{22}{@{}l}{\textbf{FLUX.2 [klein]}} \\
Stock & .40 & .11 & .71 & .43 & .17 & .62 & .49 & .16 & .62 & .41 & .14 & .71 & .54 & .18 & .58 & .62 & .19 & .55 & .48 & .16 & .63 \\
Ours w/o wrapper & .33 & \underline{.09} & .74 & .37 & .13 & .66 & .44 & \underline{.14} & .67 & .31 & \underline{.12} & \underline{.76} & .45 & .15 & .63 & .55 & \underline{.16} & .60 & .41 & .13 & .68 \\
VAE-LFA & .35 & .10 & .72 & .37 & .13 & .65 & .46 & \underline{.14} & .62 & .37 & .13 & .73 & .46 & \underline{.14} & .61 & .55 & \underline{.16} & .59 & .43 & .13 & .65 \\
FreqEdit & .39 & .11 & .71 & .42 & .15 & .68 & .50 & .17 & .65 & .39 & .13 & .73 & .51 & .17 & \underline{.66} & .59 & .18 & .62 & .46 & .15 & .67 \\
\cmidrule(l){2-22}
Ours w/o BP & .36 & .11 & .72 & .41 & .15 & .65 & .45 & .16 & .62 & .39 & .14 & .73 & .47 & .16 & .60 & .57 & .18 & .59 & .44 & .15 & .65 \\
\textbf{Ours} & \underline{.23} & \textbf{.08} & \textbf{.77} & \underline{.26} & \underline{.11} & .70 & \underline{.34} & \textbf{.13} & \textbf{.70} & \underline{.26} & \textbf{.11} & \textbf{.78} & \textbf{.34} & \textbf{.13} & \underline{.66} & \underline{.46} & \textbf{.15} & .64 & \underline{.31} & \underline{.12} & \underline{.71} \\
VAE-LFA (wrapped) & .32 & .10 & .71 & .35 & .12 & .65 & .41 & \textbf{.13} & .65 & .35 & .13 & .73 & \underline{.42} & \underline{.14} & .61 & .53 & \underline{.16} & .59 & .40 & .13 & .66 \\
FreqEdit (wrapped) & .33 & .10 & .74 & .39 & .14 & \underline{.72} & .44 & .15 & .68 & .37 & .13 & \underline{.76} & \underline{.42} & .15 & \textbf{.69} & .54 & \underline{.16} & \underline{.65} & .42 & .14 & .71 \\
\cmidrule(l){2-22}
Latent: uncorrected & .35 & .11 & .73 & .40 & .15 & .66 & .45 & .16 & .64 & .37 & .13 & .74 & .47 & .16 & .61 & .58 & .18 & .59 & .43 & .15 & .66 \\
Latent: BP & \textbf{.22} & \textbf{.08} & \underline{.76} & \textbf{.25} & \textbf{.10} & .71 & \textbf{.33} & \textbf{.13} & \underline{.69} & \textbf{.25} & \textbf{.11} & \textbf{.78} & \textbf{.34} & \textbf{.13} & \underline{.66} & \textbf{.45} & \textbf{.15} & .63 & \textbf{.31} & \textbf{.12} & .70 \\
Latent: VAE-LFA (white-box) & .32 & .10 & .71 & .33 & .12 & .67 & .41 & \textbf{.13} & .66 & .33 & \underline{.12} & .75 & \underline{.42} & \underline{.14} & .61 & .53 & \underline{.16} & .59 & .39 & .13 & .67 \\
Latent: FreqEdit & .32 & .10 & .75 & .38 & .13 & \textbf{.73} & .44 & .15 & \underline{.69} & .35 & \underline{.12} & \textbf{.78} & \underline{.42} & .15 & \textbf{.69} & .54 & \underline{.16} & \textbf{.66} & .41 & .14 & \textbf{.72} \\
\midrule
\multicolumn{22}{@{}l}{\textbf{SD3-UE}} \\
\textcolor{gray}{Released SD3-UE code$^\dagger$} & \textcolor{gray}{.84} & \textcolor{gray}{.41} & \textcolor{gray}{.04} & \textcolor{gray}{.87} & \textcolor{gray}{.48} & \textcolor{gray}{.00} & \textcolor{gray}{.88} & \textcolor{gray}{.41} & \textcolor{gray}{.02} & \textcolor{gray}{.80} & \textcolor{gray}{.45} & \textcolor{gray}{.10} & \textcolor{gray}{.86} & \textcolor{gray}{.50} & \textcolor{gray}{.04} & \textcolor{gray}{.90} & \textcolor{gray}{.44} & \textcolor{gray}{.01} & \textcolor{gray}{.86} & \textcolor{gray}{.45} & \textcolor{gray}{.03} \\
\textcolor{gray}{SD3-UE / VAE-LFA, released code$^\dagger$} & \textcolor{gray}{.68} & \textcolor{gray}{.28} & \textcolor{gray}{.44} & \textcolor{gray}{.71} & \textcolor{gray}{.33} & \textcolor{gray}{.37} & \textcolor{gray}{.71} & \textcolor{gray}{.31} & \textcolor{gray}{.43} & \textcolor{gray}{.65} & \textcolor{gray}{.29} & \textcolor{gray}{.50} & \textcolor{gray}{.71} & \textcolor{gray}{.35} & \textcolor{gray}{.39} & \textcolor{gray}{.75} & \textcolor{gray}{.34} & \textcolor{gray}{.39} & \textcolor{gray}{.70} & \textcolor{gray}{.32} & \textcolor{gray}{.42} \\
\cmidrule(l){2-22}
Stock & .53 & .17 & .64 & .56 & .21 & .56 & .63 & .21 & .58 & .58 & .20 & .62 & .64 & .24 & .53 & .69 & .23 & .55 & .60 & .21 & .58 \\
Ours w/o wrapper & \underline{.47} & \textbf{.12} & \textbf{.73} & .54 & \textbf{.16} & .61 & \underline{.56} & \textbf{.15} & \textbf{.68} & .53 & \textbf{.15} & .72 & .57 & \textbf{.16} & .64 & \underline{.62} & \textbf{.16} & \underline{.65} & .55 & \textbf{.15} & \underline{.67} \\
VAE-LFA & .50 & .14 & .64 & .52 & .18 & .54 & .60 & .19 & .55 & .56 & .19 & .58 & .62 & .23 & .46 & .66 & .21 & .49 & .58 & .19 & .54 \\
FreqEdit & .51 & .15 & .71 & .52 & .18 & .62 & .58 & .19 & .64 & \underline{.46} & \textbf{.15} & \textbf{.75} & \underline{.56} & .20 & \underline{.65} & .63 & .20 & .63 & .54 & .18 & .67 \\
\cmidrule(l){2-22}
Ours w/o BP & .52 & .16 & .66 & .55 & .20 & .57 & .62 & .21 & .59 & .57 & .20 & .62 & .63 & .24 & .53 & .68 & .22 & .56 & .60 & .21 & .59 \\
\textbf{Ours} & \textbf{.44} & \underline{.13} & \underline{.72} & \textbf{.49} & .18 & .60 & \textbf{.55} & \underline{.17} & \underline{.67} & .50 & \underline{.16} & .70 & .57 & \underline{.19} & .60 & \textbf{.59} & \underline{.17} & \textbf{.66} & \textbf{.52} & \underline{.17} & .66 \\
VAE-LFA (wrapped) & .51 & .14 & .63 & .53 & .18 & .54 & .61 & .20 & .54 & .55 & .19 & .58 & .62 & .23 & .46 & .65 & .21 & .50 & .58 & .19 & .54 \\
FreqEdit (wrapped) & .51 & .14 & \textbf{.73} & .51 & \underline{.17} & \textbf{.66} & .59 & .19 & .65 & \textbf{.45} & \textbf{.15} & \textbf{.75} & \textbf{.55} & .20 & \textbf{.66} & \underline{.62} & .19 & \underline{.65} & .54 & .17 & \textbf{.68} \\
\cmidrule(l){2-22}
Latent: uncorrected & .53 & .18 & .65 & .55 & .21 & .57 & .62 & .22 & .58 & .57 & .21 & .62 & .63 & .25 & .52 & .69 & .23 & .55 & .60 & .22 & .58 \\
Latent: BP & \textbf{.44} & .14 & .70 & \textbf{.49} & .18 & .60 & \textbf{.55} & .18 & .66 & .51 & .17 & .69 & \underline{.56} & .20 & .61 & \textbf{.59} & .18 & .64 & \underline{.52} & .17 & .65 \\
Latent: VAE-LFA (white-box) & .51 & .16 & .62 & .53 & .19 & .53 & .60 & .20 & .53 & .56 & .21 & .56 & .62 & .24 & .45 & .66 & .22 & .48 & .58 & .20 & .53 \\
Latent: FreqEdit & .50 & .15 & .71 & \underline{.50} & .18 & \underline{.65} & .59 & .20 & .63 & \textbf{.45} & \underline{.16} & \underline{.74} & \textbf{.55} & .21 & \textbf{.66} & \underline{.62} & .21 & .64 & .54 & .19 & .67 \\
\bottomrule
\end{tabular}}
\par
\caption{\footnotesize \textbf{Cycle editing on the 120 images of VAE-LFA} (their protocol and metric code: five local edit / undo pairs per image, round 10 against the input, mean of seeds 0 and 10). Rows combine a codec route between rounds (stock VAE or the \method codec with an 8-bit image hand-off, or latent pass-through without decoding) with one drift correction (none, BP, LFA, FreqEdit). BP: latent back-projection using the current round's noise level and a band profile calibrated once per editor on 8 external images. SD3-UE rows use our fix of the decoding and latent pass-through in VAE-LFA's released SD3-UE code; $^\dagger$: that code, run unmodified. Bold: best; underlined: second best per editor and column. Category ties at displayed precision share ranks; all-images columns follow the unrounded rankings in Table~\ref{tab:vaelfa-cycle}. Gray diagnostics are excluded.}
\label{tab:vaelfa-cycle-full}
\end{table*}

\subsection{\texorpdfstring{EdiVal-10$^\dagger$}{EdiVal-10\textdagger}}
\label{app:edival10-full}

Table~\ref{tab:edival10-all} gives every EdiVal-10$^\dagger$, LH-40, and NoOp-50 result for FLUX.2 [klein] and SD3-UE.

\paragraph{Long-horizon example.} The LH-40 example in Figure~\ref{fig:long-horizon} includes a saddlebag (turn 5), blue shirt (17), sneakers (21), and ``RIDE'' text (29). Ours and ACEdit complete seven of ten edits in this chain, versus six for each other comparator. 

\begin{table*}
\centering
\footnotesize
\setlength{\tabcolsep}{2.5pt}
\resizebox{\textwidth}{!}{%
\begin{tabular}{@{}ll ccccccc@{\hspace{8pt}}ccccccc@{\hspace{8pt}}ccc@{}}
\toprule
& & \multicolumn{7}{c}{\textbf{10 turns} (197 chains)} & \multicolumn{7}{c}{\textbf{40 turns}: LH-40 (197 chains)} & \multicolumn{3}{c}{\textbf{NoOp-50}} \\
\cmidrule(lr){3-9}\cmidrule(lr){10-16}\cmidrule(l){17-19}
Route & Method & Edit$\uparrow$ & IF@5$\uparrow$ & CC$\uparrow$ & O@5$\uparrow$ & \shortstack{Bg\\LPIPS$\downarrow$} & $\Delta E\downarrow$ & $\Delta$HPS$\uparrow$ & Edit$\uparrow$ & IF@5$\uparrow$ & CC$\uparrow$ & O@5$\uparrow$ & \shortstack{Bg\\LPIPS$\downarrow$} & $\Delta E\downarrow$ & $\Delta$HPS$\uparrow$ & LPIPS$\downarrow$ & $\Delta E\downarrow$ & PSNR$\uparrow$ \\
\midrule
\multicolumn{19}{@{}l}{\emph{FLUX.2 [klein] 9B, 4 steps}} \\
\multirow{4}{*}{Latent}
 & Uncorrected & \textbf{65.48} & \underline{7.11} & 77.98 & \underline{24.82} & 0.383 & 20.6 & $\underline{-0.77}$ & \underline{64.77} & \underline{8.12} & 57.04 & 24.25 & 0.671 & 38.2 & $\underline{-2.95}$ & 0.700 & 39.2 & 10.77 \\
 & VAE-LFA~\cite{vaelfa} & \underline{65.18} & 6.09 & 79.00 & 23.03 & \underline{0.325} & \underline{13.0} & $-0.93$ & \textbf{65.23} & \textbf{9.14} & \underline{62.54} & \textbf{26.27} & \underline{0.526} & \underline{20.6} & $-3.23$ & \underline{0.529} & \underline{21.3} & \underline{14.36} \\
 & FreqEdit~\cite{freqedit} & 62.18 & 5.58 & \underline{80.11} & 22.31 & 0.340 & 17.8 & $-0.82$ & 60.56 & 6.60 & 52.90 & 21.90 & 0.658 & 33.2 & $-4.11$ & 0.730 & 35.2 & 11.31 \\
 & BP & 65.13 & \textbf{7.61} & \textbf{84.28} & \textbf{26.05} & \textbf{0.206} & \textbf{8.9} & {\boldmath$-0.31$} & 64.47 & 7.61 & \textbf{83.88} & \underline{26.07} & \textbf{0.219} & \textbf{9.2} & {\boldmath$-0.48$} & \textbf{0.030} & \textbf{3.7} & \textbf{30.22} \\
\cmidrule(l){2-19}
\multirow{5}{*}{Pixel}
 & Stock & \textbf{66.35} & 5.58 & 76.29 & 21.92 & 0.402 & 21.9 & $-1.06$ & \underline{64.26} & \underline{8.63} & 50.08 & 24.14 & 0.697 & 42.3 & $\underline{-4.20}$ & 0.730 & 43.5 & 9.20 \\
 & Emu Edit~\cite{emuedit} & \underline{66.09} & \underline{7.11} & \textbf{82.15} & \underline{25.06} & \textbf{0.265} & \textbf{9.2} & {\boldmath$-0.06$} & \textbf{64.97} & 7.11 & \textbf{78.55} & \underline{24.62} & \textbf{0.357} & \textbf{13.7} & {\boldmath$-0.47$} & \textbf{0.231} & \textbf{10.6} & \textbf{18.70} \\
 & VAE-LFA~\cite{vaelfa} & \textbf{66.35} & 6.60 & 77.19 & 23.87 & 0.341 & 13.2 & $-1.19$ & 64.16 & \textbf{10.15} & 55.96 & \textbf{26.91} & \underline{0.550} & \underline{21.4} & $\underline{-4.20}$ & \underline{0.590} & \underline{21.9} & \underline{14.24} \\
 & FreqEdit~\cite{freqedit} & 61.42 & 6.09 & 77.96 & 23.15 & 0.367 & 18.5 & $-1.19$ & 59.70 & 4.06 & 42.00 & 16.29 & 0.759 & 39.1 & $-5.91$ & 0.848 & 42.2 & 8.80 \\
 & Ours w/o wrapper & 65.13 & \textbf{7.61} & \underline{80.41} & \textbf{25.90} & \underline{0.277} & \underline{12.4} & $\underline{-0.82}$ & 62.08 & 5.08 & \underline{57.29} & 19.76 & 0.670 & 32.6 & $-4.25$ & 0.763 & 42.6 & 12.36 \\
\cmidrule(l){2-19}
\multirow{4}{*}{Pixel}
 & Ours w/o BP & \textbf{66.25} & \underline{8.63} & 77.48 & \underline{27.30} & 0.386 & 20.6 & $\underline{-0.72}$ & \textbf{64.72} & \textbf{8.12} & 56.88 & \textbf{24.20} & 0.671 & 39.3 & $\underline{-2.77}$ & 0.705 & 40.7 & 10.82 \\
 & VAE-LFA (wrapped)~\cite{vaelfa} & \underline{64.97} & 6.09 & 78.94 & 22.99 & \underline{0.330} & \underline{13.1} & $-0.91$ & \underline{64.57} & \underline{7.61} & \underline{61.68} & \underline{23.99} & \underline{0.524} & \underline{20.5} & $-3.15$ & \underline{0.529} & \underline{21.1} & \underline{14.46} \\
 & FreqEdit (wrapped)~\cite{freqedit} & 61.17 & 5.08 & \underline{80.14} & 21.24 & 0.348 & 18.1 & $-0.80$ & 61.02 & 4.57 & 53.55 & 18.19 & 0.682 & 35.6 & $-3.70$ & 0.741 & 36.9 & 11.04 \\
 & \textbf{Ours} & 64.37 & \textbf{9.64} & \textbf{83.74} & \textbf{29.29} & \textbf{0.216} & \textbf{9.7} & {\boldmath$-0.39$} & 63.66 & 5.58 & \textbf{82.54} & 22.24 & \textbf{0.249} & \textbf{12.9} & {\boldmath$-0.69$} & \textbf{0.082} & \textbf{10.6} & \textbf{19.92} \\
\cmidrule(l){2-19}
Trained & ACEdit (latent)~\cite{acedit} & 66.04 & 7.11 & 82.43 & 25.12 & 0.271 & 12.5 & $-0.30$ & 63.71 & 9.14 & 63.28 & 27.06 & 0.516 & 22.4 & $-3.41$ & 0.560 & 24.2 & 13.73 \\
\midrule
\multicolumn{19}{@{}l}{\emph{SD3-UE, $512\times512$}} \\
\multirow{4}{*}{Latent}
 & Uncorrected & \textbf{46.55} & 1.02 & 65.24 & 8.59 & 0.436 & 23.5 & $\underline{-1.56}$ & 36.45 & \underline{1.02} & 44.88 & \underline{7.77} & \underline{0.637} & 41.5 & $-5.68$ & \underline{0.715} & 53.7 & 9.18 \\
 & VAE-LFA~\cite{vaelfa} & \underline{46.45} & \textbf{2.03} & 66.40 & \textbf{12.09} & 0.436 & \underline{19.3} & {\boldmath$-1.28$} & \underline{38.32} & \textbf{1.52} & \underline{45.49} & \textbf{9.51} & 0.645 & \underline{34.2} & $\underline{-4.86}$ & \underline{0.715} & \underline{42.3} & \underline{10.64} \\
 & FreqEdit~\cite{freqedit} & 40.61 & 0.51 & \textbf{69.32} & 6.33 & \underline{0.386} & 19.7 & $-2.10$ & 26.95 & 0.00 & 32.69 & 0.00 & 0.698 & 50.9 & $-8.19$ & 0.750 & 60.9 & 8.62 \\
 & BP & 46.04 & \underline{1.52} & \underline{67.71} & \underline{10.53} & \textbf{0.336} & \textbf{17.3} & $-1.76$ & \textbf{41.47} & 0.00 & \textbf{65.29} & 0.00 & \textbf{0.364} & \textbf{19.5} & {\boldmath$-2.18$} & \textbf{0.254} & \textbf{13.5} & \textbf{18.21} \\
\cmidrule(l){2-19}
\multirow{5}{*}{Pixel}
 & Stock & \underline{46.14} & \underline{0.51} & 64.37 & 6.05 & 0.453 & 23.4 & $-1.37$ & 40.76 & \textbf{1.02} & 46.01 & \underline{7.78} & 0.655 & 44.8 & $-4.60$ & 0.755 & 62.2 & 8.36 \\
 & Emu Edit~\cite{emuedit} & \textbf{47.87} & \underline{0.51} & \underline{66.99} & 6.09 & 0.415 & 19.6 & {\boldmath$-0.79$} & \textbf{44.21} & \underline{0.51} & \textbf{64.28} & 6.02 & \textbf{0.444} & \textbf{22.2} & {\boldmath$-1.58$} & \textbf{0.332} & \textbf{16.7} & \textbf{15.75} \\
 & VAE-LFA~\cite{vaelfa} & 46.04 & \textbf{1.02} & 65.10 & \underline{8.53} & 0.432 & \underline{18.8} & $\underline{-1.12}$ & \underline{43.35} & \textbf{1.02} & \underline{50.58} & \textbf{7.84} & \underline{0.598} & \underline{28.8} & $\underline{-3.15}$ & \underline{0.703} & 41.2 & \underline{11.34} \\
 & FreqEdit~\cite{freqedit} & 42.13 & \underline{0.51} & \textbf{67.40} & 6.28 & \textbf{0.400} & 19.7 & $-2.04$ & 33.71 & \textbf{1.02} & 31.82 & 6.98 & 0.738 & 56.2 & $-7.51$ & 0.784 & 67.5 & 8.05 \\
 & Ours w/o wrapper & 40.25 & \textbf{1.02} & 62.96 & \textbf{8.56} & \underline{0.411} & \textbf{16.3} & $-3.21$ & 23.60 & \underline{0.51} & 24.02 & 4.86 & 0.806 & 41.6 & $-10.12$ & 0.843 & \underline{37.7} & 10.91 \\
\cmidrule(l){2-19}
\multirow{4}{*}{Pixel}
 & Ours w/o BP & \underline{44.06} & \underline{0.51} & 65.06 & 6.06 & 0.441 & 23.4 & $\underline{-1.44}$ & 33.05 & \textbf{0.00} & 40.92 & \textbf{0.00} & 0.676 & 44.0 & $-5.66$ & 0.739 & 56.0 & 9.26 \\
 & VAE-LFA (wrapped)~\cite{vaelfa} & \textbf{46.45} & \textbf{1.02} & 65.40 & \underline{8.56} & 0.428 & \underline{18.7} & {\boldmath$-1.23$} & \underline{36.19} & \textbf{0.00} & \underline{43.08} & \textbf{0.00} & \underline{0.663} & \underline{35.1} & $\underline{-5.14}$ & \underline{0.722} & \underline{41.6} & \underline{11.00} \\
 & FreqEdit (wrapped)~\cite{freqedit} & 41.07 & \underline{0.51} & \textbf{68.68} & 6.32 & \underline{0.390} & 19.6 & $-2.13$ & 26.35 & \textbf{0.00} & 30.80 & \textbf{0.00} & 0.712 & 51.9 & $-7.86$ & 0.762 & 61.7 & 8.88 \\
 & \textbf{Ours} & 43.55 & \textbf{1.02} & \underline{68.24} & \textbf{8.61} & \textbf{0.339} & \textbf{17.4} & $-1.65$ & \textbf{40.81} & \textbf{0.00} & \textbf{64.40} & \textbf{0.00} & \textbf{0.408} & \textbf{22.5} & {\boldmath$-1.89$} & \textbf{0.314} & \textbf{18.6} & \textbf{16.40} \\
\bottomrule
\end{tabular}}
\par
\caption{\footnotesize \textbf{EdiVal-10$^\dagger$: cumulative local edits with DiT editors} (FLUX.2 at $1024\times1024$, SD3-UE at $512\times512$, same seed per turn for all methods). Left: one edit per turn. Middle: LH-40, each edit followed by three no-op turns, measured after them. Right: NoOp-50, fifty turns of ``Keep the image as is.'' and no edit, on the same 197 inputs. Rows: drift corrections in each route (latent: the output latent is passed on, never decoded; pixel: an 8-bit image is handed off every turn, through the stock VAE or the \method codec); ACEdit is fine-tuned FLUX.2, with no available SD3-UE checkpoint; ours keeps codec and editor weights frozen on both editors. Edit: fraction of edits that succeeded. IF@5: fraction of chains whose first five edits all succeeded. CC: final content consistency (DINOv3~\cite{dinov3} similarity to the input for untouched objects and background). O@5: $\sqrt{\mathrm{IF@5}\cdot\mathrm{CC@5}}$, with CC after five edits. EdiVal columns in \%. Bg LPIPS and $\Delta E$ (mean CIELAB color difference): last turn vs.\ the input, outside the boxes of all the chain's objects; for NoOp-50, LPIPS, $\Delta E$ and PSNR (dB) of turn 50 vs.\ the input, over the whole image. $\Delta$HPS: change in HPSv3 image quality from the first to the last scored turn. All metrics at EdiVal's $512\times512$ (at $1024$: Table~\ref{tab:edival10-full}). Bold/underlined: best/second within each route/codec block. Ties at displayed precision share ranks. Metric definitions: Table~\ref{tab:metrics}.}
\label{tab:edival10-all}
\end{table*}

Table~\ref{tab:edival10-full} gives FLUX.2 drift (background, and NoOp-50 over the whole image) at EdiVal's scoring size and at the generation size.

\begin{table*}
\centering
\footnotesize
\setlength{\tabcolsep}{4pt}
\resizebox{\textwidth}{!}{%
\begin{tabular}{@{}ll cccc@{\hspace{12pt}}cccc@{\hspace{12pt}}cccccc@{}}
\toprule
& & \multicolumn{4}{c}{\textbf{10 turns}} & \multicolumn{4}{c}{\textbf{40 turns} (LH-40)} & \multicolumn{6}{c}{\textbf{50 no-op turns} (NoOp-50)} \\
\cmidrule(lr){3-6}\cmidrule(lr){7-10}\cmidrule(l){11-16}
& & \multicolumn{2}{c}{Bg LPIPS$\downarrow$} & \multicolumn{2}{c}{$\Delta E\downarrow$} & \multicolumn{2}{c}{Bg LPIPS$\downarrow$} & \multicolumn{2}{c}{$\Delta E\downarrow$} & \multicolumn{2}{c}{LPIPS$\downarrow$} & \multicolumn{2}{c}{$\Delta E\downarrow$} & \multicolumn{2}{c}{PSNR$\uparrow$} \\
\cmidrule(lr){3-4}\cmidrule(lr){5-6}\cmidrule(lr){7-8}\cmidrule(lr){9-10}\cmidrule(lr){11-12}\cmidrule(lr){13-14}\cmidrule(l){15-16}
Route & Method & 512 & 1024 & 512 & 1024 & 512 & 1024 & 512 & 1024 & 512 & 1024 & 512 & 1024 & 512 & 1024 \\
\midrule
\multirow{4}{*}{Latent}
 & Uncorrected & 0.383 & 0.420 & 20.6 & 21.0 & 0.671 & 0.710 & 38.2 & 38.8 & 0.700 & 0.736 & 39.2 & 39.8 & 10.77 & 10.55 \\
 & VAE-LFA~\cite{vaelfa} & \underline{0.325} & \underline{0.339} & \underline{13.0} & \underline{13.3} & \underline{0.526} & \underline{0.538} & \underline{20.6} & \underline{20.8} & \underline{0.529} & \underline{0.546} & \underline{21.3} & \underline{21.6} & \underline{14.36} & \underline{14.17} \\
 & FreqEdit~\cite{freqedit} & 0.340 & 0.370 & 17.8 & 18.1 & 0.658 & 0.697 & 33.2 & 33.7 & 0.730 & 0.785 & 35.2 & 35.5 & 11.31 & 11.14 \\
 & BP & \textbf{0.206} & \textbf{0.194} & \textbf{8.9} & \textbf{9.2} & \textbf{0.219} & \textbf{0.203} & \textbf{9.2} & \textbf{9.5} & \textbf{0.030} & \textbf{0.053} & \textbf{3.7} & \textbf{4.0} & \textbf{30.22} & \textbf{29.07} \\
\cmidrule(l){2-16}
\multirow{4}{*}{Pixel}
 & Stock & 0.402 & 0.435 & 21.9 & 22.4 & 0.697 & \underline{0.716} & 42.3 & 43.1 & \underline{0.730} & \underline{0.753} & 43.5 & 44.2 & 9.20 & 8.96 \\
 & VAE-LFA~\cite{vaelfa} & \underline{0.341} & \underline{0.353} & \underline{13.2} & \underline{13.5} & \textbf{0.550} & \textbf{0.562} & \textbf{21.4} & \textbf{21.7} & \textbf{0.590} & \textbf{0.624} & \textbf{21.9} & \textbf{22.3} & \textbf{14.24} & \textbf{14.03} \\
 & FreqEdit~\cite{freqedit} & 0.367 & 0.405 & 18.5 & 18.9 & 0.759 & 0.818 & 39.1 & 39.7 & 0.848 & 0.905 & \underline{42.2} & \underline{42.7} & 8.80 & 8.64 \\
 & Ours w/o wrapper & \textbf{0.277} & \textbf{0.324} & \textbf{12.4} & \textbf{13.1} & \underline{0.670} & 0.785 & \underline{32.6} & \underline{34.2} & 0.763 & 0.873 & 42.6 & 44.7 & \underline{12.36} & \underline{11.41} \\
\cmidrule(l){2-16}
\multirow{4}{*}{Pixel}
 & Ours w/o BP & 0.386 & 0.424 & 20.6 & 21.1 & 0.671 & 0.714 & 39.3 & 39.7 & 0.705 & 0.744 & 40.7 & 41.1 & 10.82 & 10.67 \\
 & VAE-LFA (wrapped)~\cite{vaelfa} & \underline{0.330} & \underline{0.344} & \underline{13.1} & \underline{13.5} & \underline{0.524} & \underline{0.535} & \underline{20.5} & \underline{20.9} & \underline{0.529} & \underline{0.546} & \underline{21.1} & \underline{21.4} & \underline{14.46} & \underline{14.28} \\
 & FreqEdit (wrapped)~\cite{freqedit} & 0.348 & 0.378 & 18.1 & 18.5 & 0.682 & 0.724 & 35.6 & 36.0 & 0.741 & 0.789 & 36.9 & 37.3 & 11.04 & 10.93 \\
 & \textbf{Ours} & \textbf{0.216} & \textbf{0.206} & \textbf{9.7} & \textbf{10.2} & \textbf{0.249} & \textbf{0.249} & \textbf{12.9} & \textbf{13.3} & \textbf{0.082} & \textbf{0.124} & \textbf{10.6} & \textbf{11.0} & \textbf{19.92} & \textbf{19.70} \\
\bottomrule
\end{tabular}}
\par
\caption{\footnotesize \textbf{Background drift at two resolutions}, EdiVal-10$^\dagger$, FLUX.2 [klein] 9B: at EdiVal's $512\times512$ (Table~\ref{tab:edival10}) and at the generation size $1024\times1024$, with the same boxes; NoOp-50: LPIPS, $\Delta E$ and PSNR (dB) of turn 50 vs.\ the input, over the whole image. The sizes rank the methods the same way except BP with the stock VAE on LH-40, which is above no correction at 1024 and below it at 512: its extra error is fine texture. Chains with an empty background at the end (3--6 of 197 for 10 turns, 2--12 for LH-40) are left out. Pixel-route LFA: VAE-LFA's alignment on the output latent before the 8-bit hand-off; latent-route LFA: VAE-LFA's own white-box mode. Bold/underlined: best/second within each route/codec block. Ties at displayed precision share ranks.}
\label{tab:edival10-full}
\end{table*}

\subsection{PasteBench object retention}
\label{app:pastebench-retention}

 Table~\ref{tab:pastebench}'s Kept column gives the mean number of the ten pasted objects still recognizable anywhere in each turn-10 image, over the 20 scenes, with a 95\% interval in brackets. A VLM assessor (OpenAI Codex) scored all 120 turn-10 images (20 scenes, six methods; 1,200 decisions) under a rubric fixed before scoring. An object counts if a recognizable counterpart of its type is visible anywhere in the image, allowing changes in color, size, pose, style, and position; it does not count if it is gone or has turned into a different kind of object. Each object scores at most once, and one visible instance cannot satisfy two objects. For each scene, the assessor saw the original scene, the numbered cutouts, and the six outputs under random IDs in a shuffled order; it searched the whole frame and enlarged uncertain regions before declaring an object absent, and boxed the evidence for each object found. Method names were hidden in the scoring views, though the assessor had seen some outputs earlier. Intervals resample the 20 scenes (10,000 bootstrap resamples). The full audit is a separate PDF in the supplementary material: rubric and procedure, all 20 scene figures with evidence boxes, and every per-image score. The supplementary material also includes an HTML viewer (\texttt{pastebench\_viewer.html}) of all 20 scenes and six methods at turns 0, 10, and 20, with per-scene LPIPS and PSNR at turn 20.

\section{Ablations and analysis}
\label{app:ablations}

\subsection{Codec ablations}
\label{app:ablations-codec}

\subsubsection{Empirical deviation from the pretrained VAE}
\label{app:deviation}

Table~\ref{tab:deviation} measures the deviations of Section~\ref{sec:codec} with the codec settings of our editing experiments. Encoder: the relative change of the latent of a real image, $\|\Ep(x)-E(x)\|_2/\|E(x)\|_2$. Decoder: the PSNR between $\Dp(z)$ and $D(z)$, on latents of real images, $z=E(x)$, and on latents generated in one turn (Table~\ref{tab:deviation} lists the generators; a no-op turn for SD3-UE). Images: the first 50 REED images ($512\times512$) for SD~1.5 and SD~2.1, the first 30 for SD3-UE, and the first 30 EdiVal-10$^\dagger$ photographs ($1024\times1024$) for FLUX.2; none is in the PCA calibration set.

\paragraph{Interpreting the deviations.}
A decoder agreement of 42--46\,dB PSNR corresponds to a root-mean-square difference of 1.3--2.0 gray levels on an 8-bit scale, while the pretrained VAE's reconstruction of the same images lies 28--37\,dB from the input (Table~\ref{tab:deviation}). With the codec alone, edit success on EdiVal-10$^\dagger$ is 66.25\% versus 66.35\% for the stock VAE with FLUX.2 [klein], and 44.06\% versus 46.14\% with SD3-UE (Table~\ref{tab:edival10-all}).
The low-variance PCA choice keeps these deviations small: with a random orthogonal basis, encoder deviations reach 245--852\%, and first-reconstruction PSNR falls to 5--14\,dB, versus 28 and 36\,dB with PCA (Appendix~\ref{app:basis}).

\begin{table*}
\centering
\small
\setlength{\tabcolsep}{4pt}
\begin{tabular}{@{}lrrrr@{}}
\toprule
& Encoder & \multicolumn{2}{c}{Decoder PSNR (dB)$\uparrow$} & Stock VAE \\
\cmidrule(l){3-4}
VAE & change (\%)$\downarrow$ & real $z$ & generated $z$ & recon.\ (dB) \\
\midrule
SD~1.5 & 2.21 & 44.5 & 43.6 & 28.0 \\
SD~2.1 & 2.21 & 45.4 & 42.2 & 29.1 \\
FLUX.2 [klein] & 5.85 & 43.3 & 42.3 & 36.7 \\
SD3-UE & 1.31 & 45.7 & 43.5 & 34.3 \\
\bottomrule
\end{tabular}
\par
\caption{\footnotesize \textbf{How far the \method codec is from the pretrained VAE it wraps.} Encoder: relative change of the latent. Decoder: $\Dp(z)$ against $D(z)$. Generated latents: InstructPix2Pix (SD~1.5), SD~2.1 image-to-image, FLUX.2 [klein] edits, SD3-UE no-op turns. Stock VAE recon.: the pretrained VAE's own reconstruction $D(E(x))$ against $x$, for scale; it does not involve the codec.}
\label{tab:deviation}
\end{table*}

\subsubsection{Choice of basis}
\label{app:basis}
\begin{figure*}
\centering
\input{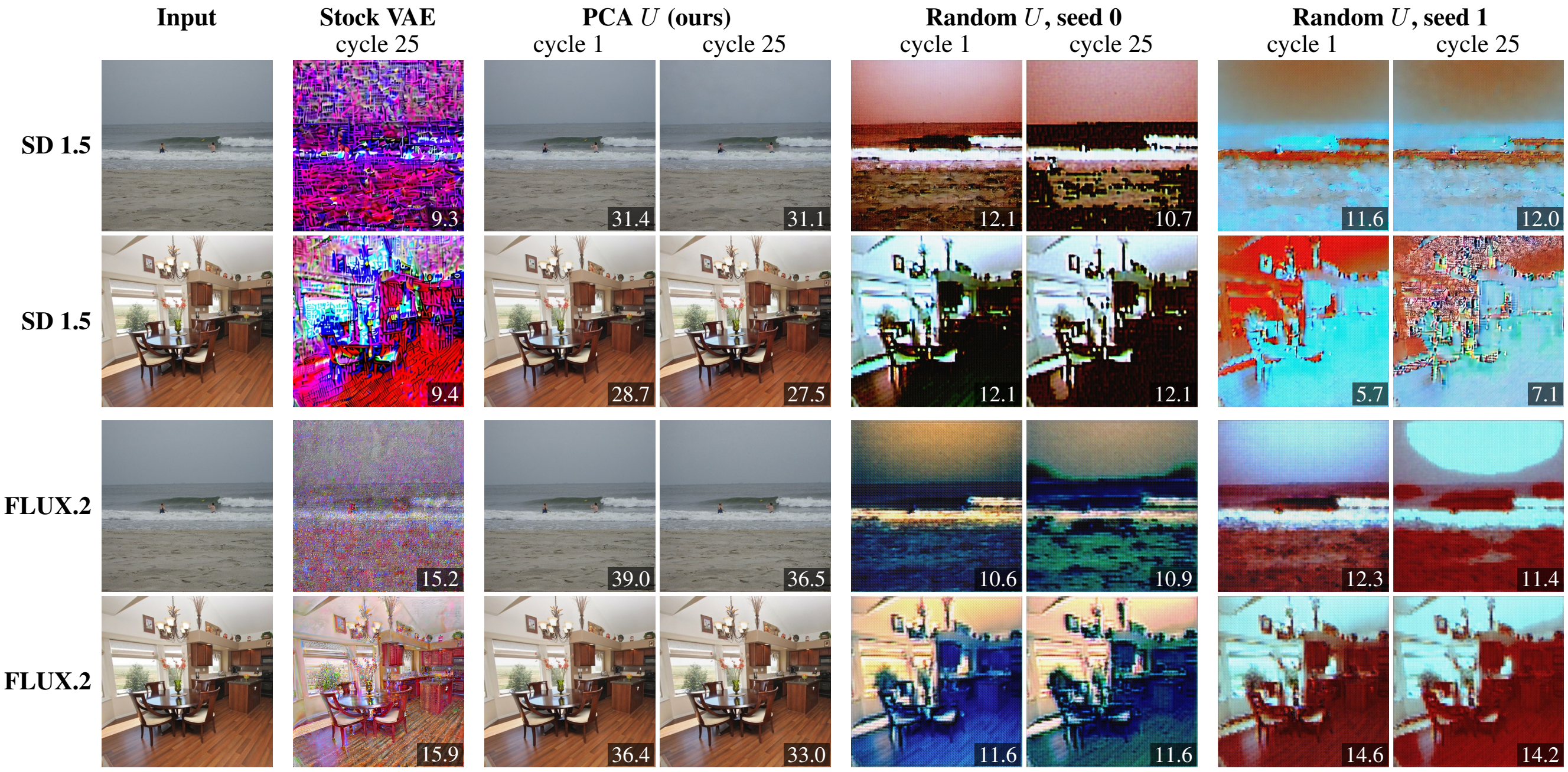}
\par
\caption{\footnotesize \textbf{PCA vs.\ random orthogonal basis, examples.} Two of REED's ImagenHub images (Section~\ref{sec:cycles}) through the SD~1.5 and FLUX.2 codecs of Table~\ref{tab:basis}, with 8-bit hand-offs; corner: PSNR (dB) to the input. The stock VAE drifts over 25 cycles. With either basis, the codec stays near its first reconstruction, but only with the PCA basis is that reconstruction close to the input.}
\label{fig:basis}
\end{figure*}

The codec reserves the lowest-variance PCA directions of $8\times8$ RGB patches (Section~\ref{sec:codec}). Here we replace the whole PCA basis $U$ with a random orthogonal matrix (Haar-distributed, three seeds), keeping $\mu$, $k$, $c$, and the rest of the reconstruction codec of Table~\ref{tab:cycles} fixed. We use the images of that test, the 179 ImagenHub images of REED's benchmark (Section~\ref{sec:cycles}). Table~\ref{tab:basis} reports codec deviations and 25-cycle reconstruction scores. Figure~\ref{fig:basis} shows examples.

Since $\Ep(\Dp(z))=z$ holds for any orthogonal $U$, the random codec drifts little: its PSNR changes by at most 1.6\,dB over 25 cycles. However, natural images are no longer near zero along its reserved directions (RMS 0.39--0.57, versus 0.003 with PCA), so $c\,x_0$ dominates the latent. The encoder then moves latents by 245--852\%, 1--4\% of decoded pixel values leave $[-1,1]$, and the codec settles at a fixed point far from the input, with false colors and blocky structure. Orthogonality thus provides consistency, while low-variance PCA keeps that fixed point close to the input.

\begin{table}
\centering
\footnotesize
\setlength{\tabcolsep}{3pt}
\begin{tabular}{@{}lrrrr@{}}
\toprule
& \multicolumn{2}{c}{SD~1.5} & \multicolumn{2}{c}{FLUX.2} \\
\cmidrule(lr){2-3}\cmidrule(l){4-5}
$U$ & PCA & random & PCA & random \\
\midrule
Encoder change (\%)$\downarrow$ & 1.7 & 245--320 & 5.3 & 763--852 \\
Decoder PSNR (dB)$\uparrow$ & 44.5 & 28.8--31.0 & 44.3 & 19.2--20.2 \\
PSNR, cycle 1 (dB)$\uparrow$ & 28.2 & 5.2--12.6 & 35.7 & 12.0--14.1 \\
PSNR, cycle 25 (dB)$\uparrow$ & 27.5 & 6.7--12.3 & 32.0 & 12.1--13.4 \\
LPIPS, cycle 25$\downarrow$ & 0.07 & 0.64--0.83 & 0.03 & 0.72--0.83 \\
\bottomrule
\end{tabular}
\par
\caption{\footnotesize \textbf{PCA vs.\ random orthogonal basis.} The \method reconstruction codec of Table~\ref{tab:cycles}, with only $U$ changed; the random columns give the range over three seeds. Encoder change and decoder PSNR are measured as in Table~\ref{tab:deviation}, on the first 50 of REED's 179 images. They differ from Table~\ref{tab:deviation}, whose SD~1.5 row uses the REED editing codec and whose FLUX.2 row uses EdiVal-10$^\dagger$ photographs. The cycle rows compare each image with the original after 1 and 25 encode--decode cycles with 8-bit hand-offs, on all 179 images.}
\label{tab:basis}
 \end{table}

\subsubsection{Range handling}
\label{app:no-softlimit}

The soft-limit control in Table~\ref{tab:reed-editing-full} measures the effect of the soft limit on the stock decoder, but does not isolate its contribution within the wrapper. The ablation in Table~\ref{tab:no-softlimit} compares the plain construction in Eqs.~\ref{eq:encoder} and~\ref{eq:decoder} with the version using the soft limit and pixel map, under the same 8-bit editing protocol. 

Without range handling, the share of pixels with a channel outside $[-1,1]$ at the 8-bit hand-off, which serialization clips, grows along the chain: averaged over the 179 chains, it rises from 1--4\% in round~1 to 25--39\% in round~25 for InstructPix2Pix, MagicBrush, DiffEdit and SD Inpainting. With range handling, the inverse pixel map keeps the final output in $(-1,1)$, preventing clipping. For these four editors, range handling improves round-25 PSNR by 0.97--1.60\,dB and LPIPS by 0.021--0.065. For Paint-by-Example, where only 0.8\% of pixels are clipped (mean over rounds), the two variants agree within 0.1\,dB and 0.003 LPIPS.

\begin{table}[h]
\centering
\small
\setlength{\tabcolsep}{3pt}
\resizebox{\columnwidth}{!}{%
\begin{tabular}{@{}llcccccc@{}}
\toprule
& & \multicolumn{3}{c}{LPIPS$\downarrow$} & \multicolumn{3}{c}{PSNR$\uparrow$} \\
\cmidrule(lr){3-5}\cmidrule(l){6-8}
Editor & Method & 5 & 15 & 25 & 5 & 15 & 25 \\
\midrule
\multirow{3}{*}{IP2P} & REED-VAE (paper) & 0.180 & \textbf{0.450} & \textbf{0.580} & 19.810 & \underline{13.360} & \textbf{11.360} \\
 & Codec, no range handling & \underline{0.175} & 0.484 & 0.651 & \underline{20.321} & 12.941 & 10.356 \\
 & Ours w/o BP & \textbf{0.172} & \underline{0.458} & \underline{0.608} & \textbf{20.653} & \textbf{13.653} & \underline{11.332} \\
\midrule
\multirow{3}{*}{MagicBrush} & REED-VAE (paper) & 0.190 & 0.510 & 0.690 & 21.530 & \underline{16.660} & \textbf{14.090} \\
 & Codec, no range handling & \underline{0.148} & \underline{0.453} & \underline{0.649} & \underline{21.862} & 16.027 & 12.477 \\
 & Ours w/o BP & \textbf{0.147} & \textbf{0.437} & \textbf{0.615} & \textbf{22.315} & \textbf{16.795} & \underline{13.596} \\
\midrule
\multirow{3}{*}{DiffEdit} & REED-VAE (paper) & \underline{0.300} & 0.550 & 0.680 & \textbf{16.240} & \textbf{12.910} & \underline{11.440} \\
 & Codec, no range handling & \textbf{0.294} & \underline{0.484} & \underline{0.572} & 15.377 & 11.699 & 10.658 \\
 & Ours w/o BP & \textbf{0.294} & \textbf{0.473} & \textbf{0.551} & \underline{15.804} & \underline{12.502} & \textbf{11.627} \\
\midrule
\multirow{3}{*}{PbE} & REED-VAE (paper) & 0.200 & 0.440 & 0.590 & 19.620 & 16.190 & 14.430 \\
 & Codec, no range handling & \textbf{0.147} & \textbf{0.270} & \underline{0.379} & \underline{20.493} & \underline{17.950} & \underline{16.711} \\
 & Ours w/o BP & \underline{0.149} & \underline{0.273} & \textbf{0.378} & \textbf{20.573} & \textbf{18.010} & \textbf{16.800} \\
\midrule
\multirow{3}{*}{SD Inp.} & REED-VAE (paper) & 0.170 & 0.470 & 0.650 & \textbf{23.140} & \underline{16.660} & \underline{13.610} \\
 & Codec, no range handling & \textbf{0.139} & \underline{0.417} & \underline{0.599} & \underline{23.096} & 16.536 & 13.065 \\
 & Ours w/o BP & \underline{0.143} & \textbf{0.403} & \textbf{0.534} & 23.059 & \textbf{17.170} & \textbf{14.665} \\
\bottomrule
\end{tabular}}
\par
\caption{\footnotesize \textbf{\method codec with and without range handling, vs.\ REED-VAE}. Protocol of Table~\ref{tab:reed-editing-full}, without BP. No range handling: the same codec without the soft limit and the pixel map, with the same PCA coordinates and $c=29.8$; Ours w/o BP: the codec with both. Bold: best; underlined: second best per editor and column.}
\label{tab:no-softlimit}
 \end{table}

\subsection{BP ablations}
\label{app:ablations-bp}

\subsubsection{Threshold rule}
\label{app:bp-threshold-ablation}

Across eight settings (two editors, two protocols, two routes), Eq.~\ref{eq:bp-threshold} improves mean LPIPS, L1, and SSIM over each ablation, giving 16 paired comparisons (Tables~\ref{tab:vaelfa-bp-rules} and~\ref{tab:vaelfa-bp-ablation}). The LPIPS and L1 improvements are significant in all 16 and the SSIM improvements in 15 (95\% paired bootstrap). In every comparison, more image--seed pairs improve than degrade by over 0.05 LPIPS.

\begin{table}
\centering
\small
\setlength{\tabcolsep}{3pt}
\resizebox{\columnwidth}{!}{%
\begin{tabular}{@{}lcccccccc@{}}
\toprule
& \multicolumn{4}{c}{No-op} & \multicolumn{4}{c}{Edit/undo} \\
\cmidrule(lr){2-5}\cmidrule(l){6-9}
Route & No BP & $\rho_{b,\ell}=1$ & Fixed & Ours & No BP & $\rho_{b,\ell}=1$ & Fixed & Ours \\
\midrule
\multicolumn{9}{@{}l}{\textbf{FLUX.2 [klein]}} \\
Wrapped pixel & .349 & .182 & .207 & \textbf{.164} & .441 & .337 & .359 & \textbf{.314} \\
Latent & .337 & .163 & .169 & \textbf{.148} & .434 & .325 & .343 & \textbf{.308} \\
\multicolumn{9}{@{}l}{\textbf{SD3-UE}} \\
Wrapped pixel & .469 & .345 & .305 & \textbf{.282} & .596 & .554 & .567 & \textbf{.522} \\
Latent & .470 & .341 & .296 & \textbf{.280} & .596 & .554 & .569 & \textbf{.524} \\
\bottomrule
\end{tabular}}
\par
\caption{\footnotesize \textbf{Final LPIPS per BP threshold rule} (VAE-LFA's protocols and metric code; round 10 against the input, mean over 240 image--seed pairs). \emph{No BP}: the same route without correction. \emph{$\rho_{b,\ell}=1$}: one per-turn level for all bands. \emph{Fixed}: $\rho_{b,\ell}\hat\sigma_\ell$ replaced by $\sigma^{\mathrm{cal}}_{b,\ell}$. \emph{Ours}: Eq.~\ref{eq:bp-threshold}.}
\label{tab:vaelfa-bp-rules}
 \end{table}

\begin{table*}
\centering
\small
\resizebox{\textwidth}{!}{%
\begin{tabular}{@{}lcccccccc@{}}
\toprule
& \multicolumn{4}{c}{Versus $\rho_{b,\ell}=1$} & \multicolumn{4}{c}{Versus fixed $\sigma^{\mathrm{cal}}_{b,\ell}$} \\
\cmidrule(lr){2-5}\cmidrule(l){6-9}
Editor, route & LPIPS$\downarrow$ & L1$\downarrow$ & SSIM$\uparrow$ & Worse/better & LPIPS$\downarrow$ & L1$\downarrow$ & SSIM$\uparrow$ & Worse/better \\
\midrule
\multicolumn{9}{@{}l}{\emph{No-op, round 10}} \\
FLUX.2 [klein], wrapped pixel & $-.018\,[-.022,-.015]$ & $-.010\,[-.012,-.008]$ & $+.007\,[+.004,+.009]$ & 2/18 & $-.043\,[-.057,-.031]$ & $-.009\,[-.012,-.006]$ & $+.017\,[+.011,+.025]$ & 3/55 \\
FLUX.2 [klein], latent & $-.015\,[-.018,-.012]$ & $-.011\,[-.014,-.009]$ & $+.006\,[+.004,+.009]$ & 2/9 & $-.021\,[-.031,-.013]$ & $-.005\,[-.008,-.003]$ & $+.011\,[+.005,+.017]$ & 3/24 \\
SD3-UE, wrapped pixel & $-.063\,[-.073,-.053]$ & $-.030\,[-.033,-.026]$ & $+.032\,[+.026,+.039]$ & 6/133 & $-.023\,[-.036,-.011]$ & $-.009\,[-.014,-.005]$ & $+.019\,[+.010,+.029]$ & 16/58 \\
SD3-UE, latent & $-.061\,[-.071,-.051]$ & $-.038\,[-.042,-.034]$ & $+.034\,[+.027,+.041]$ & 8/119 & $-.016\,[-.028,-.004]$ & $-.008\,[-.013,-.004]$ & $+.015\,[+.005,+.025]$ & 20/45 \\
\midrule
\multicolumn{9}{@{}l}{\emph{Edit/undo, round 10}} \\
FLUX.2 [klein], wrapped pixel & $-.023\,[-.029,-.018]$ & $-.010\,[-.013,-.007]$ & $+.017\,[+.008,+.026]$ & 3/39 & $-.045\,[-.057,-.034]$ & $-.011\,[-.015,-.007]$ & $+.034\,[+.022,+.046]$ & 2/74 \\
FLUX.2 [klein], latent & $-.017\,[-.022,-.012]$ & $-.008\,[-.012,-.005]$ & $+.007\,[-.001,+.016]$ & 5/26 & $-.035\,[-.047,-.025]$ & $-.008\,[-.013,-.004]$ & $+.018\,[+.009,+.029]$ & 1/51 \\
SD3-UE, wrapped pixel & $-.032\,[-.040,-.024]$ & $-.023\,[-.026,-.020]$ & $+.026\,[+.018,+.035]$ & 8/72 & $-.044\,[-.055,-.035]$ & $-.025\,[-.030,-.021]$ & $+.050\,[+.039,+.061]$ & 8/88 \\
SD3-UE, latent & $-.030\,[-.038,-.023]$ & $-.029\,[-.033,-.025]$ & $+.028\,[+.021,+.036]$ & 6/59 & $-.045\,[-.055,-.035]$ & $-.031\,[-.036,-.025]$ & $+.054\,[+.040,+.068]$ & 8/96 \\
\bottomrule
\end{tabular}}
\par
\caption{\footnotesize \textbf{BP threshold ablations on the 120 images of VAE-LFA} (their protocols and metric code; round 10 against the input, seeds 0 and 10). Each cell gives Eq.~\ref{eq:bp-threshold} minus the ablation: the mean paired difference over 240 image--seed pairs, with its 95\% bootstrap interval over the 120 images (10,000 resamples, both seeds of an image kept together). Negative LPIPS and L1 and positive SSIM favor Eq.~\ref{eq:bp-threshold}. \emph{Versus $\rho_{b,\ell}=1$}: one per-turn level $\hat\sigma_\ell$ for all bands. \emph{Versus fixed $\sigma^{\mathrm{cal}}_{b,\ell}$}: $\rho_{b,\ell}\hat\sigma_\ell$ replaced by the calibrated level, with no per-turn estimate. \emph{Worse/better}: image--seed pairs, of 240, where Eq.~\ref{eq:bp-threshold} has LPIPS more than 0.05 above or below the ablation.}
\label{tab:vaelfa-bp-ablation}
\end{table*}

\subsubsection{Where the correction acts}
\label{app:bp-placement}

We test whether BP's advantage comes simply from acting after sampling. FreqEdit~\cite{freqedit} also corrects each turn with a wavelet-domain map: during the first denoising steps (here two of four), it blends the high-frequency db4 wavelet bands of the denoiser's predicted velocity with those of the reference, with per-pixel weights that are largest where the two differ least, relative to the largest difference in the image. BP instead acts once after sampling and keeps or reverts each Haar wavelet coefficient of the final latent's change against the turn's noise level (Eq.~\ref{eq:bp-threshold}). We move FreqEdit's map and blend to the end of sampling: onto the velocity of the last denoising step, or once onto the final latent with the turn's input latent as reference, which matches BP's position and reference. When moved, FreqEdit's default map ($\alpha_0=1.6$, $\gamma=2$, weights up to about 10) severely degrades the image from round 2 (round-10 LPIPS 0.89--0.97), so we also run the moved variants with $\gamma=-10$ and $\alpha_0=1/(e^{-10}-1)$, which bound the weights to $[0,1]$. Table~\ref{tab:vaelfa-freqedit-last} compares these variants, with the stock VAE as in FreqEdit's own setting, against \method on drift (LPIPS between round 10 and the input) and EdiVal's verification of the 1,200 edit/undo pairs, as in Table~\ref{tab:vaelfa-cycle}. With $\gamma=-10$, both moved variants keep fewer edits: EdiVal verifies 86.2--87.3\% of their edits and both the edit and the undo in 64.8--65.5\% of pairs, against 98.6\% and 87.0\% for \method. They also drift more than \method. With the \method codec or latent pass-through (not shown in the table), their no-op LPIPS is similar or higher (0.144--0.203 against 0.148--0.164) and their edit/undo LPIPS is lower (0.242--0.282 against 0.308--0.314), a drop that comes with fewer verified edits. Placement alone therefore does not reproduce BP's combination of low drift and verified edits. BP also retains the practical advantage of acting once at the end of each turn without interfering with the denoising process, unlike FreqEdit.

\begin{table}
\centering
\small
\setlength{\tabcolsep}{3pt}
\resizebox{\columnwidth}{!}{%
\begin{tabular}{@{}llccccc@{}}
\toprule
& & \multicolumn{2}{c}{Drift (LPIPS)} & \multicolumn{3}{c}{Edit success (\%)} \\
\cmidrule(lr){3-4}\cmidrule(l){5-7}
Method & Map & No-op$\downarrow$ & Edit/undo$\downarrow$ & Edit$\uparrow$ & Undo$\uparrow$ & Cycle$\uparrow$ \\
\midrule
\textbf{Ours} (\method + BP) & -- & \textbf{.164} & \textbf{.314} & 98.6 & \textbf{87.5} & \textbf{87.0} \\
\midrule
FreqEdit, early steps (original) & default & .384 & .465 & \textbf{98.7} & 73.8 & 73.6 \\
FreqEdit, last-step velocity & default & .889 & .965 & 54.7 & 19.0 & 12.2 \\
FreqEdit, last-step velocity & $\gamma=-10$ & .368 & .416 & 86.2 & 68.0 & 64.8 \\
FreqEdit, final latent & default & .891 & .967 & 51.0 & 19.5 & 11.7 \\
FreqEdit, final latent & $\gamma=-10$ & .386 & .429 & 87.3 & 68.0 & 65.5 \\
\bottomrule
\end{tabular}}
\par
\caption{\footnotesize \textbf{Where FreqEdit's correction acts} (FLUX.2 [klein], the 120 images of VAE-LFA, seeds 0 and 10). Ours (\method codec + BP) against FreqEdit with the stock VAE and 8-bit image hand-off, its own setting: at its original early denoising steps, on the last-step velocity, or once on the final latent with the turn's input latent as reference (BP's position and reference). Map: FreqEdit's default ($\alpha_0=1.6$, $\gamma=2$) or $\gamma=-10$ with $\alpha_0=1/(e^{-10}-1)$. Drift: LPIPS between round 10 and the input in the no-op and edit/undo protocols, with VAE-LFA's metric code. Edit success: percentage of the 1,200 edit/undo pairs for which EdiVal verifies the edit (Edit), the undo (Undo), or both (Cycle), as in Table~\ref{tab:vaelfa-cycle}. Bold: best per column.}
\label{tab:vaelfa-freqedit-last}
 \end{table}

{\section{Global edits}
\label{app:global-edits}

Figure~\ref{fig:global-edits} illustrates the limitation in Section~\ref{sec:limitations} on FLUX.2 [klein] 9B with 8-bit hand-offs. Each chain is a no-op turn followed by one global edit. In all six global edits we tried (watercolor, night, winter, golden-sunset lighting, black and white, and Van Gogh), BP reduced the LPIPS change of the edit turn, from 0.27--0.80 to 0.06--0.36; the codec alone follows edits like the stock pipeline.

\begin{figure*}
\centering
\input{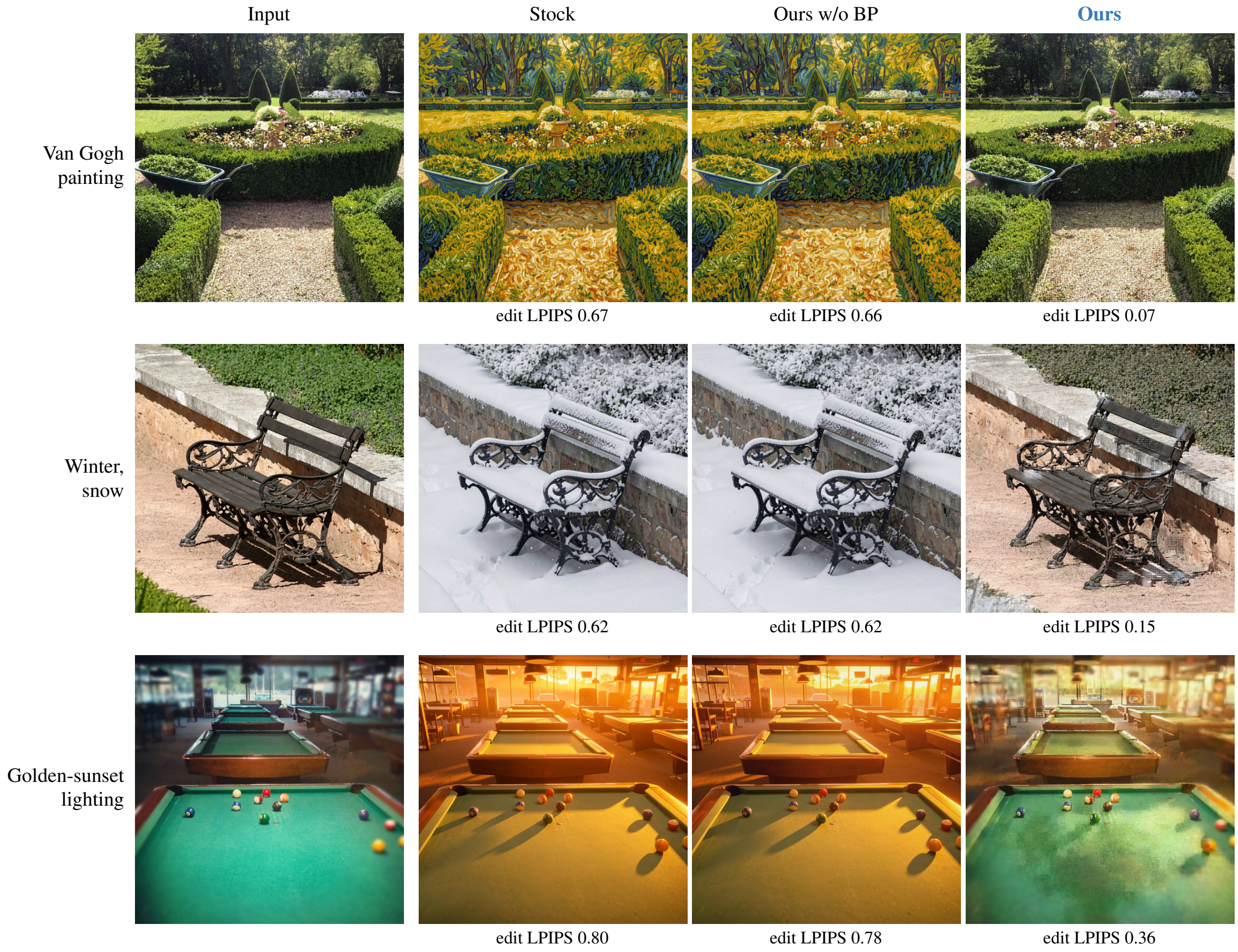}
\par
\caption{\footnotesize \textbf{Global edits are suppressed by BP.} Three of the six global edits we tried, on PasteBench scenes with FLUX.2 [klein] 9B and 8-bit hand-offs. Each row runs a no-op turn and then the global edit named on the left; the images are the outputs of the edit turn. Below each output: the LPIPS between the images before and after the edit turn; larger means a stronger edit. The codec alone (Ours w/o BP) applies the edits like the stock pipeline. With BP, the style edit is almost fully reverted.}
\label{fig:global-edits}
\end{figure*}
}
\FloatBarrier
\end{document}